\documentclass[]{spie}  

\usepackage{amsmath,amsfonts,amssymb}
\usepackage{graphicx}
\usepackage[colorlinks=true, allcolors=blue]{hyperref}
\usepackage{subcaption}
\usepackage{tabularx}

\title{Post-mastoidectomy Surface Multi-View Synthesis from a Single Microscopy Image}

\author[a]{Yike Zhang}
\author[a,b]{Jack H. Noble}
\affil[a]{Dept. of Computer Science, Vanderbilt University}
\affil[b]{Dept. of Electrical and Computer Engineering, Vanderbilt University}

\begin{document} 
\maketitle

\begin{abstract}
Cochlear Implant (CI) procedures involve performing an invasive mastoidectomy to insert an electrode array into the cochlea. In this paper, we introduce a novel pipeline that is capable of generating synthetic multi-view videos from a single CI microscope image. In our approach, we use a patient’s pre-operative CT scan to predict the post-mastoidectomy surface using a method designed for this purpose. We manually align the surface with a selected microscope frame to obtain an accurate initial pose of the reconstructed CT mesh relative to the microscope. We then perform UV projection to transfer the colors from the frame to surface textures. Novel views of the textured surface can be used to generate a large dataset of synthetic frames with ground truth poses. We evaluated the quality of synthetic views rendered using Pytorch3D and PyVista. We found both rendering engines lead to similarly high-quality synthetic novel-view frames compared to ground truth with a structural similarity index for both methods averaging $\sim$0.86. A large dataset of novel views with known poses is critical for ongoing training of a method to automatically estimate microscope pose for 2D to 3D registration with the pre-operative CT to facilitate augmented reality surgery. This dataset will empower various downstream tasks, such as integrating Augmented Reality (AR) in the OR, tracking surgical tools, and supporting other video analysis studies.
\end{abstract}


\section{INTRODUCTION}
Cochlear Implant (CI) procedures are performed to restore hearing in patients with moderate-to-profound hearing loss \cite{labadie2018preliminary} . CI surgery requires an invasive mastoidectomy procedure, which involves removing part of the temporal bone, to obtain a clear view and safe access to the cochlea. This surgery demands precision and accuracy, as some of the important structures, such as facial nerve, chorda, and ossicles, are hidden or heavily occluded (Zhang et al. \cite{yike2022, yike2023}). One of our primary goals is to develop an augmented reality (AR) system to improve CI procedures. AR would enable enhancing the surgical field visualization with hidden structures that are reconstructed from a patient’s CT scan. AR could also play a crucial role in helping to optimally place the electrodes within the cochlea, thereby maximizing the success rates of restoring hearing \cite{labadie2018preliminary} . A critical step for AR surgery is the image-to-physical co-registration of the pre-operative plan created using CT to the microscope video. To facilitate this 3D to 2D registration, we aim to investigate modern deep learning techniques that can estimate the pose of the microscope while only having microscope video as input. However, training such a deep learning-based pose estimation model for CI surgery requires a large dataset that needs to contain diverse microscopy views. Unlike other medical procedures where large amounts of videos are available, CI surgeries are less frequently recorded and do not include pose tracking, leading to a data shortage problem. To address this challenge, there is a need for a method that can generate realistic microscopy views from limited data. Herein, we rely on a single microscopy video frame registered to a post-mastoidectomy surface estimated from the corresponding CT scan to generate and colorize the surface so that it can be virtually viewed at novel poses to create large amounts of realistic synthetic visualizations. 

\section{Method}
\begin{figure}[h]
    \centering
    \begin{subfigure}[b]{0.3\textwidth}
        \centering
        \includegraphics[width=0.5\textwidth]{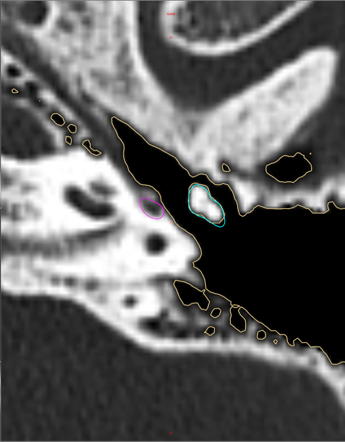}
        \caption{CT scan}
        \label{fig:ct}
    \end{subfigure}
    \begin{subfigure}[b]{0.3\textwidth}
        \centering
        \includegraphics[width=0.5\textwidth]{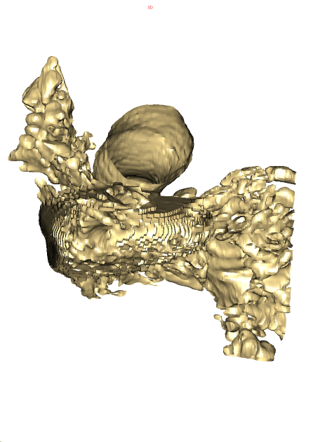}
        \caption{Reconstructed CT mesh}
        \label{fig:ct_mesh}
    \end{subfigure}
    \begin{subfigure}[b]{0.3\textwidth}
        \centering
        \includegraphics[width=0.5\textwidth]{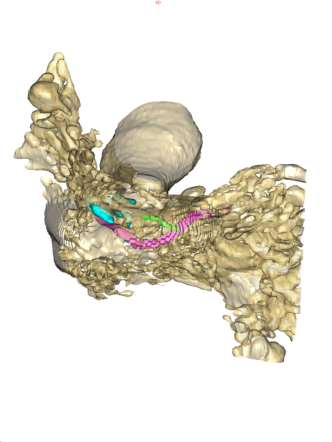}
        \caption{Ear structures overlay}
        \label{fig:meshes_overlay}
    \end{subfigure}
    \caption{Extracted Meshes from CT scan.}
    \label{fig:extracted_meshes}
\end{figure}

\begin{figure}[h]
    \centering
    \begin{subfigure}[b]{0.45\textwidth}
        \centering
        \includegraphics[width=0.5\textwidth]{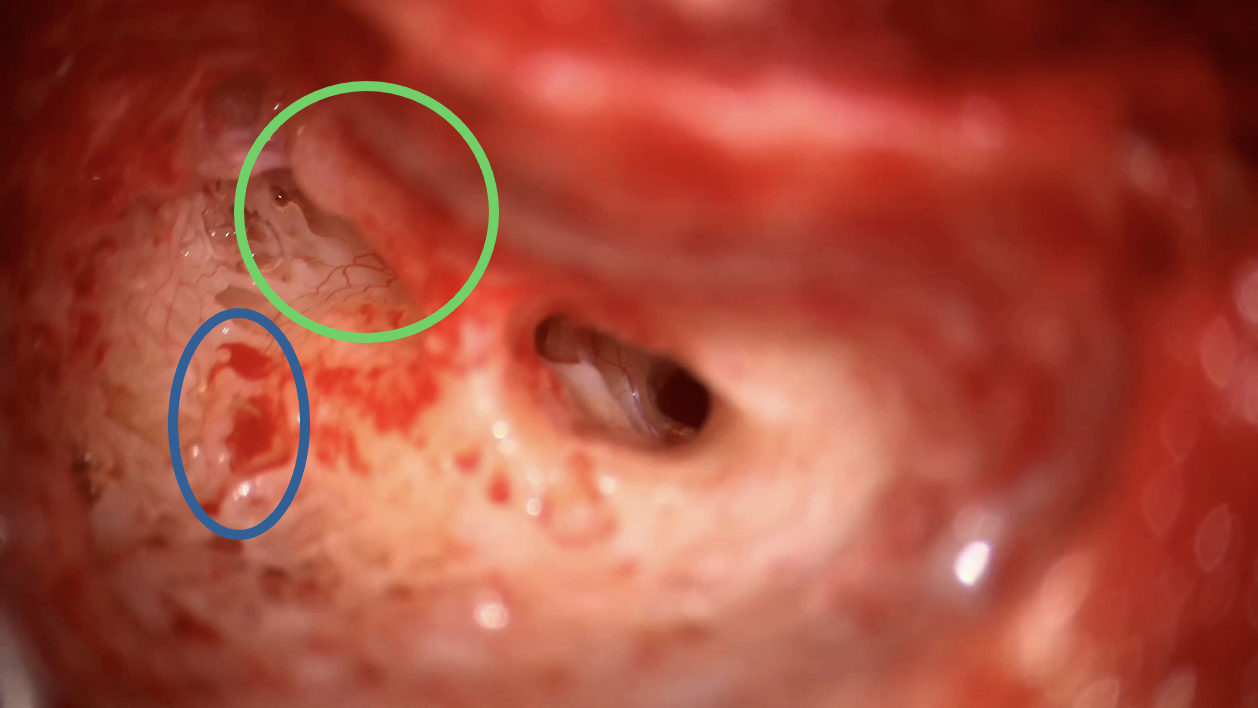}
        \caption{Selected microscope view}
    \end{subfigure}
    \begin{subfigure}[b]{0.45\textwidth}
        \centering
        \includegraphics[width=0.5\textwidth]{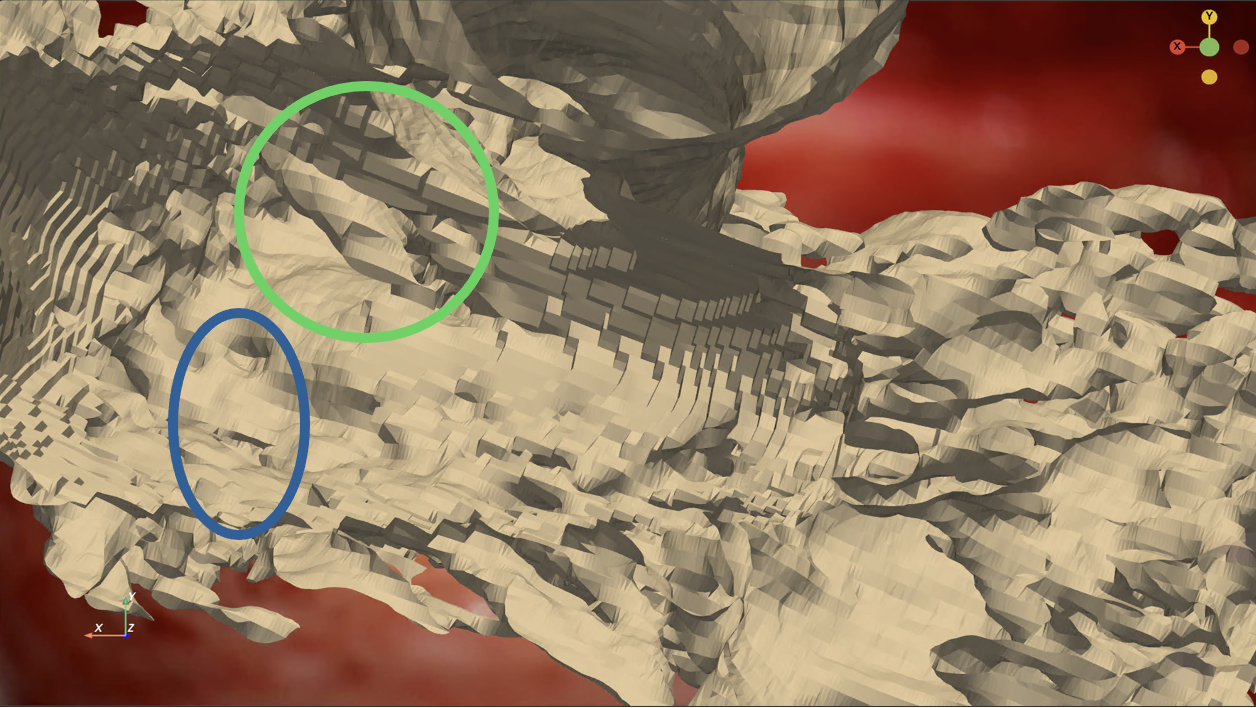}
        \caption{Register CT mesh to the microscope view}
    \end{subfigure}
    \caption{Register CT mesh to the microscope view.}
    \label{fig:vision6D}
\end{figure}

First, we process a patient’s pre-operative CT scan to predict the post-mastoidectomy surface mesh using the proposed method in Zhang et al. \cite{zhang2024unsupervisedmastoidectomycochlearct} isolating only the bone structure while discarding the soft tissue and air present in the scan. The CT scan is displayed in Fig. \ref{fig:ct}, and the whole surface is shown in Fig. \ref{fig:ct_mesh}. We also demonstrate the essential ear structures, such as ossicles (highlighted in aqua), facial nerve (highlighted in magenta), and chorda (highlighted in green), and overlay them onto the reconstructed CT mesh in Fig. \ref{fig:meshes_overlay}. After obtaining the vertex coordinates $V \in \mathbb{R}^{N_v\times3}$ from the surface mesh, we demean it using $V' = V - \frac{1}{N_v}\sum^{N_v}_{i=1}V_{i}$ to get $V'$, where $N_v$ is the total number of vertices in $V$. 
A single frame $I$ is selected from the recorded video that presents a clear microscopy view, free from any occlusion either caused by the surgeon’s hands or the surgical tools. Given a preset camera intrinsic matrix, we manually aligned the surface with the selected frame in our customized open-source software Vision6D\cite{Zhang_Vision6D} to obtain an accurate initial pose of the reconstructed CT mesh, as shown in Fig. \ref{fig:vision6D}. The corresponding features between the microscopy view and the CT mesh are circled in matching colors.
To paint the texture-less CT mesh, we perform UV projection to color $V_i$ with the color shown in $I$. We use Eq.\ref{eq:projection} first to project $V_i$ to the image plane, sample the pixel values, and linearly interpolate the RGB colors $I(\frac{u}{w}, \frac{v}{w})$, and assign them to color $V_i$. This is repeated for each vertex $V_i, i = 1, ..., N_v$.
\begin{equation}
    \begin{pmatrix} 
        u \\
        v \\
        w \\
    \end{pmatrix}
    =
    \begin{pmatrix}
        f_x & 0 & c_x & 0 \\
        0 & f_y & c_y & 0 \\
        0 & 0 & 1 & 0 \\
    \end{pmatrix}
    \begin{pmatrix}
        R_{3 \times 3} & t_{3 \times 1} \\
        0_{1 \times 3} & 1_{1 \times 1} \\
    \end{pmatrix}
    \begin{pmatrix}
        X \\
        Y \\
        Z \\
        1
    \end{pmatrix}
\label{eq:projection}
\end{equation}
where $\begin{pmatrix}
        f_x & 0 & c_x & 0 \\
        0 & f_y & c_y & 0 \\
        0 & 0 & 1 & 0 \\
    \end{pmatrix}$ is the camera intrinsic matrix, 
    $\begin{pmatrix}
        R_{3 \times 3} & t_{3 \times 1} \\
        0_{1 \times 3} & 1_{1 \times 1} \\
    \end{pmatrix}$ is the camera extrinsic matrix, $\begin{pmatrix}
        X, Y, Z, 1
    \end{pmatrix}$ is the homogeneous coordinates representation of the vertices in $V$, and $\begin{pmatrix} 
        u, v, w
    \end{pmatrix}$ is the image coordinates. In Eq. \ref{eq:projection}, the world coordinates $\begin{pmatrix}
        X, Y, Z, 1
    \end{pmatrix}$ are transformed by the camera extrinsic matrix, which combines the rotation matrix $R$ and the translation vector $t$. The resulting camera coordinates are then transformed by the camera intrinsic matrix, which applies the camera's focal length $f_x$, $f_y$ and principle points $c_x$, $c_y$, resulting in projected image coordinates $\begin{pmatrix} u, v, w \end{pmatrix}$. After that, the image coordinates $I(x, y)$ can be computed as $I(\frac{u}{w}, \frac{v}{w})$. Thus, the RGB color can be mapped from the linearly interpolated point $I(x, y)$ in the image plane to $V_i$. Finally, $V$ is textured with the RGB information $C \in \mathbb{R}^{N_v\times3}$ obtained from $I$. For synthesizing the microscopy multi-view, we randomly generate a set of camera poses $P_i \in SE(3)$, and thus we obtain synthetic images with known ground-truth poses $P$. We compare the performance of two widely-used 3D rendering libraries, PyTorch3D\cite{ravi2020pytorch3d} and PyVista\cite{sullivan2019pyvista}, in rendering novel views relative to new surgical video frames.

\section{results}
\label{sec:sections}
\begin{figure}[h]
    \centering
    \begin{minipage}{0.11\textwidth}
        \includegraphics[width=\textwidth]{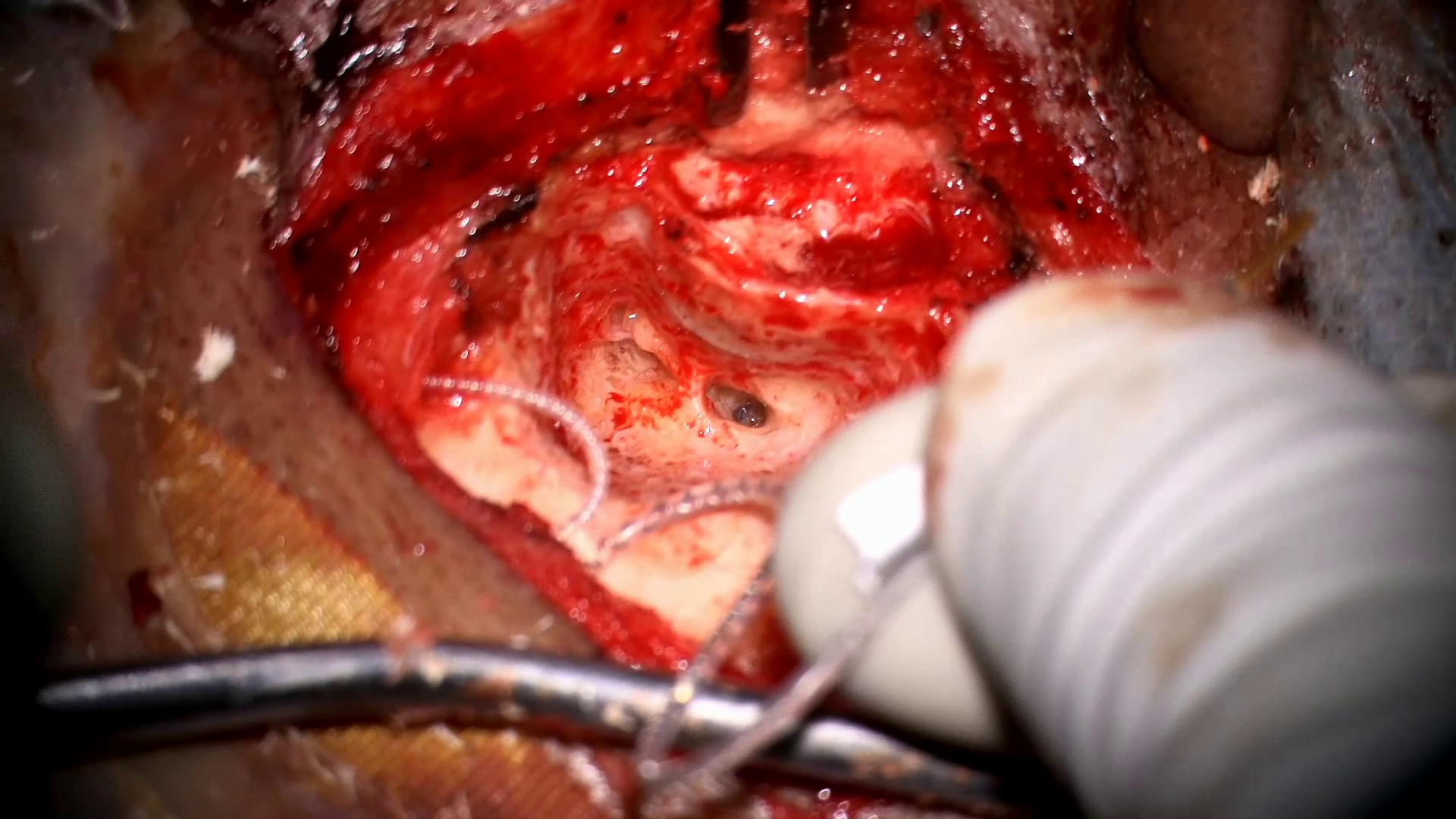}
    \end{minipage}
    \hfill
    \begin{minipage}{0.11\textwidth}
        \includegraphics[width=\textwidth]{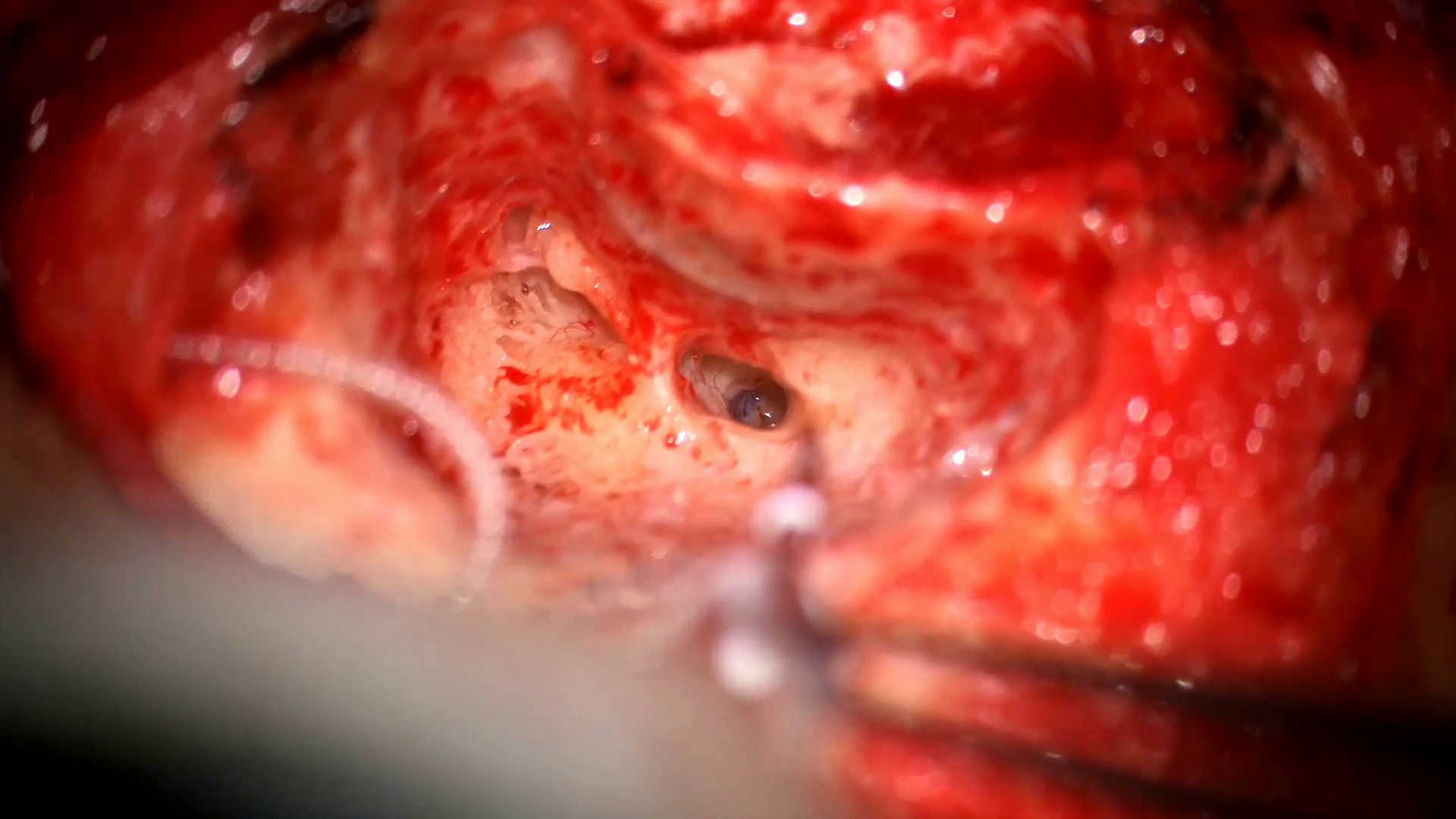}
    \end{minipage}
    \hfill
    \begin{minipage}{0.11\textwidth}
        \includegraphics[width=\textwidth]{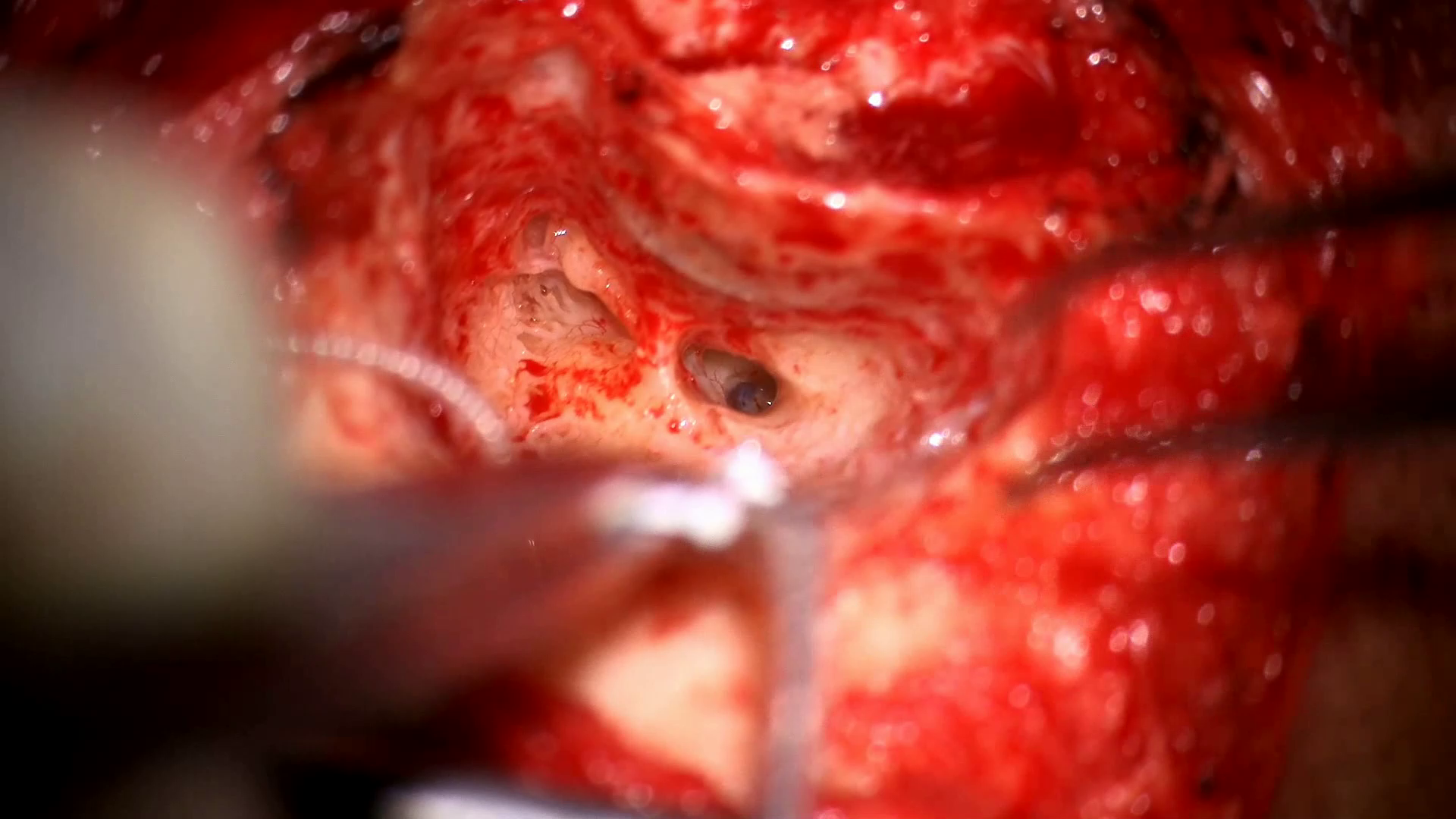}
    \end{minipage}
    \hfill
    \begin{minipage}{0.11\textwidth}
        \includegraphics[width=\textwidth]{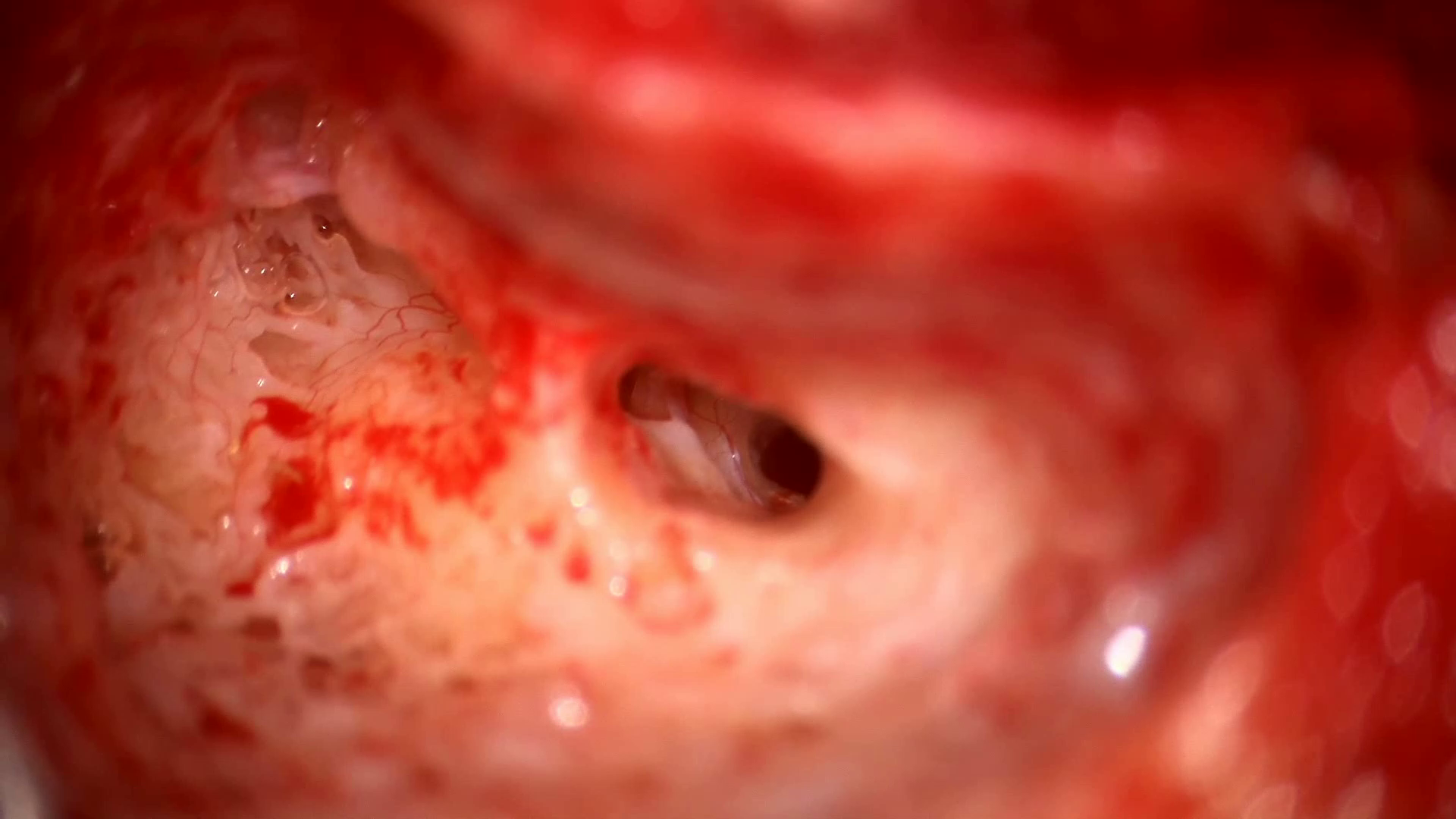}
    \end{minipage}
    \hfill
    \begin{minipage}{0.11\textwidth}
        \includegraphics[width=\textwidth]{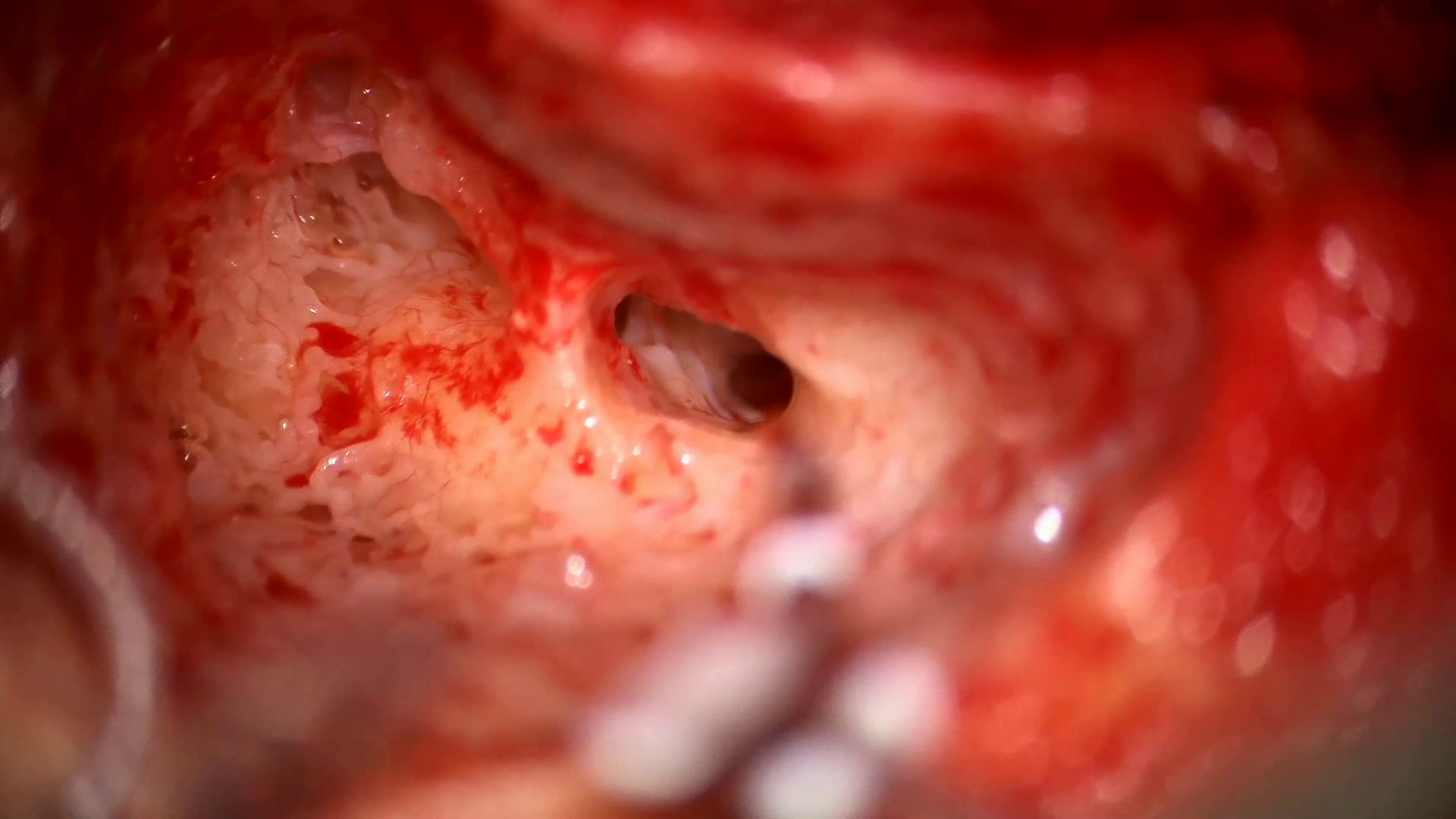}
    \end{minipage}
    \hfill
    \begin{minipage}{0.11\textwidth}
        \includegraphics[width=\textwidth]{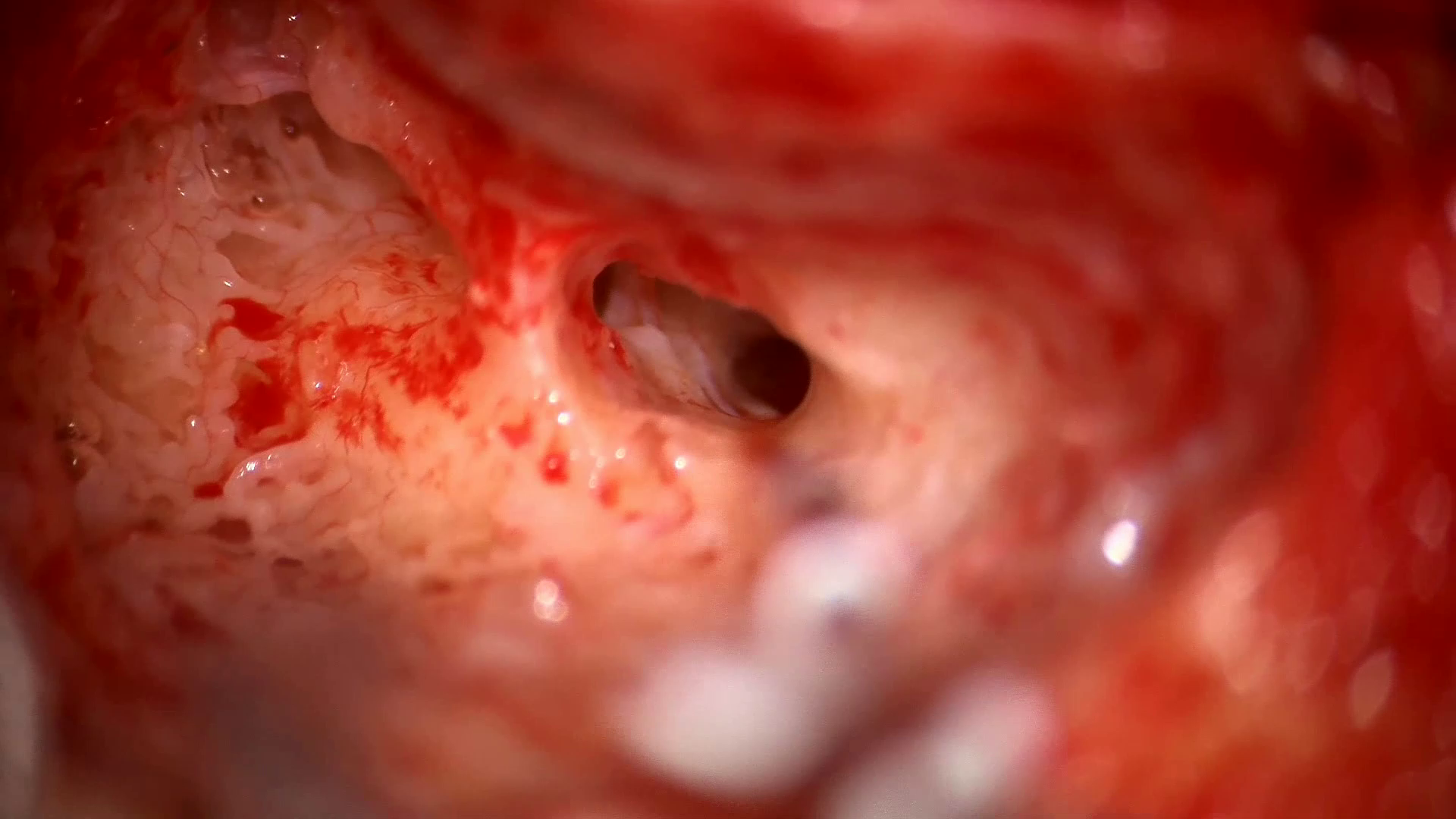}
    \end{minipage}
    \hfill
    \begin{minipage}{0.11\textwidth}
        \includegraphics[width=\textwidth]{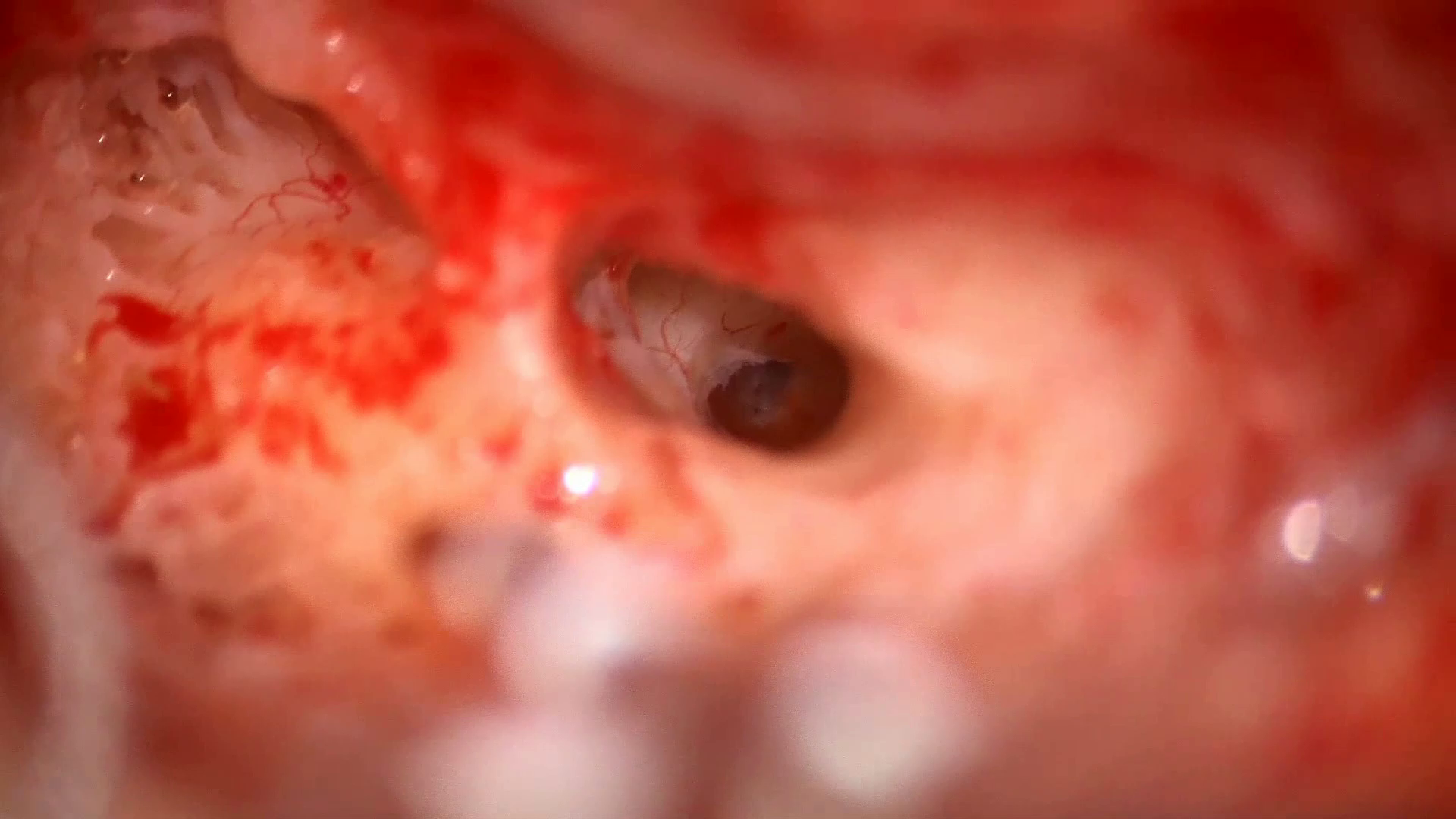}
    \end{minipage}
    \hfill
    \begin{minipage}{0.11\textwidth}
        \includegraphics[width=\textwidth]{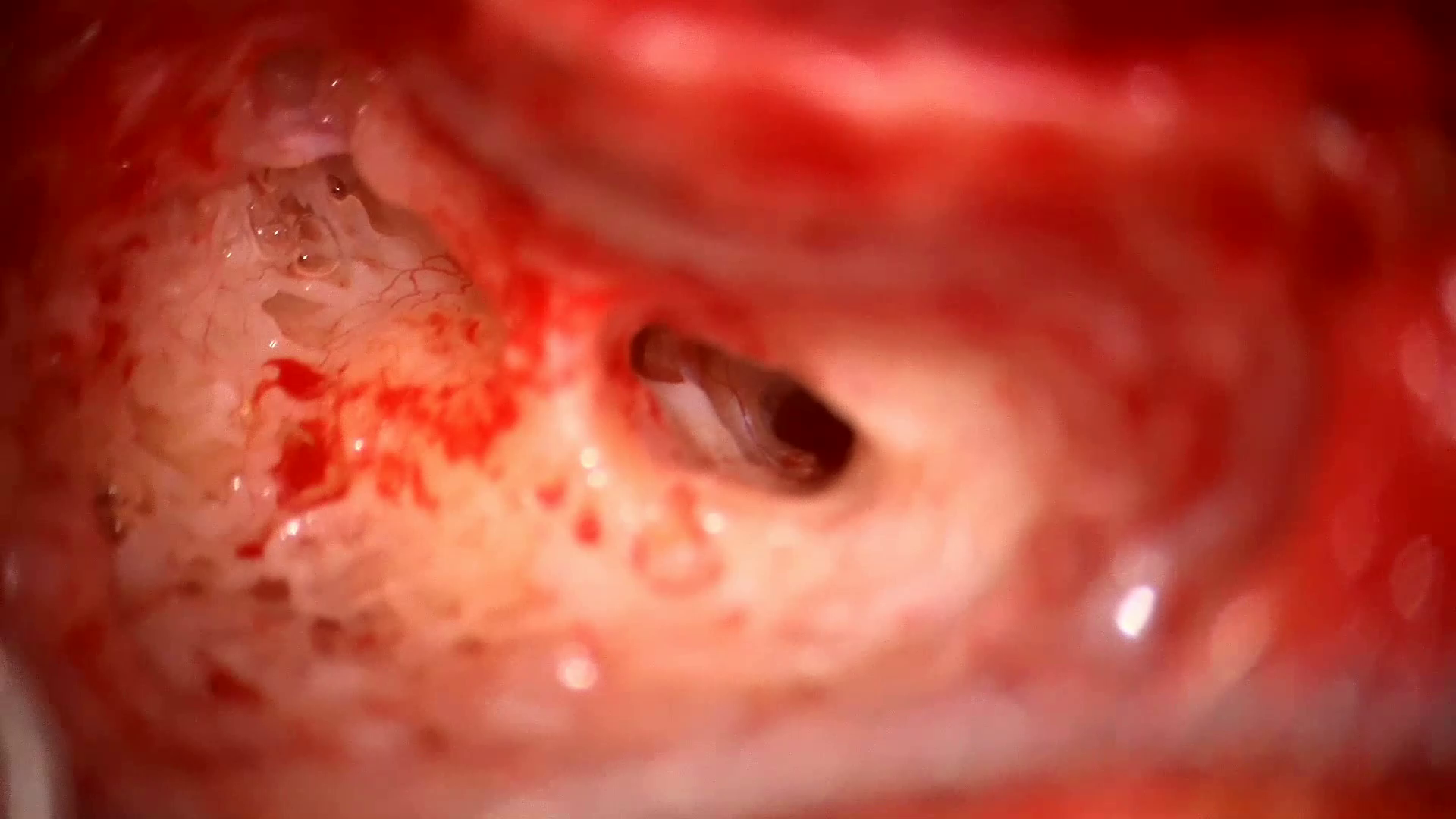}
    \end{minipage}
    \hfill
    \begin{minipage}{0.11\textwidth}
        \includegraphics[width=\textwidth]{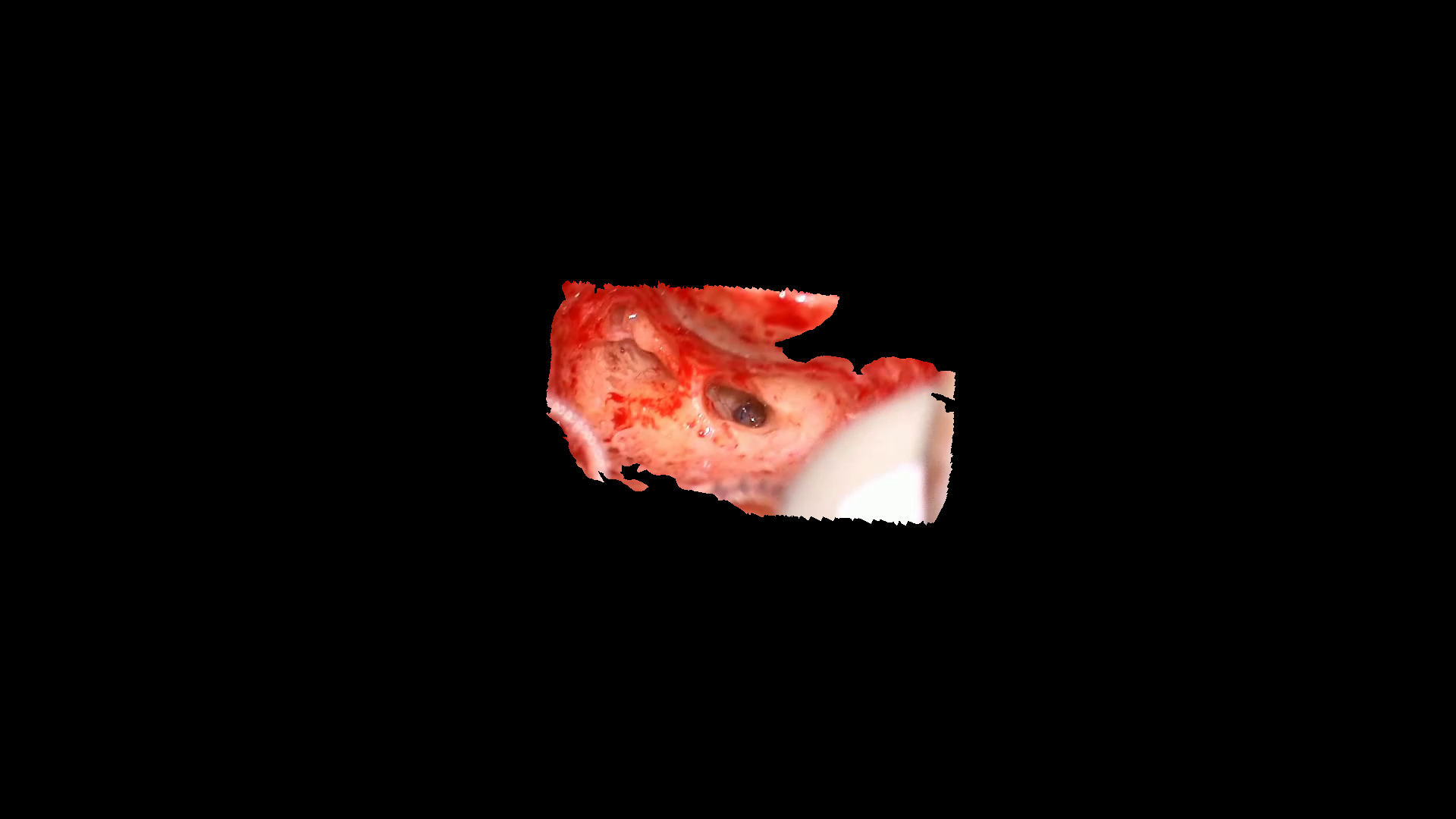}
    \end{minipage}
    \hfill
    \begin{minipage}{0.11\textwidth}
        \includegraphics[width=\textwidth]{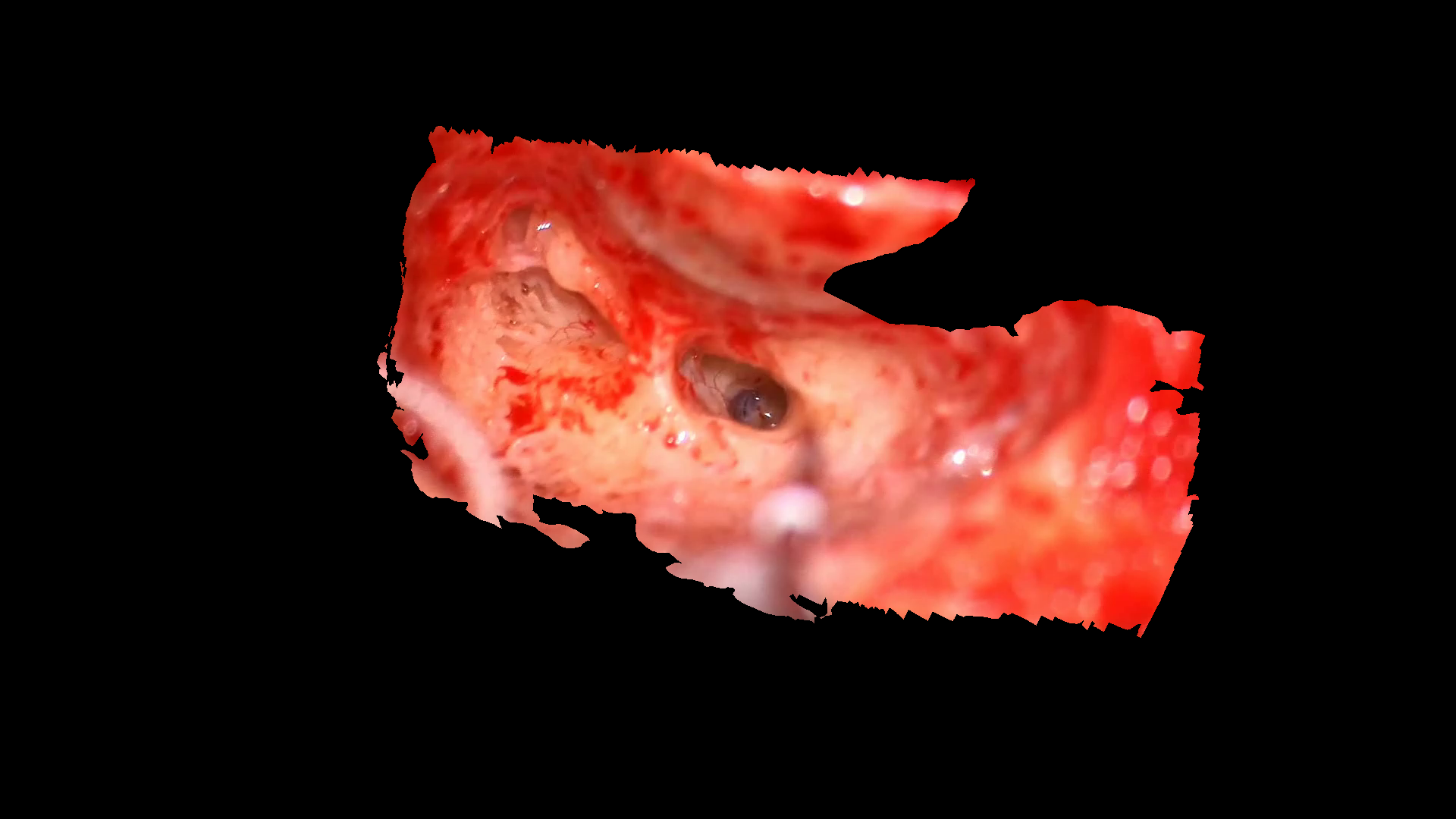}
    \end{minipage}
    \hfill
    \begin{minipage}{0.11\textwidth}
        \includegraphics[width=\textwidth]{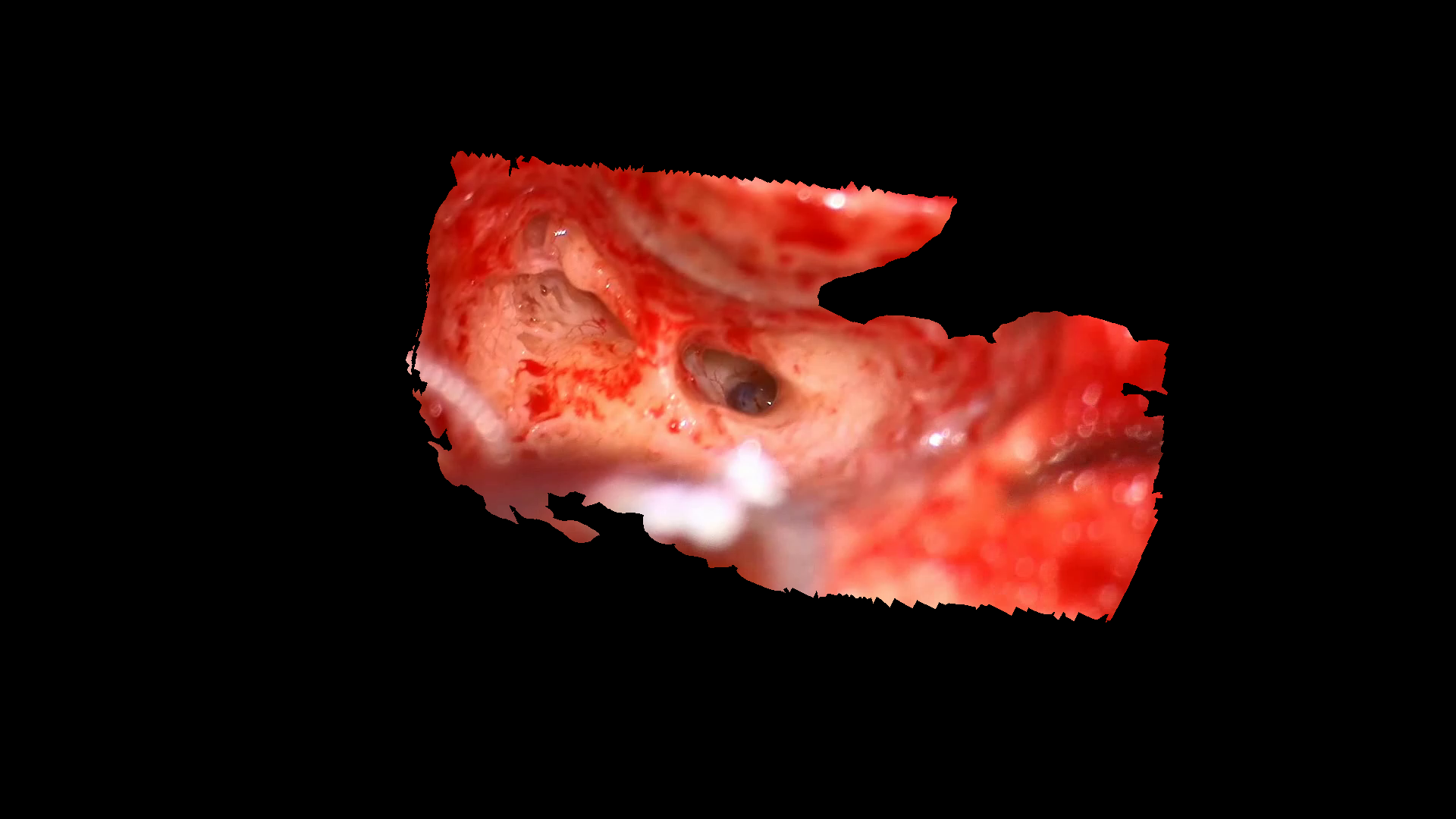}
    \end{minipage}
    \hfill
    \begin{minipage}{0.11\textwidth}
        \includegraphics[width=\textwidth]{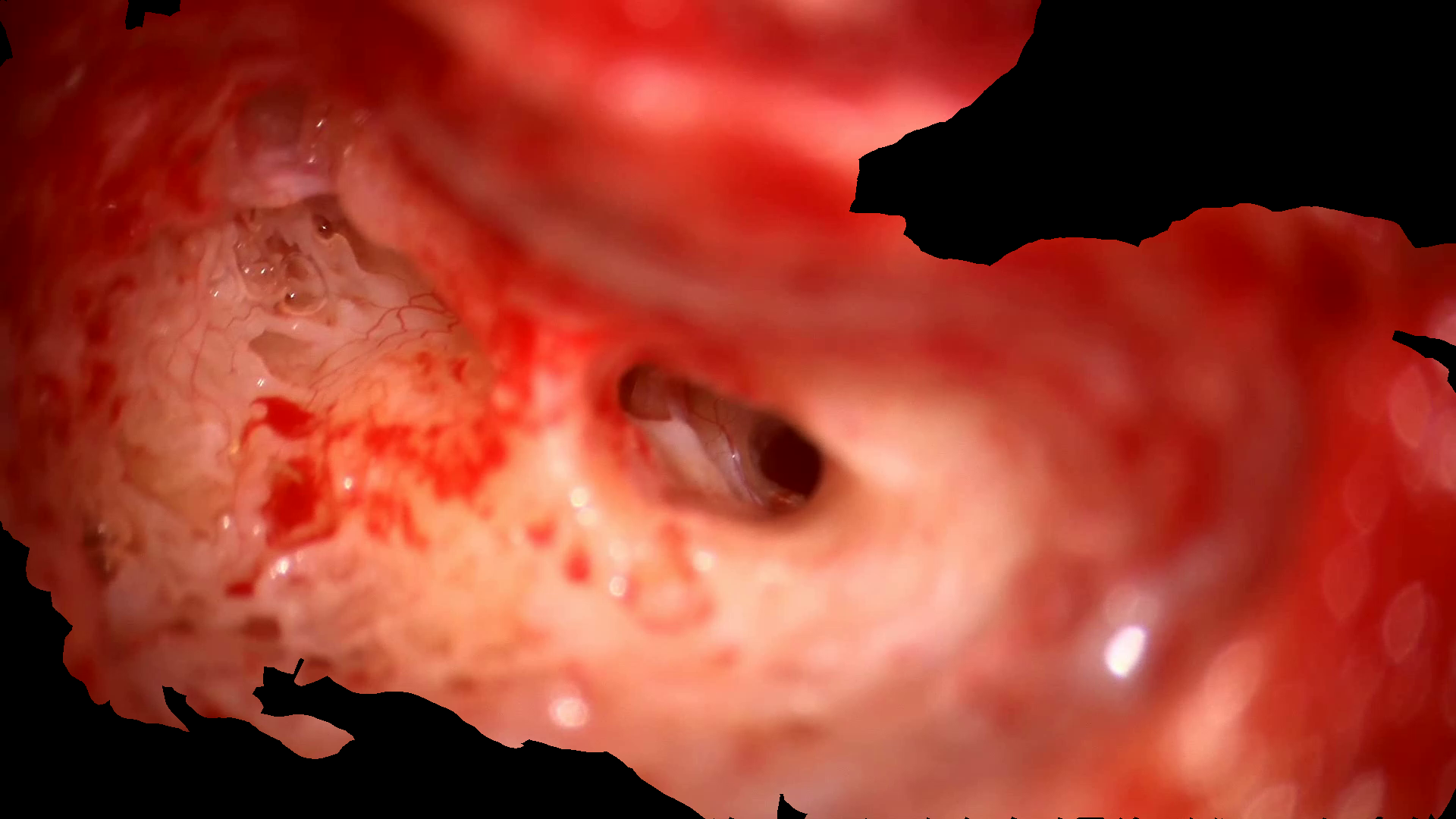}
    \end{minipage}
    \hfill
    \begin{minipage}{0.11\textwidth}
        \includegraphics[width=\textwidth]{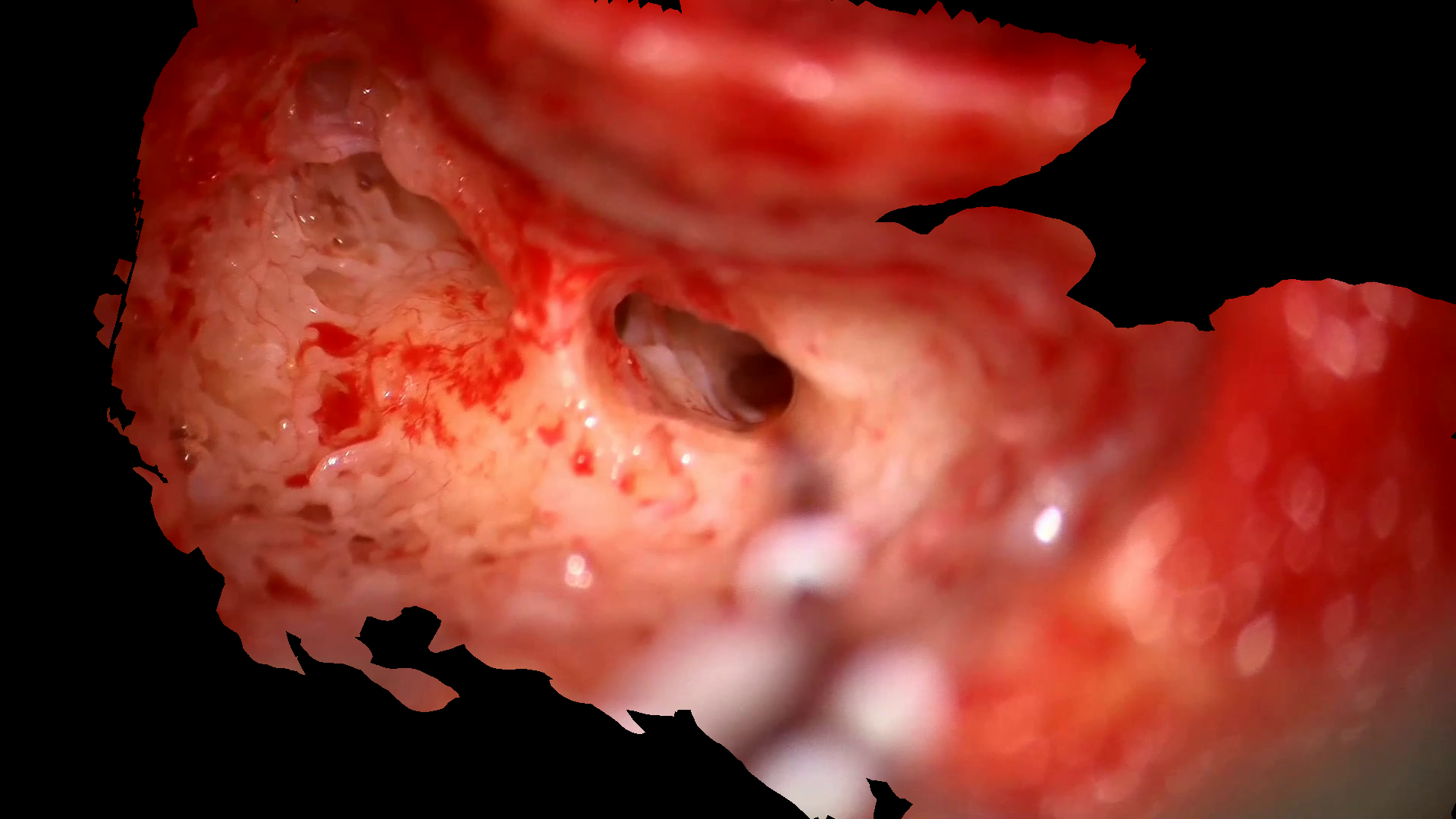}
    \end{minipage}
    \hfill
    \begin{minipage}{0.11\textwidth}
        \includegraphics[width=\textwidth]{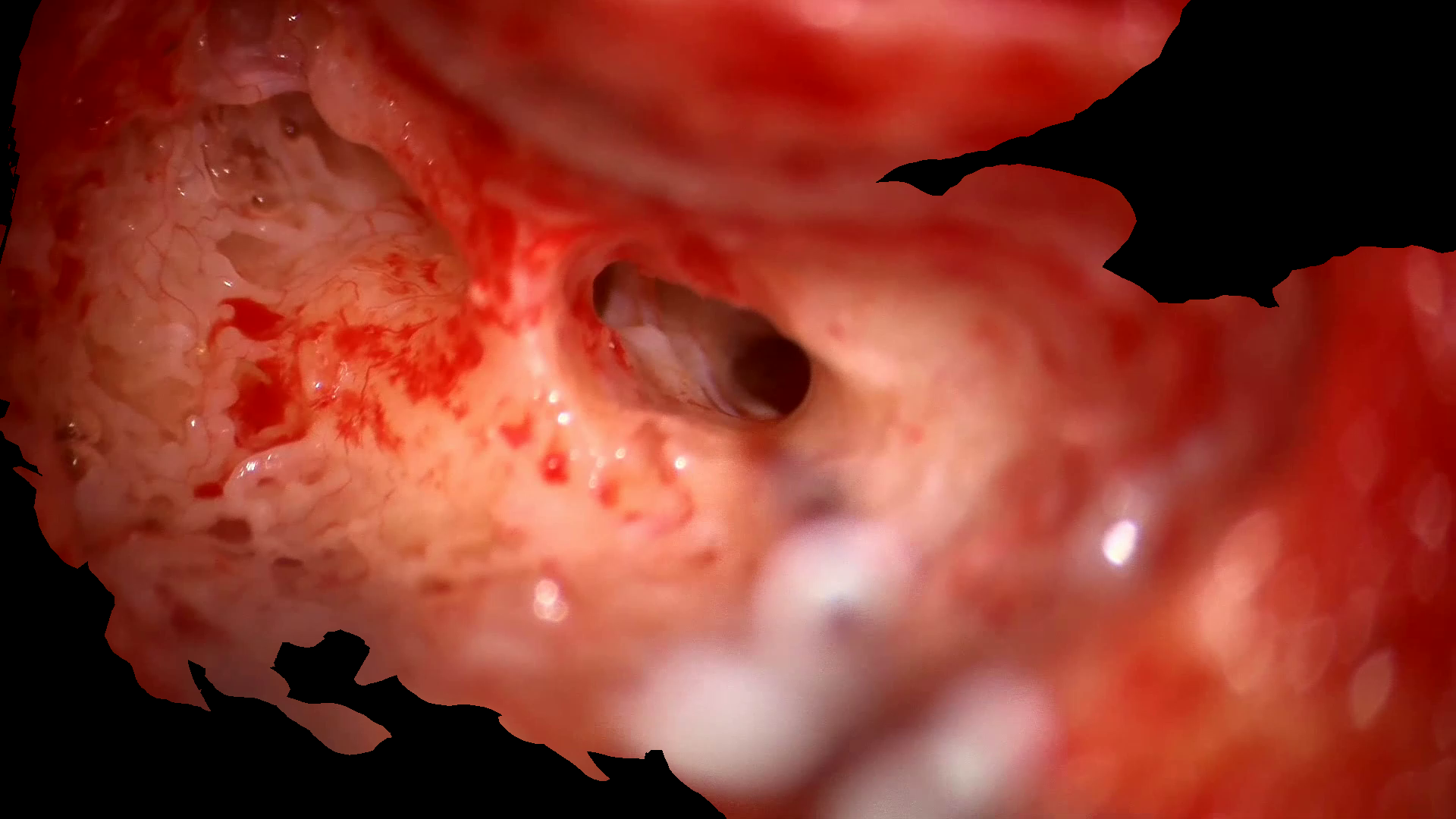}
    \end{minipage}
    \hfill
    \begin{minipage}{0.11\textwidth}
        \includegraphics[width=\textwidth]{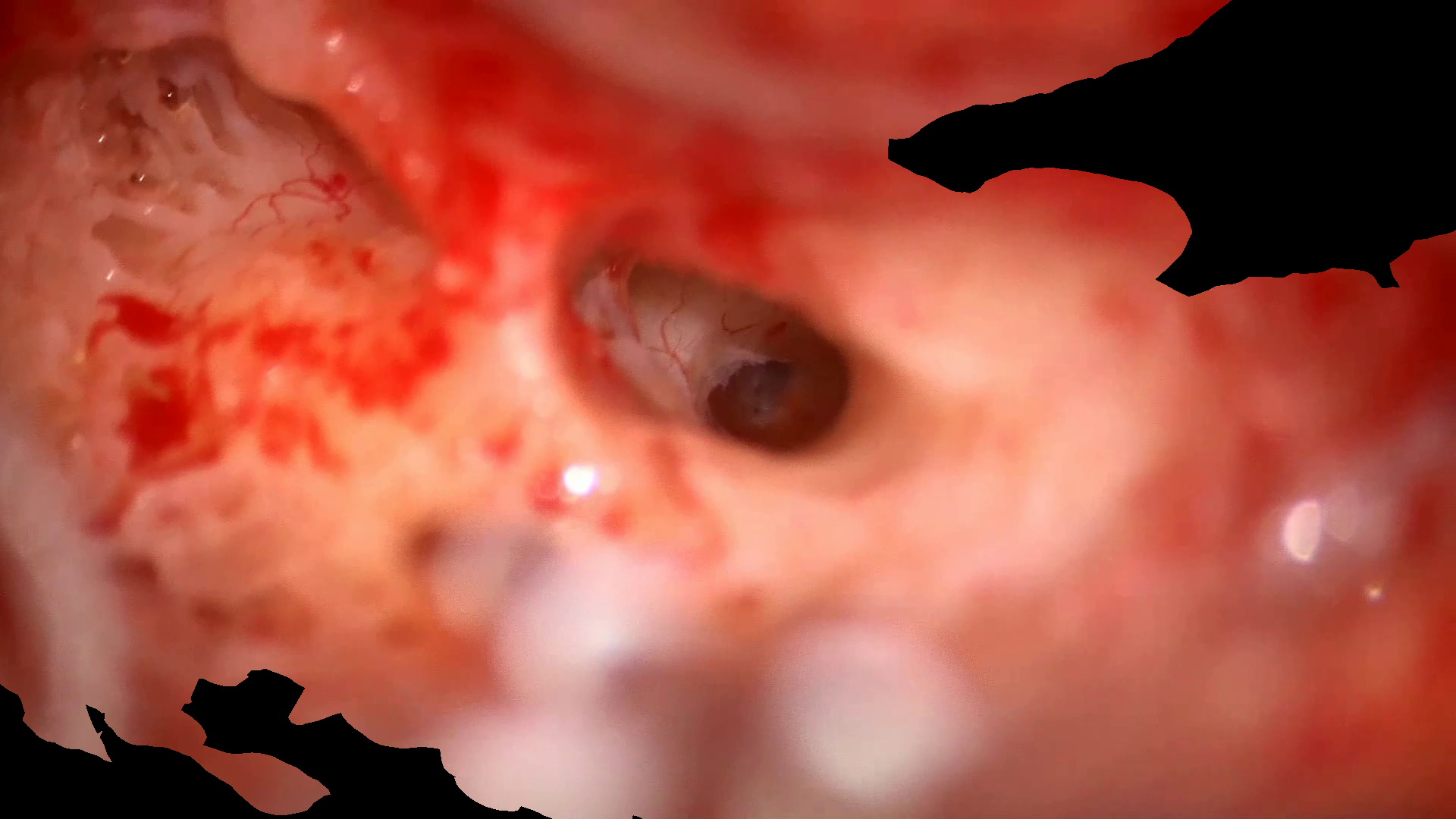}
    \end{minipage}
    \hfill
    \begin{minipage}{0.11\textwidth}
        \includegraphics[width=\textwidth]{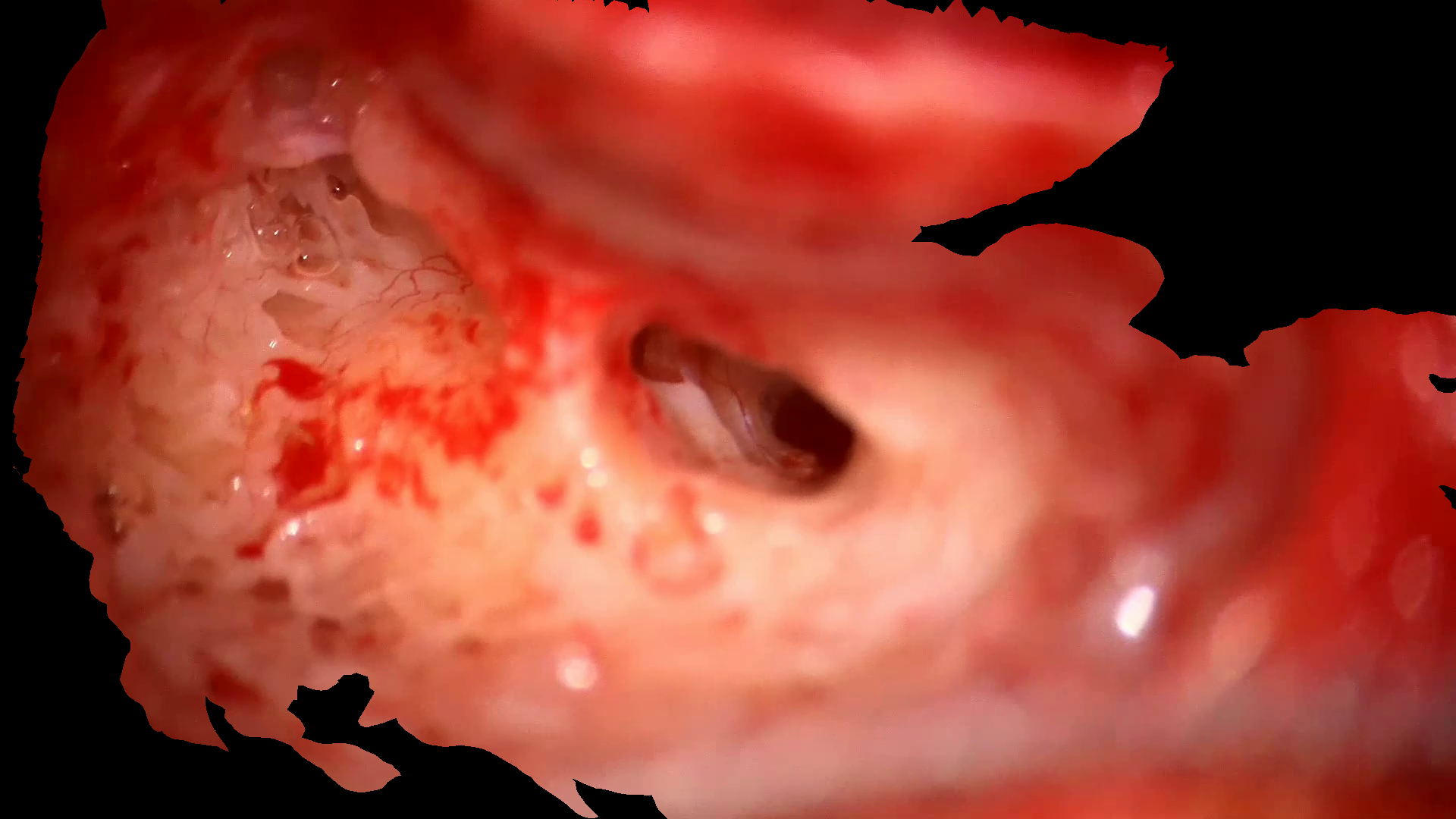}
    \end{minipage}
    \begin{minipage}{0.11\textwidth}
        \includegraphics[width=\textwidth]{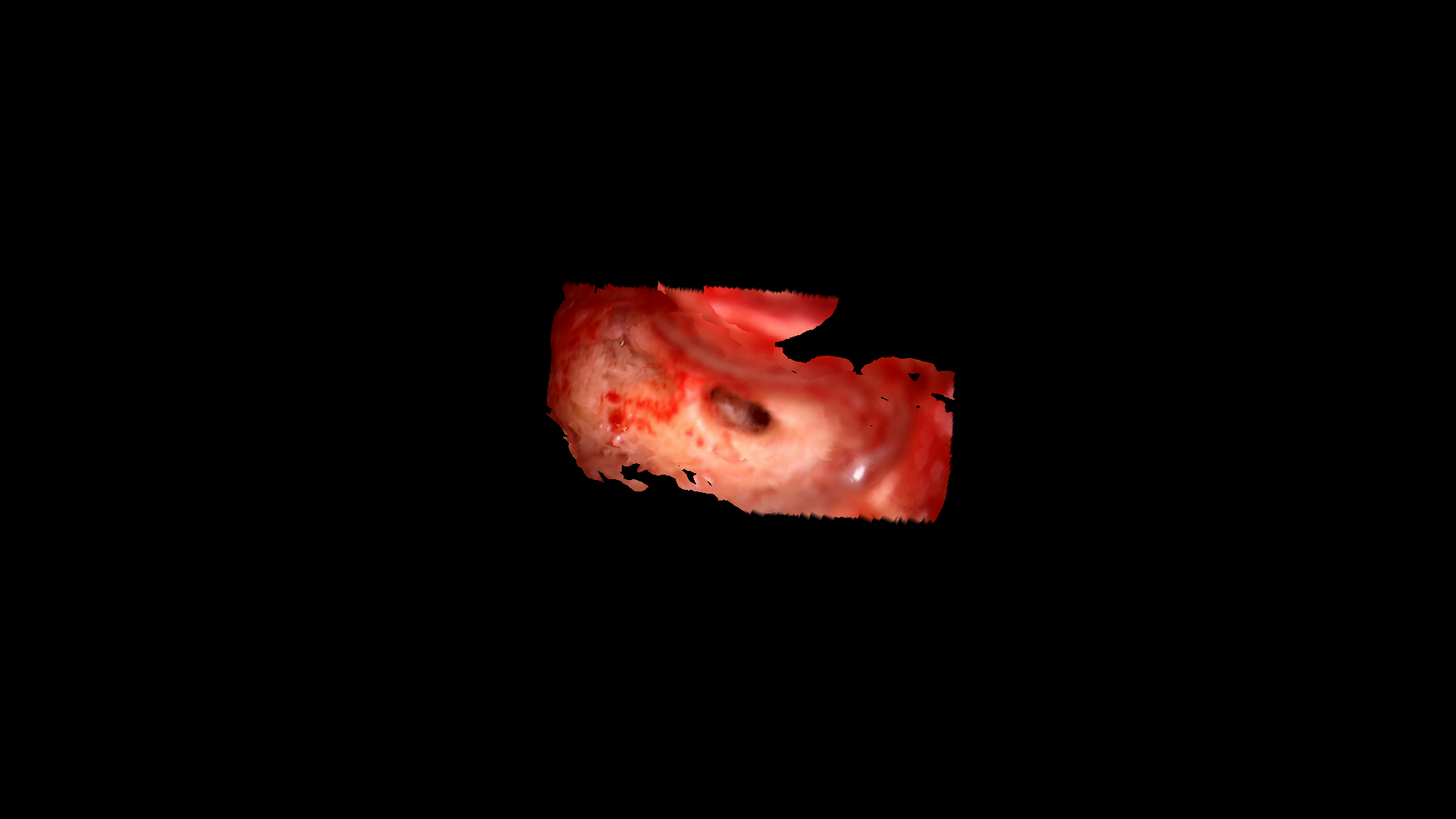}
    \end{minipage}
    \hfill
    \begin{minipage}{0.11\textwidth}
        \includegraphics[width=\textwidth]{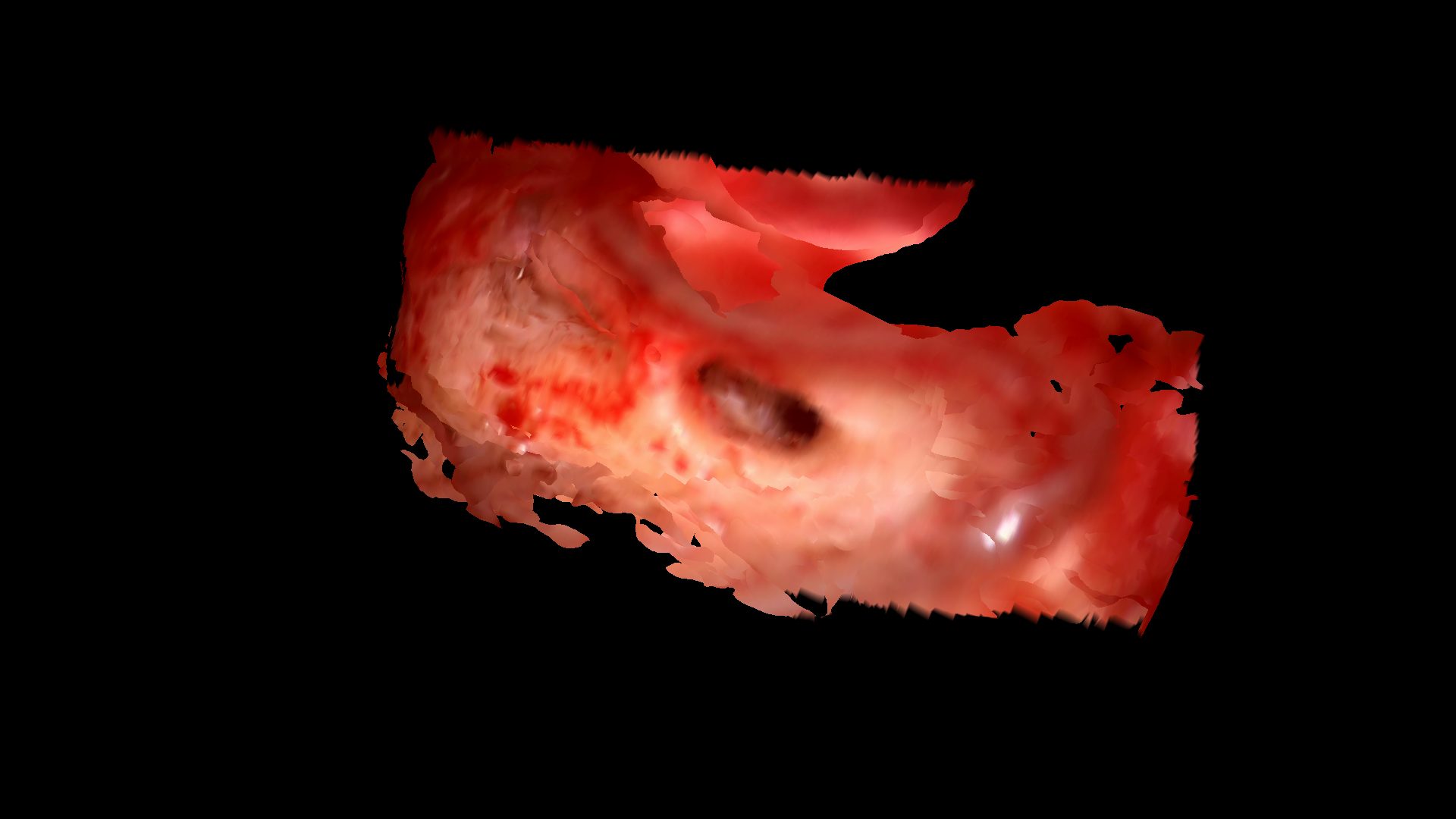}
    \end{minipage}
    \hfill
    \begin{minipage}{0.11\textwidth}
        \includegraphics[width=\textwidth]{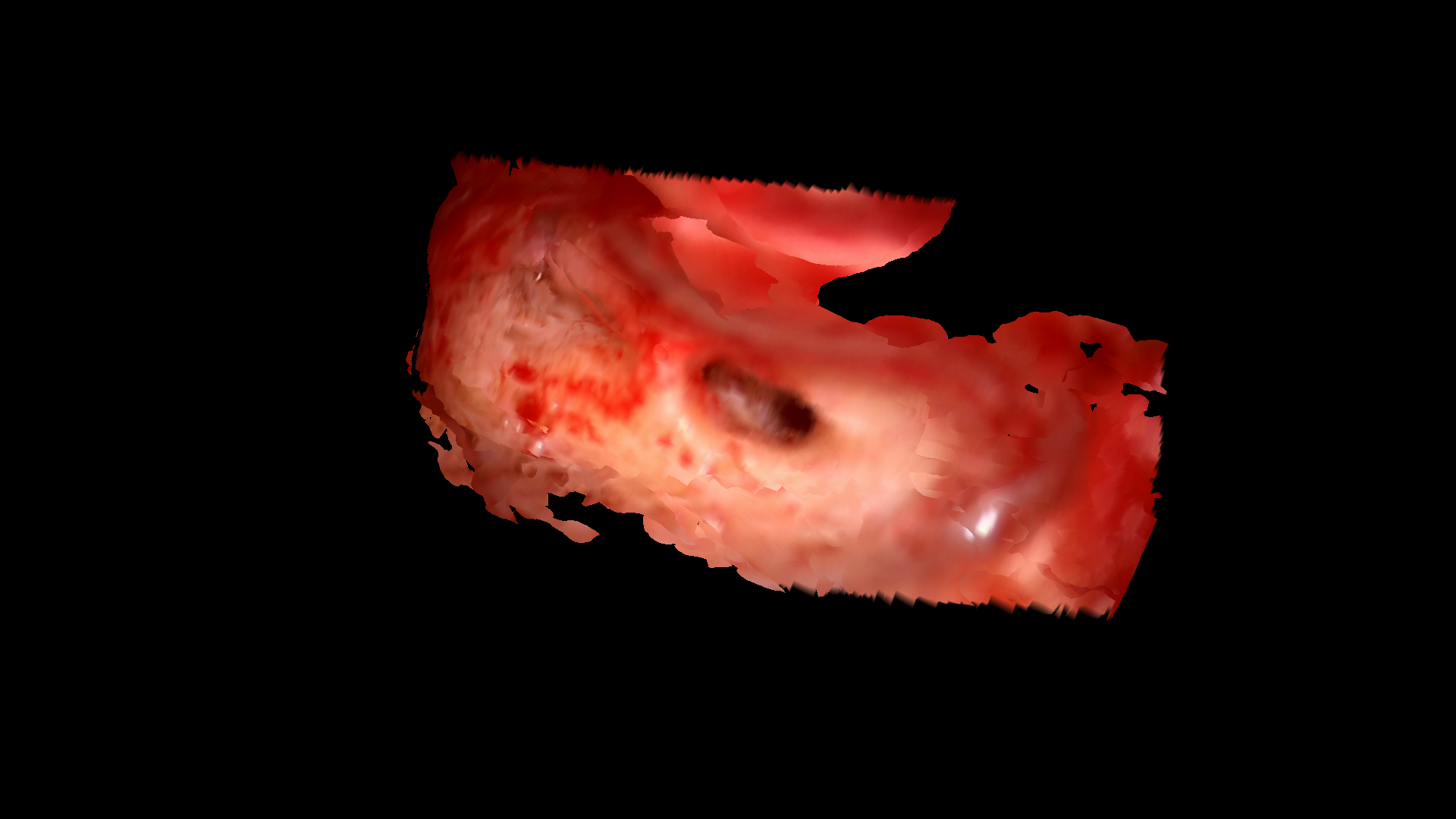}
    \end{minipage}
    \hfill
    \begin{minipage}{0.11\textwidth}
        \includegraphics[width=\textwidth]{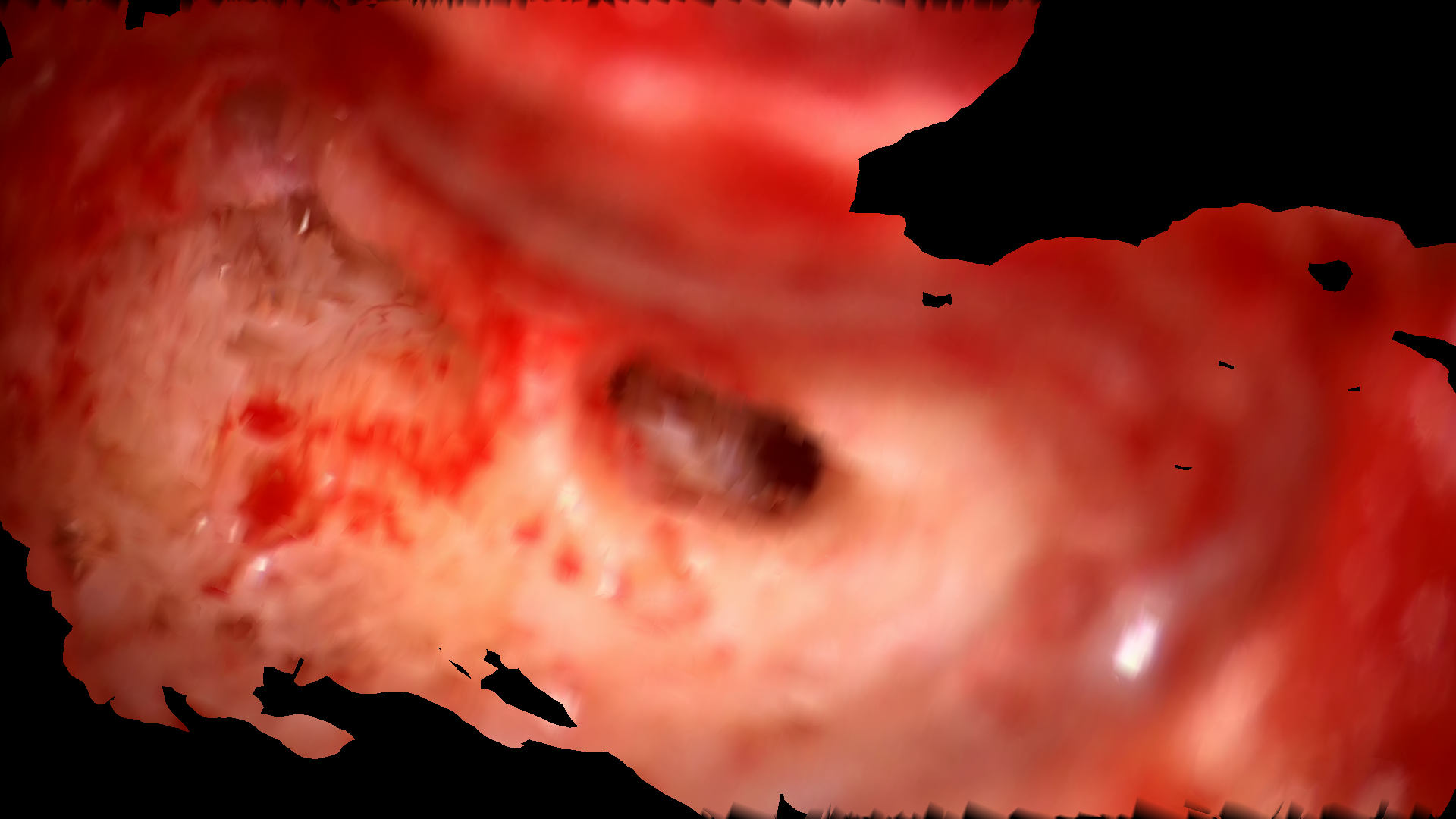}
    \end{minipage}
    \hfill
    \begin{minipage}{0.11\textwidth}
        \includegraphics[width=\textwidth]{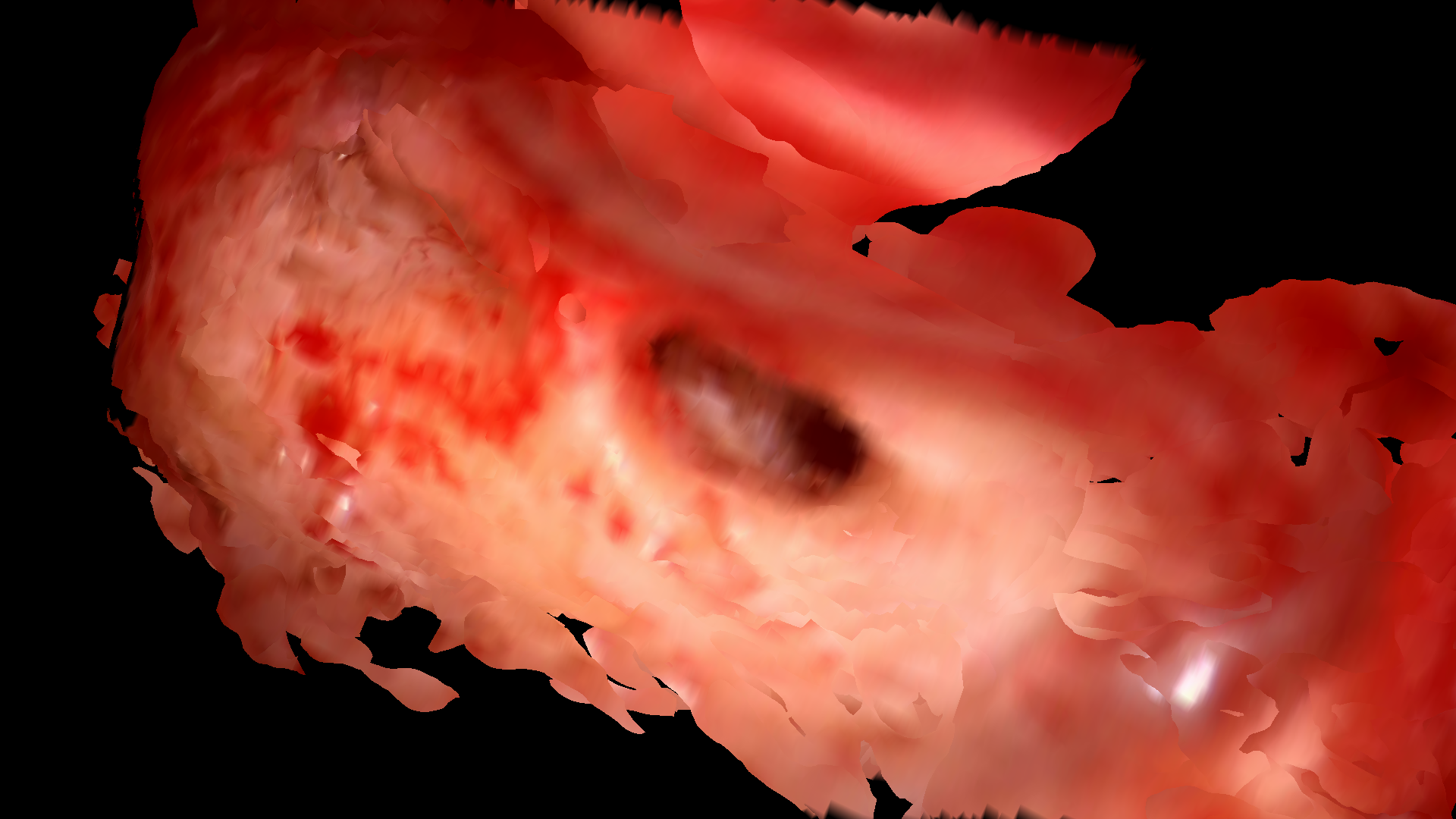}
    \end{minipage}
    \hfill
    \begin{minipage}{0.11\textwidth}
        \includegraphics[width=\textwidth]{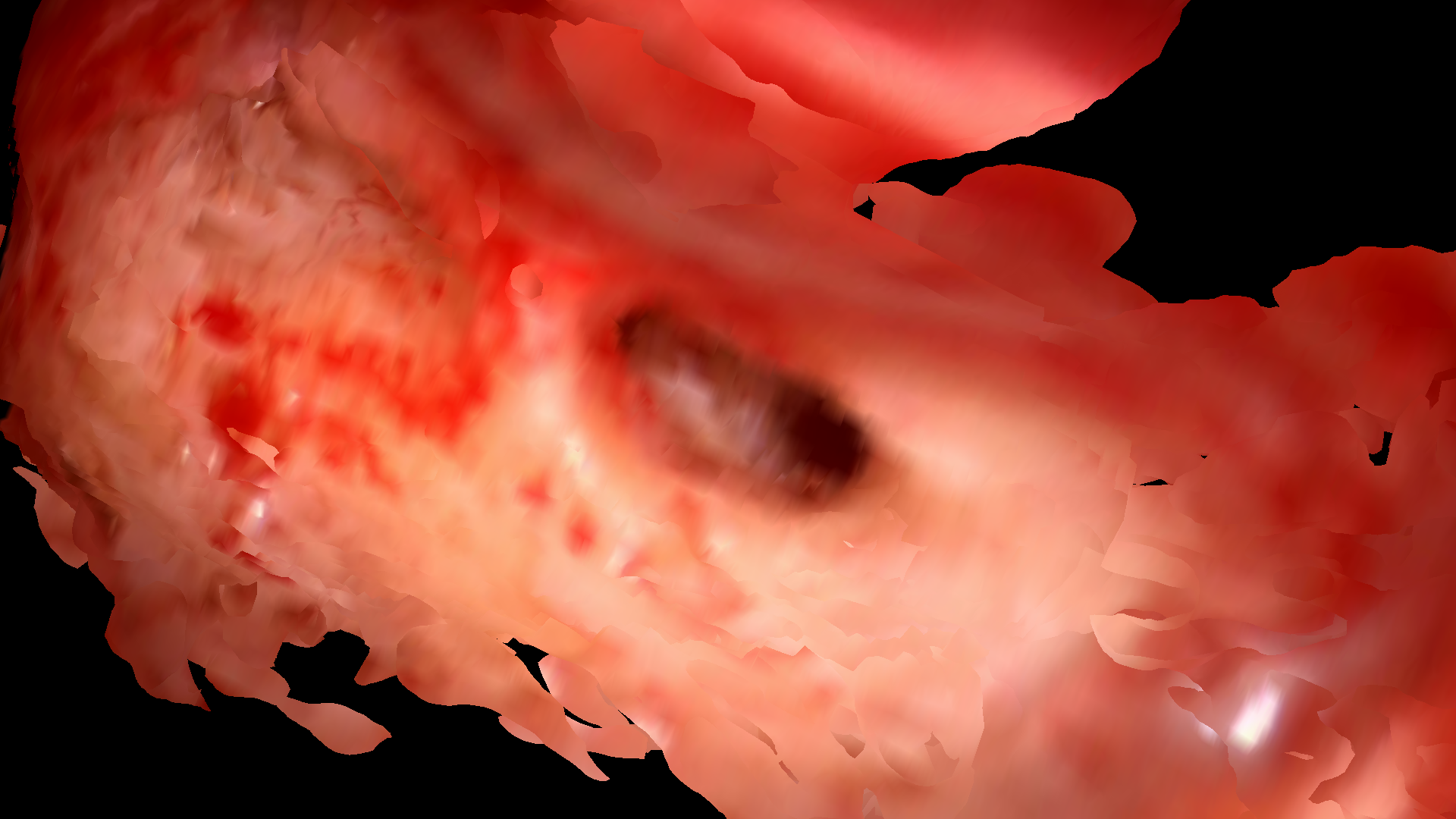}
    \end{minipage}
    \hfill
    \begin{minipage}{0.11\textwidth}
        \includegraphics[width=\textwidth]{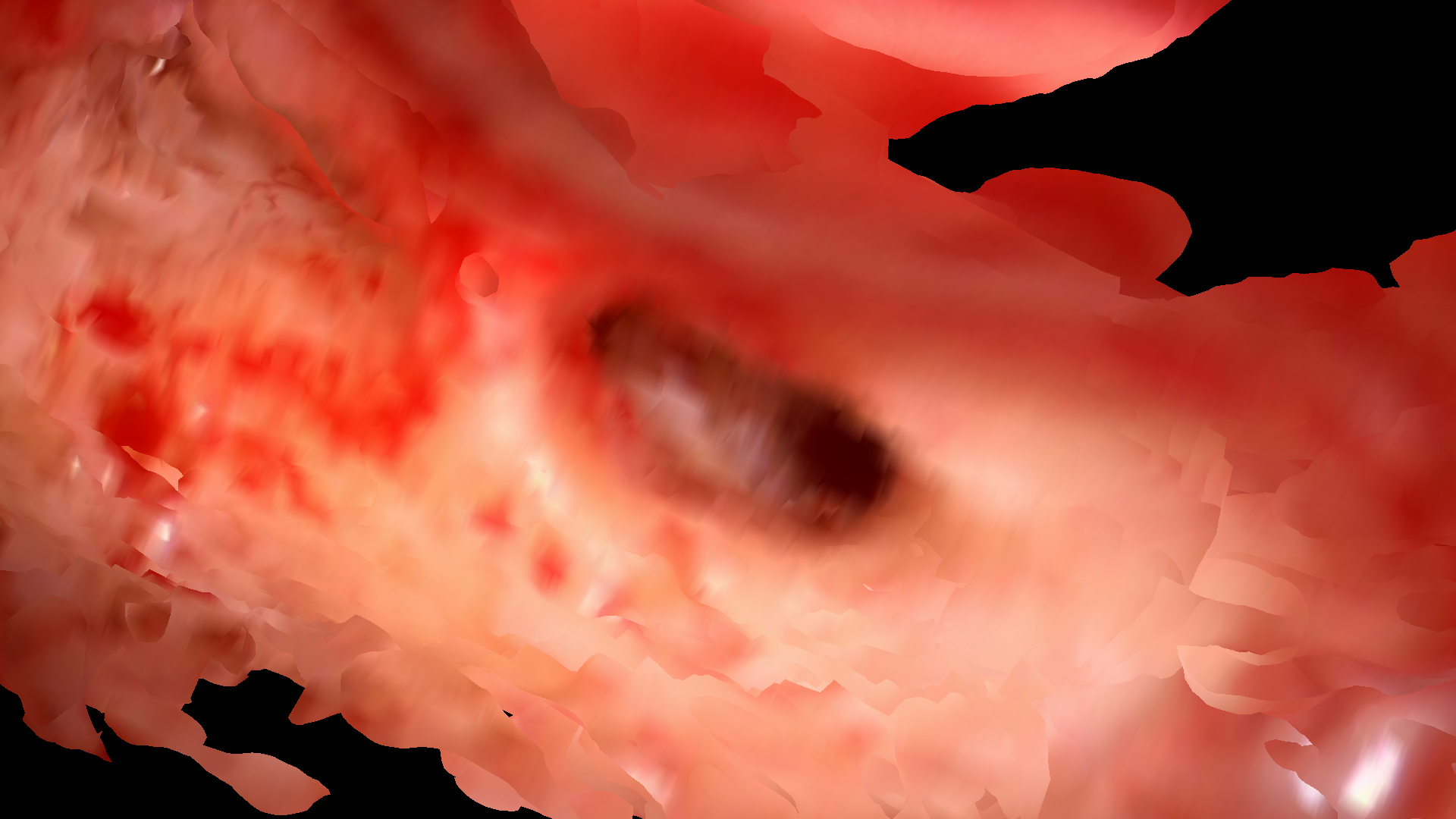}
    \end{minipage}
    \hfill
    \begin{minipage}{0.11\textwidth}
        \includegraphics[width=\textwidth]{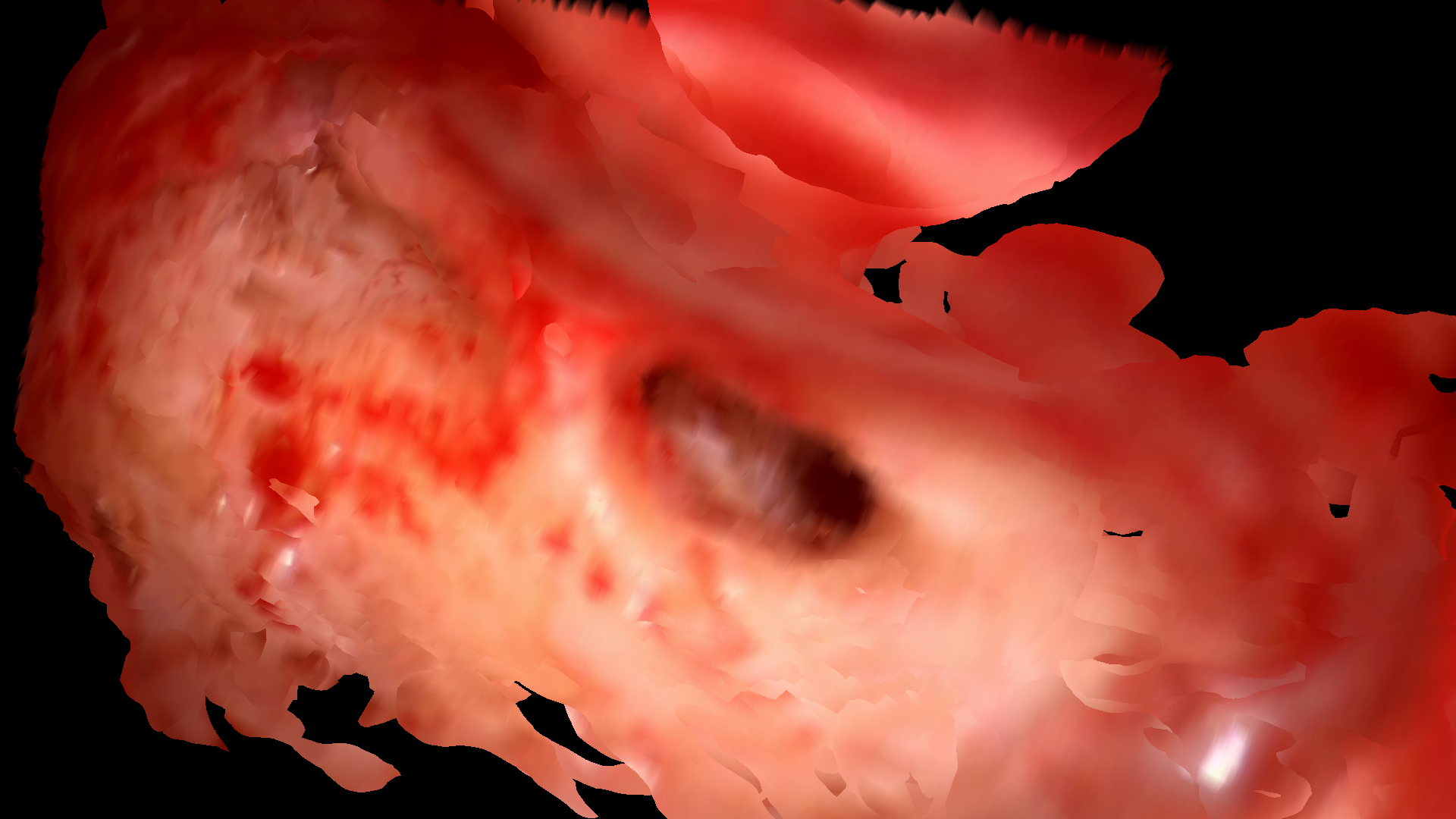}
    \end{minipage}
    \begin{minipage}{0.11\textwidth}
        \includegraphics[width=\textwidth]{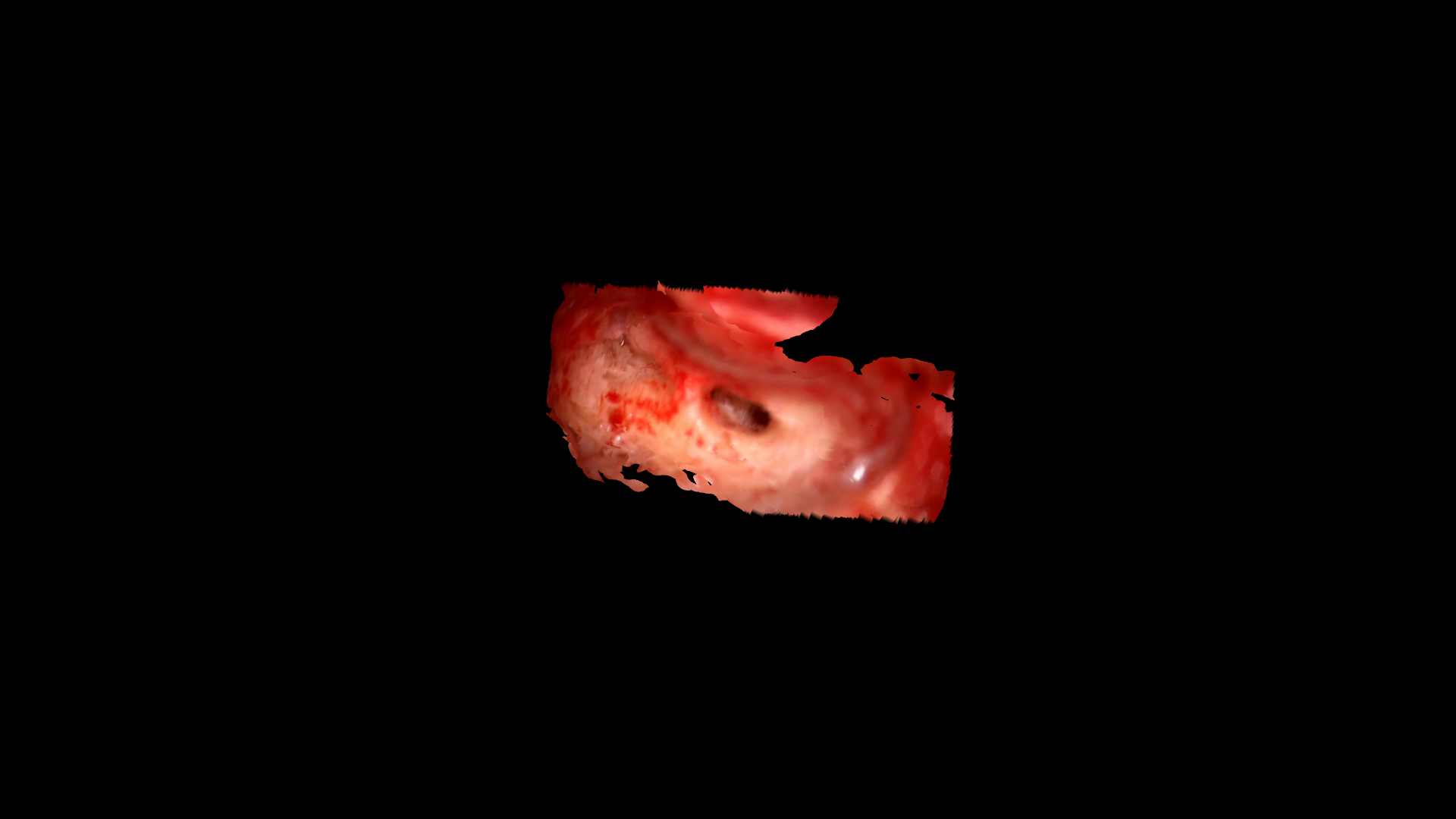}
    \end{minipage}
    \hfill
    \begin{minipage}{0.11\textwidth}
        \includegraphics[width=\textwidth]{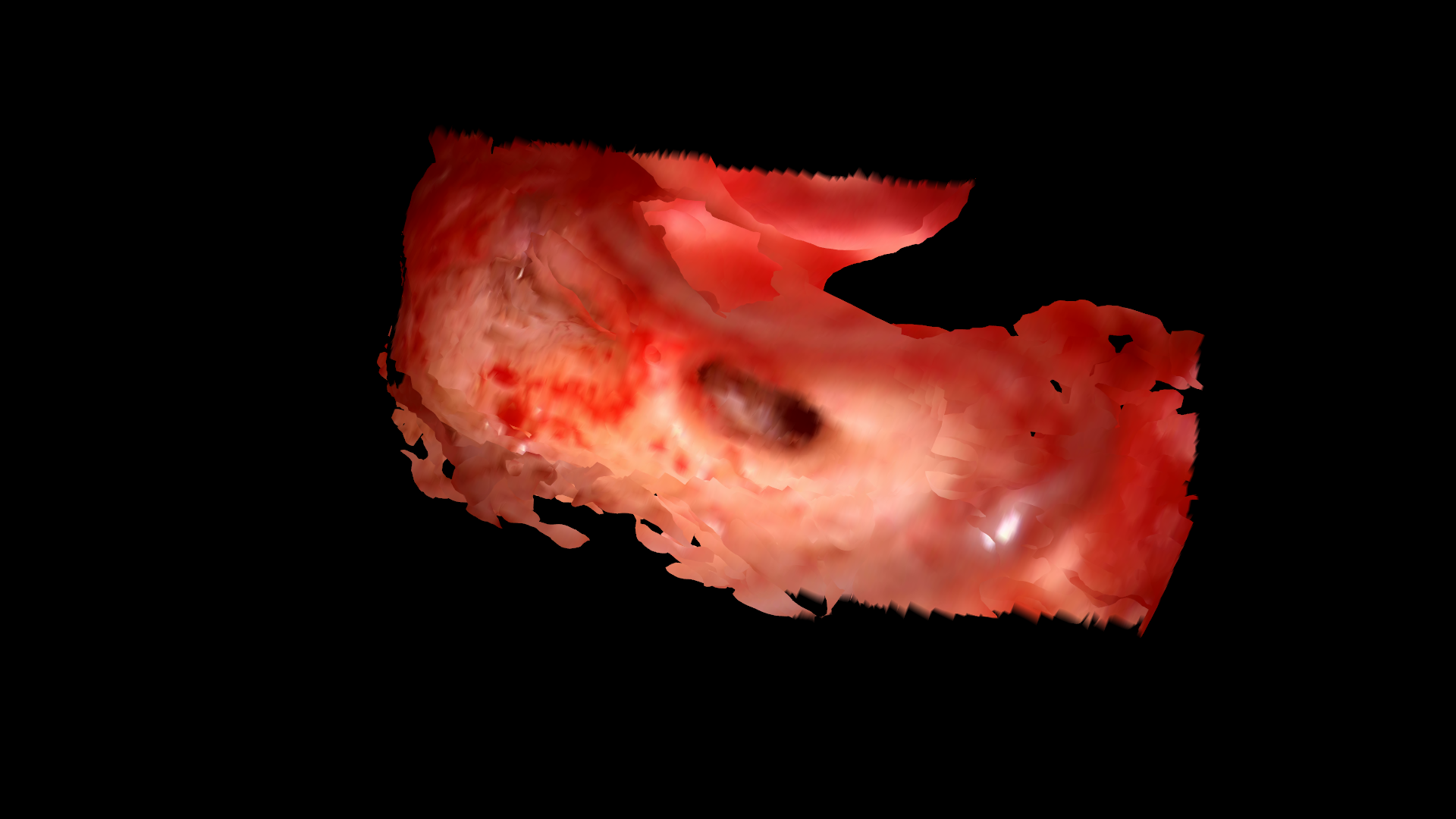}
    \end{minipage}
    \hfill
    \begin{minipage}{0.11\textwidth}
        \includegraphics[width=\textwidth]{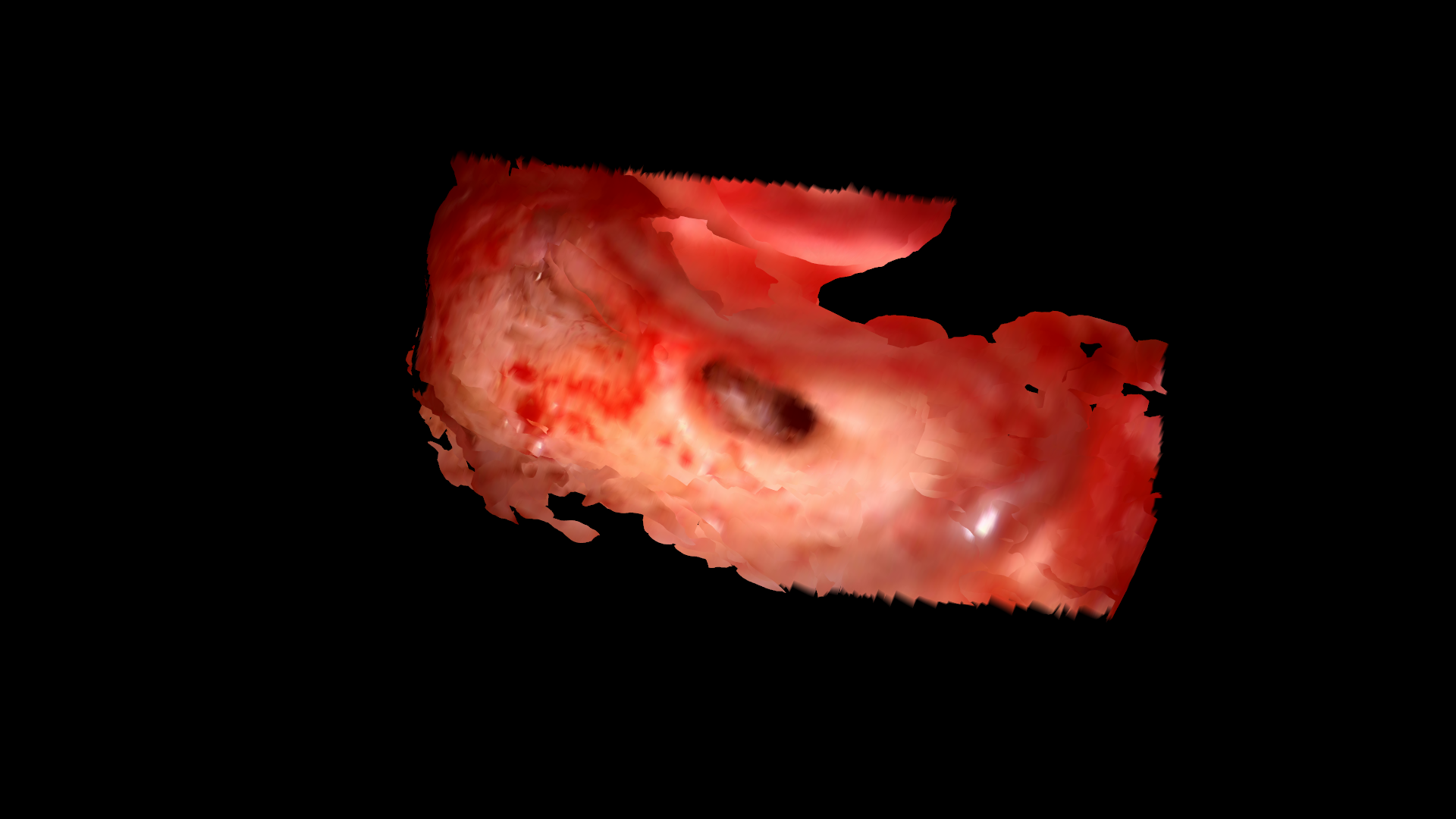}
    \end{minipage}
    \hfill
    \begin{minipage}{0.11\textwidth}
        \includegraphics[width=\textwidth]{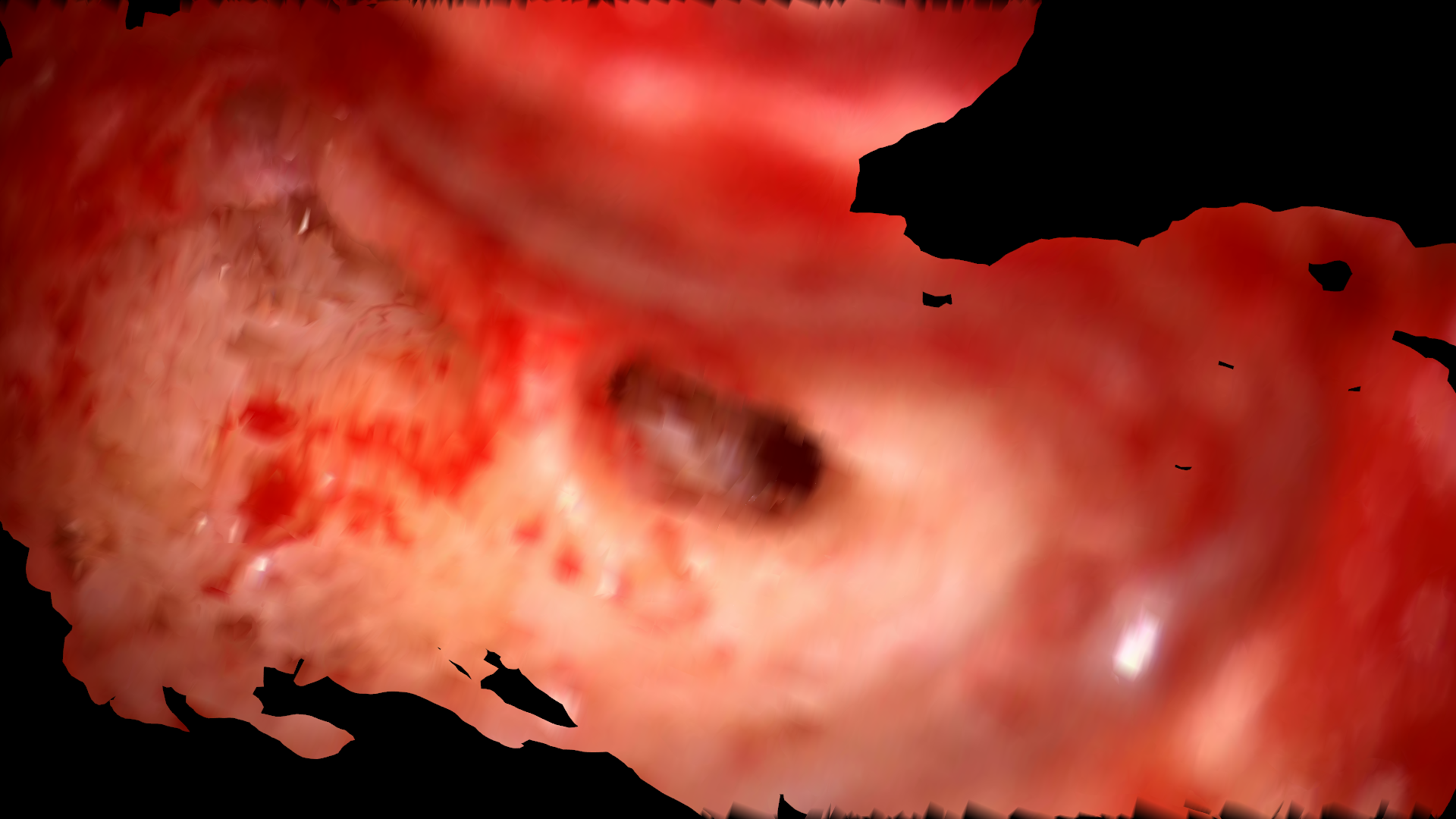}
    \end{minipage}
    \hfill
    \begin{minipage}{0.11\textwidth}
        \includegraphics[width=\textwidth]{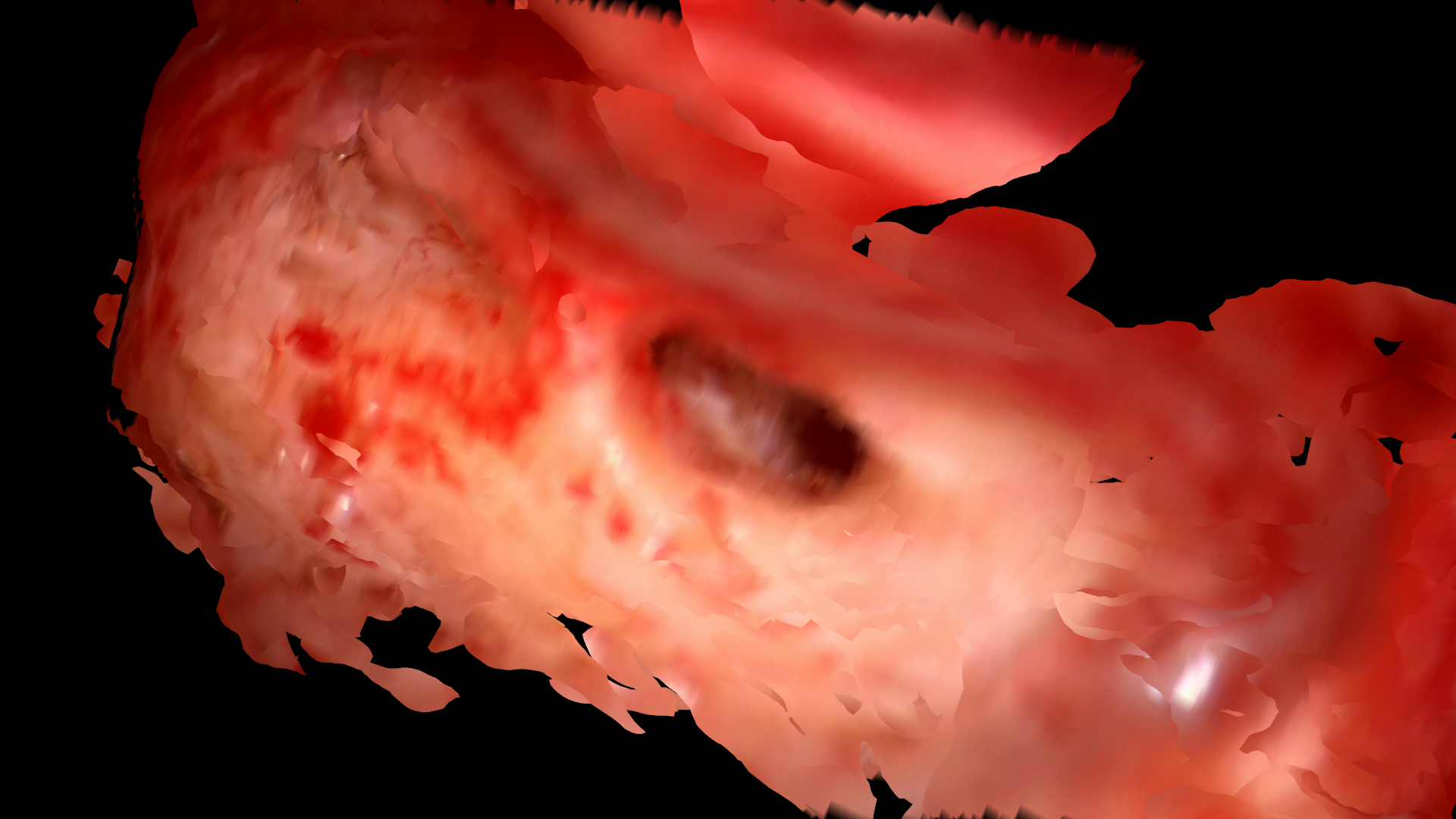}
    \end{minipage}
    \hfill
    \begin{minipage}{0.11\textwidth}
        \includegraphics[width=\textwidth]{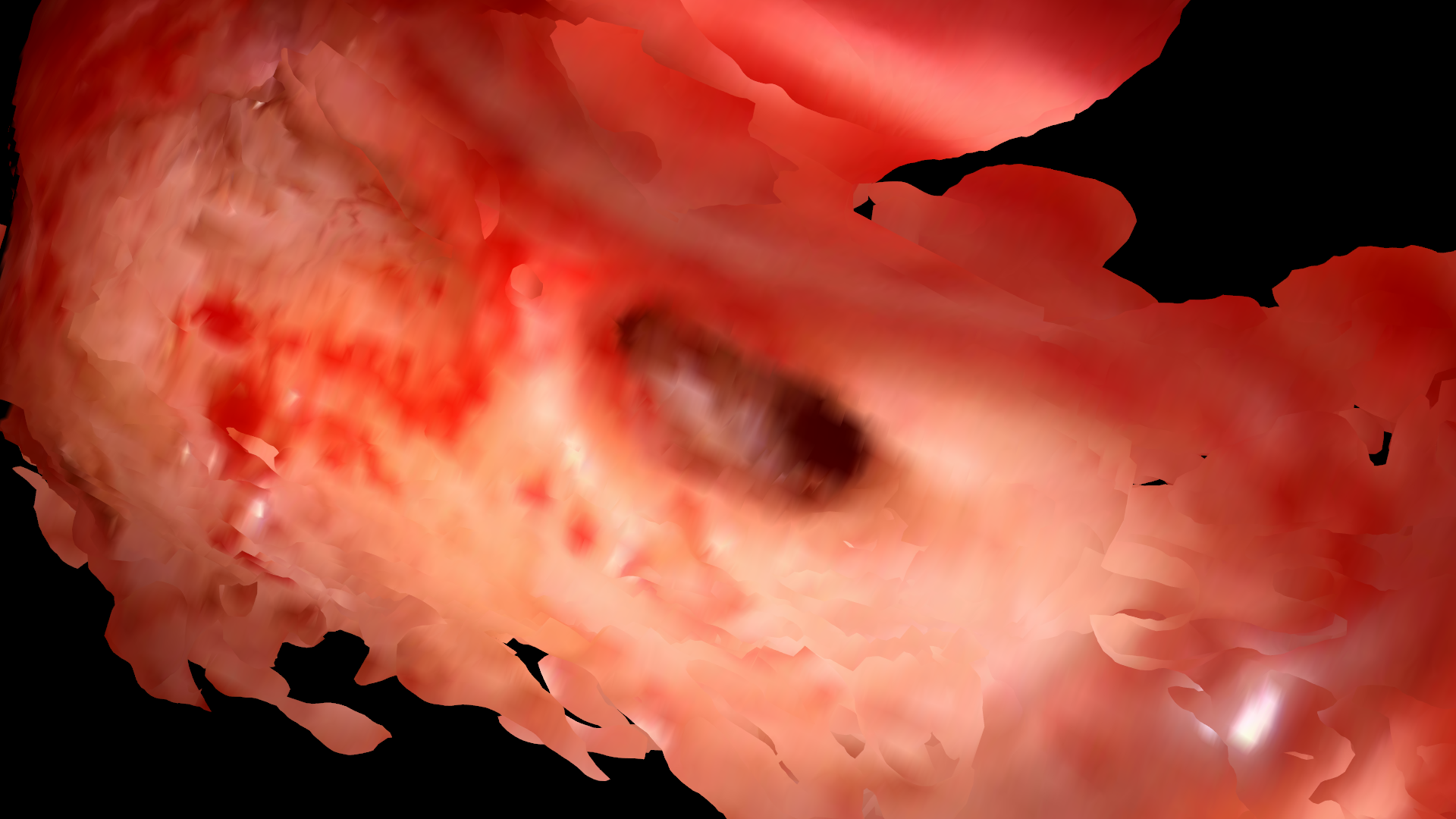}
    \end{minipage}
    \hfill
    \begin{minipage}{0.11\textwidth}
        \includegraphics[width=\textwidth]{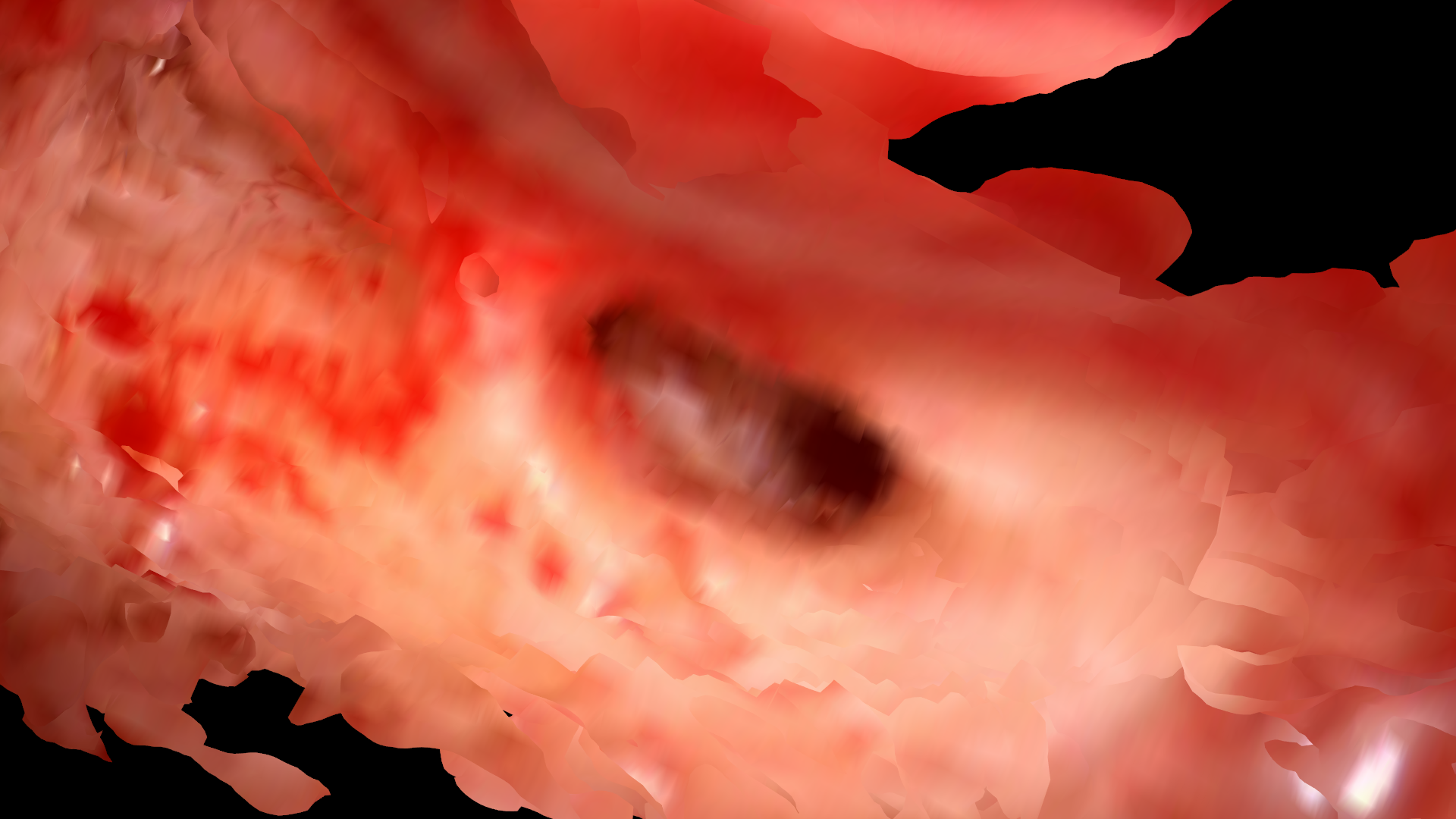}
    \end{minipage}
    \hfill
    \begin{minipage}{0.11\textwidth}
        \includegraphics[width=\textwidth]{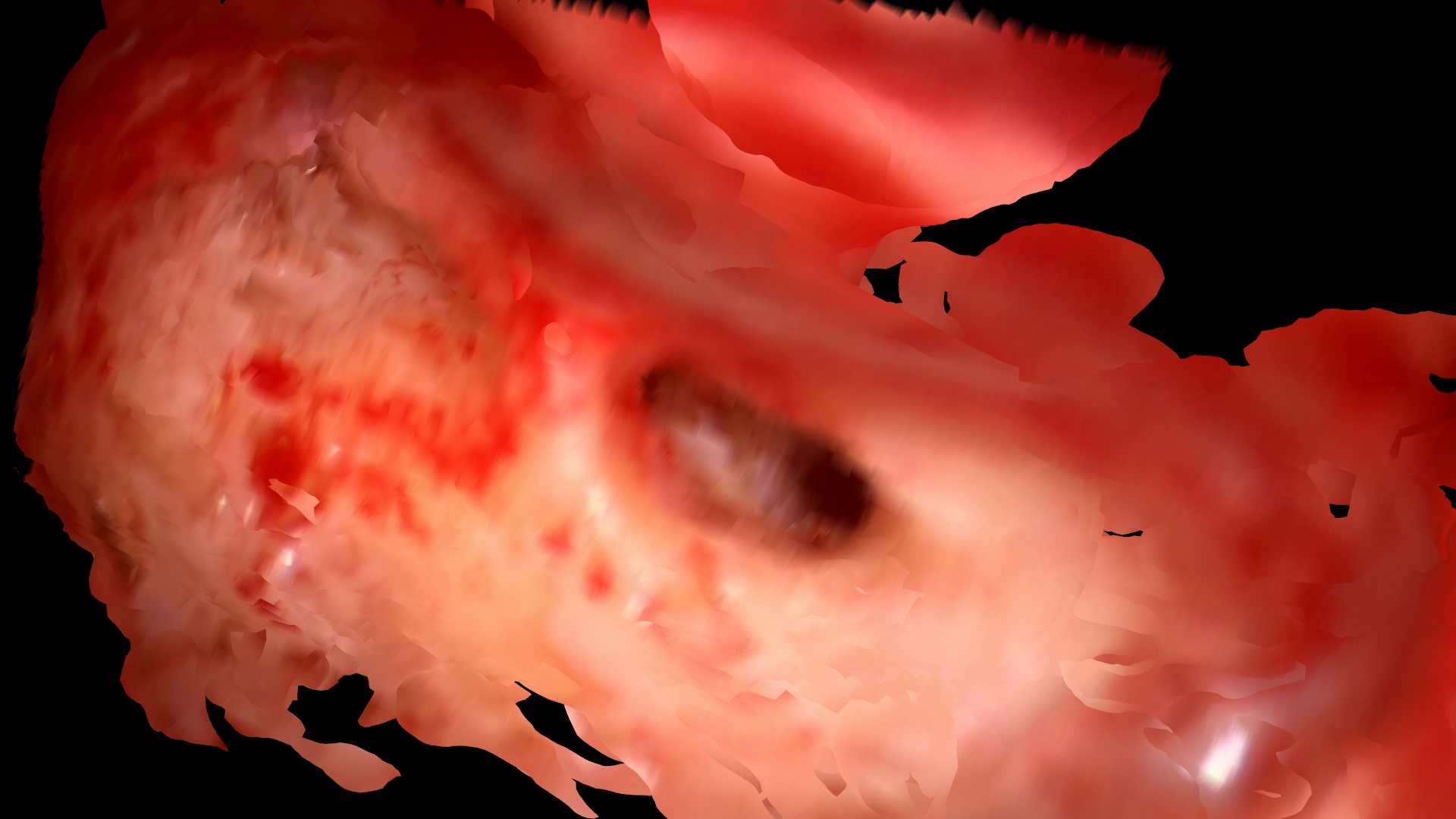}
    \end{minipage}
\caption{Real surgical image (1st row), segmented real surgical image (2nd row), and synthetic views rendered by PyTorch3D (3rd row) and PyVista (4th row). Each column represents an individual test case.}
\label{fig:qualitative_results} 
\end{figure}

Fig. \ref{fig:qualitative_results} shows the performance on eight test samples, with each column representing an individual test case. The first row displays the real surgical image, the second row shows the segmented real surgical image containing only the region that overlaps with the registered surface mesh, the third row demonstrates the synthesized view rendered by Pytorch3D, and the last row indicates the rendered scene generated by PyVista.
For evaluating the synthesized results with two different renders, we manually annotate 30 ground truth poses in Vision6D, and compare the real images with the synthesized ones in Table \ref{Tab:analysis} using Fréchet Inception Distance (FID) \cite{fid} , Kernel Inception Distance (KID) \cite{kid} , and Structural Similarity Index Measure (SSIM) \cite{ssim} metrics with their mean values. The overall results show that both rendering methods perform similarly, effectively generating realistic renderings that closely match the original surgical scenes, despite variations in lighting conditions, environmental noise, and occlusions in the real surgical frames from the surgeons’ hands and tools.
\begin{table*}[ht]
    \centering
    \footnotesize
    
    \begin{tabularx}{\textwidth}{ |c|>{\centering\arraybackslash}X|>{\centering\arraybackslash}X|>{\centering\arraybackslash}X| }
        \hline
        \multicolumn{1}{|c|}{Methods} &
        \multicolumn{1}{c|}{FID $\downarrow$} & 
        \multicolumn{1}{c|}{KID $\downarrow$} & 
        \multicolumn{1}{c|}{SSIM $\uparrow$} \\
        \hline
        Synthetic views generated by Pytorch3D & 180.2680 & 0.3989 & 0.8607 \\
        \hline
        Synthetic views generated by PyVista & 179.9417 & 0.3979 & 0.8595 \\
        \hline
    \end{tabularx}
    \caption{Performance comparison between two different renders for realism.}
    \label{Tab:analysis}
\end{table*}
Fig. \ref{fig:pytorch3d_pyvista_render} shows a gallery of synthesized novel microscopy views in pairs through various randomly generated poses of the post-mastoidectomy mesh.

\renewcommand{\thesubfigure}{\Roman{subfigure}}

\newcommand\pairsize{0.48}
\newcommand\imgsize{0.49}
\newcommand\vspacing{0.5em}

\begin{figure}[ht]
\begin{minipage}{\pairsize\textwidth}
    \centering
    \begin{minipage}{\imgsize\linewidth}
        \centering
        \includegraphics[width=\linewidth]{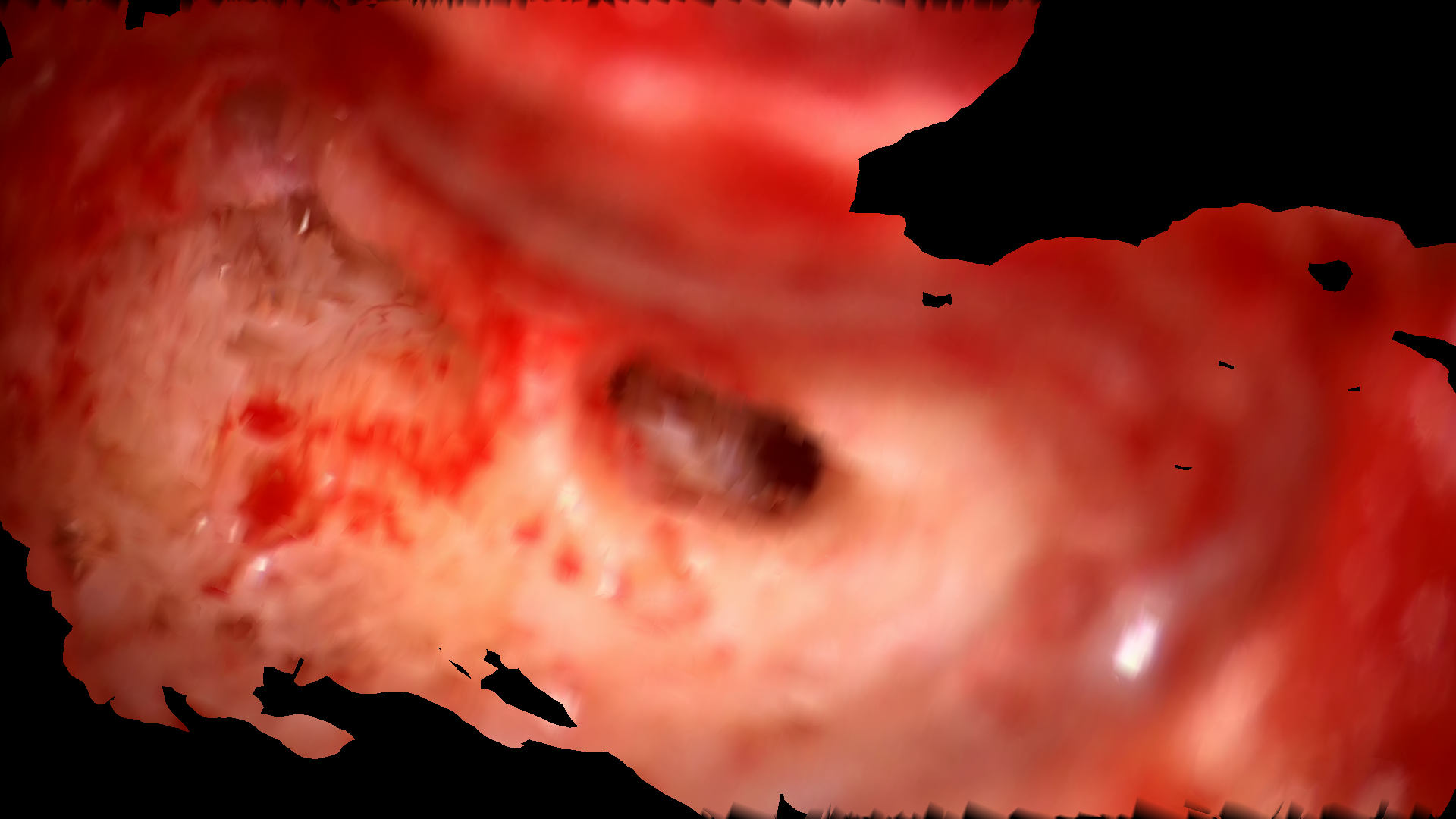}
        \subcaption{Pose 1 - Pytorch3D}
    \end{minipage}\hfill
    \begin{minipage}{\imgsize\linewidth}
        \centering
        \includegraphics[width=\linewidth]{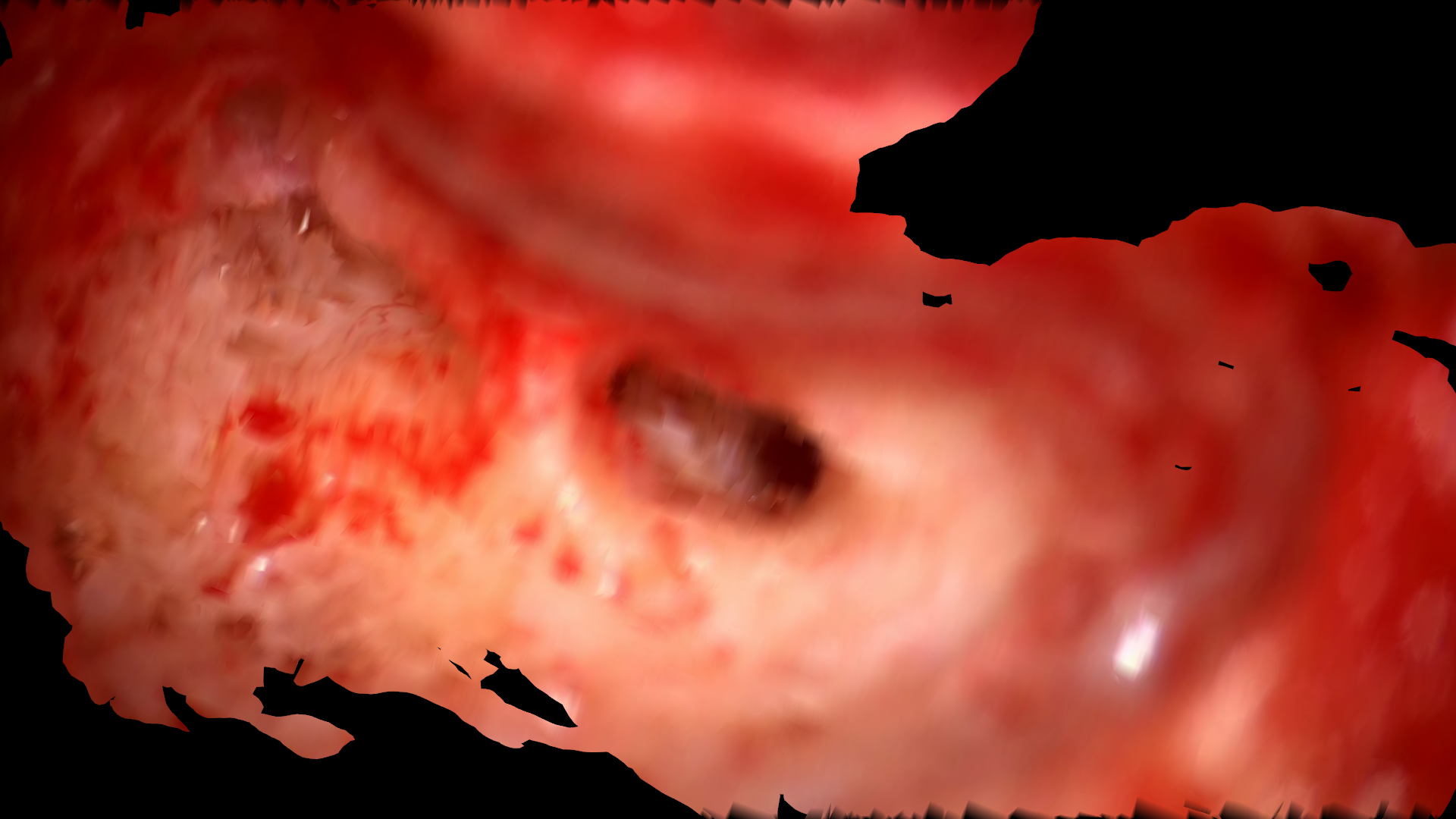}
        \subcaption{Pose 1 - PyVista}
    \end{minipage}
\end{minipage}\hfill
\begin{minipage}{\pairsize\textwidth}
    \centering
    \begin{minipage}{\imgsize\linewidth}
        \centering
        \includegraphics[width=\linewidth]{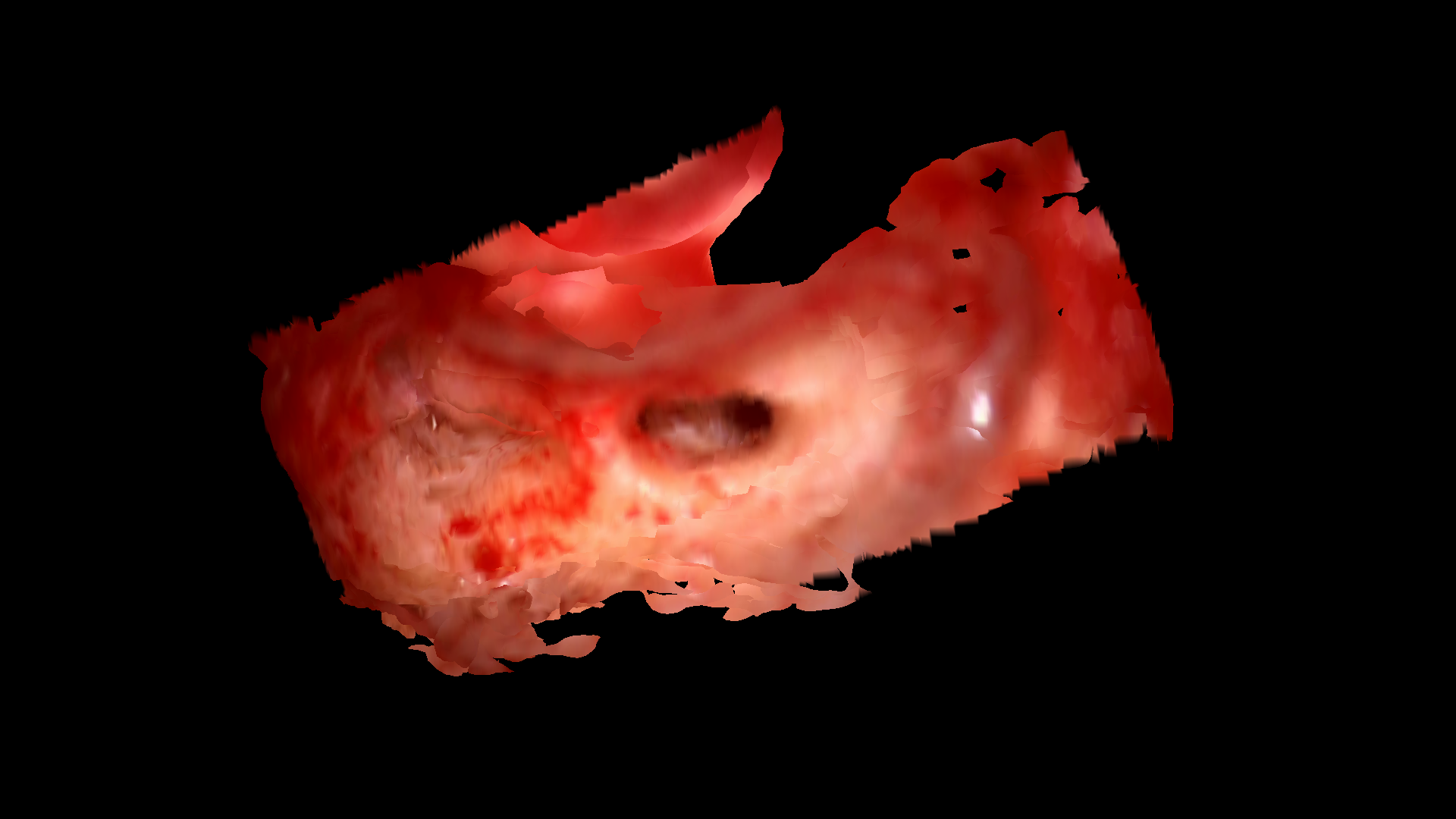}
        \subcaption{Pose 2 - Pytorch3D}
    \end{minipage}\hfill
    \begin{minipage}{\imgsize\linewidth}
        \centering
        \includegraphics[width=\linewidth]{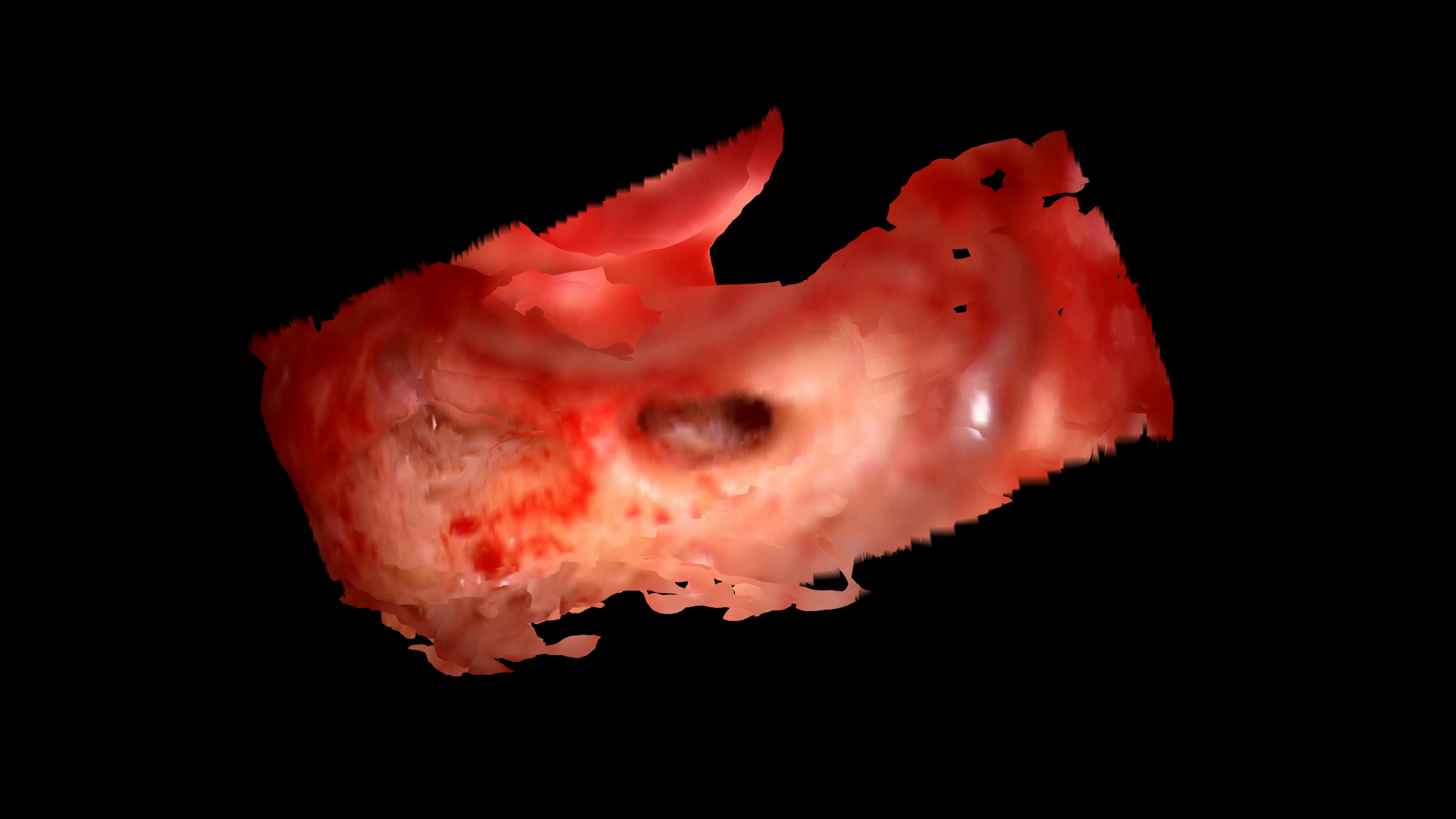}
        \subcaption{Pose 2 - PyVista}
    \end{minipage}
\end{minipage}

\vspace{\vspacing}
\begin{minipage}{\pairsize\textwidth}
    \centering
    \begin{minipage}{\imgsize\linewidth}
        \centering
        \includegraphics[width=\linewidth]{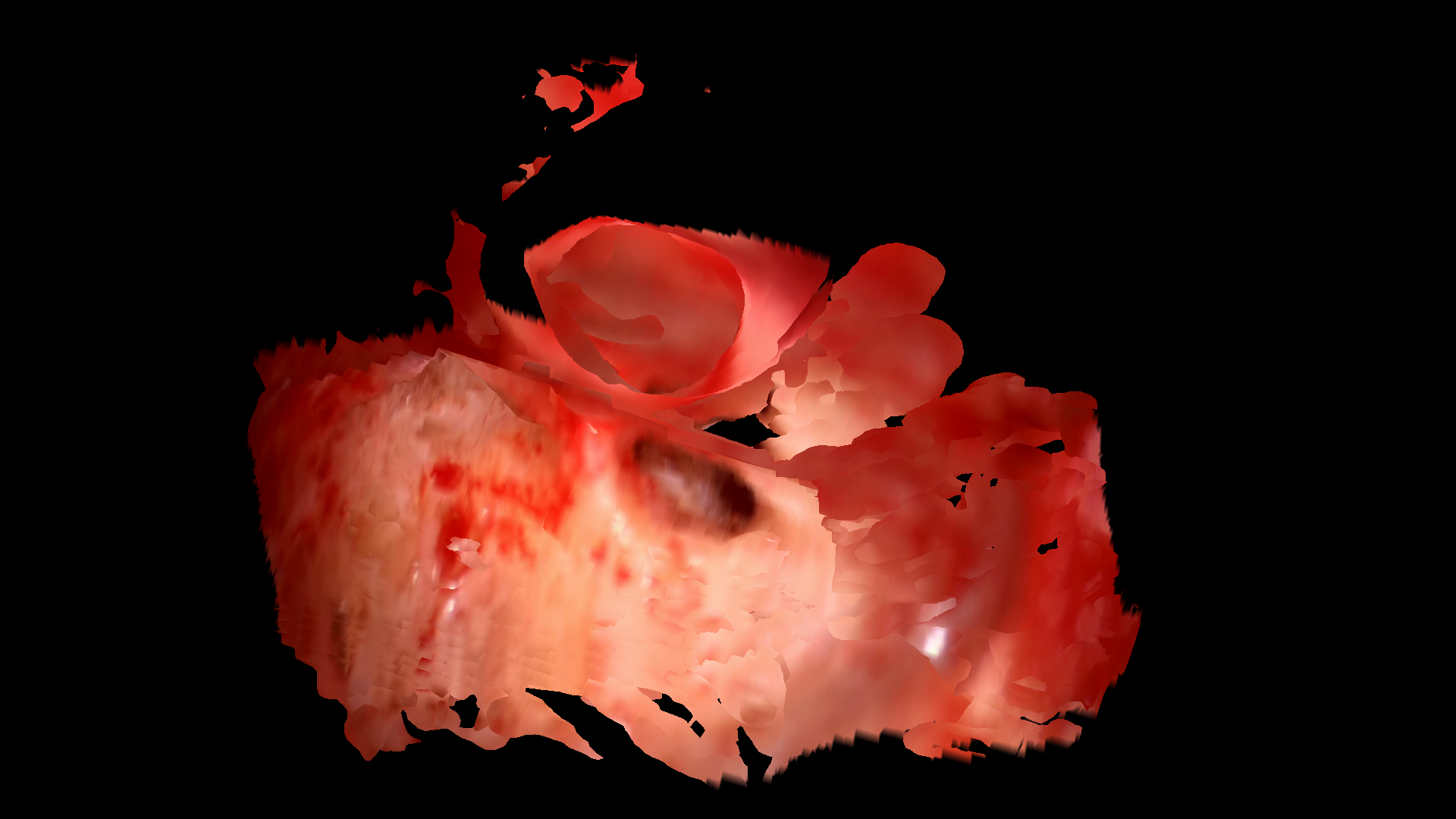}
        \subcaption{Pose 3 - Pytorch3D}
    \end{minipage}\hfill
    \begin{minipage}{\imgsize\linewidth}
        \centering
        \includegraphics[width=\linewidth]{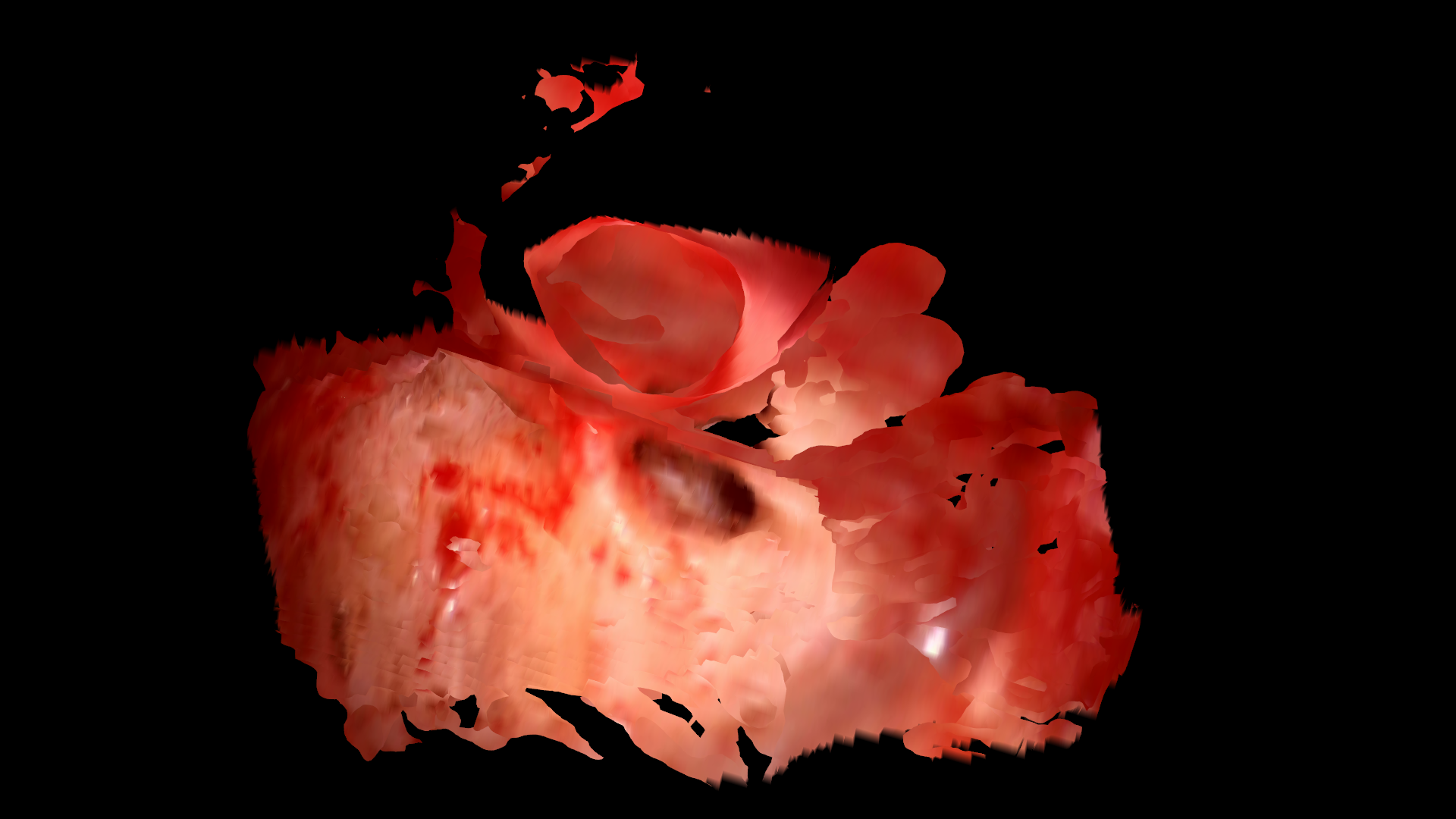}
        \subcaption{Pose 3 - PyVista}
    \end{minipage}
\end{minipage}\hfill
\begin{minipage}{\pairsize\textwidth}
    \centering
    \begin{minipage}{\imgsize\linewidth}
        \centering
        \includegraphics[width=\linewidth]{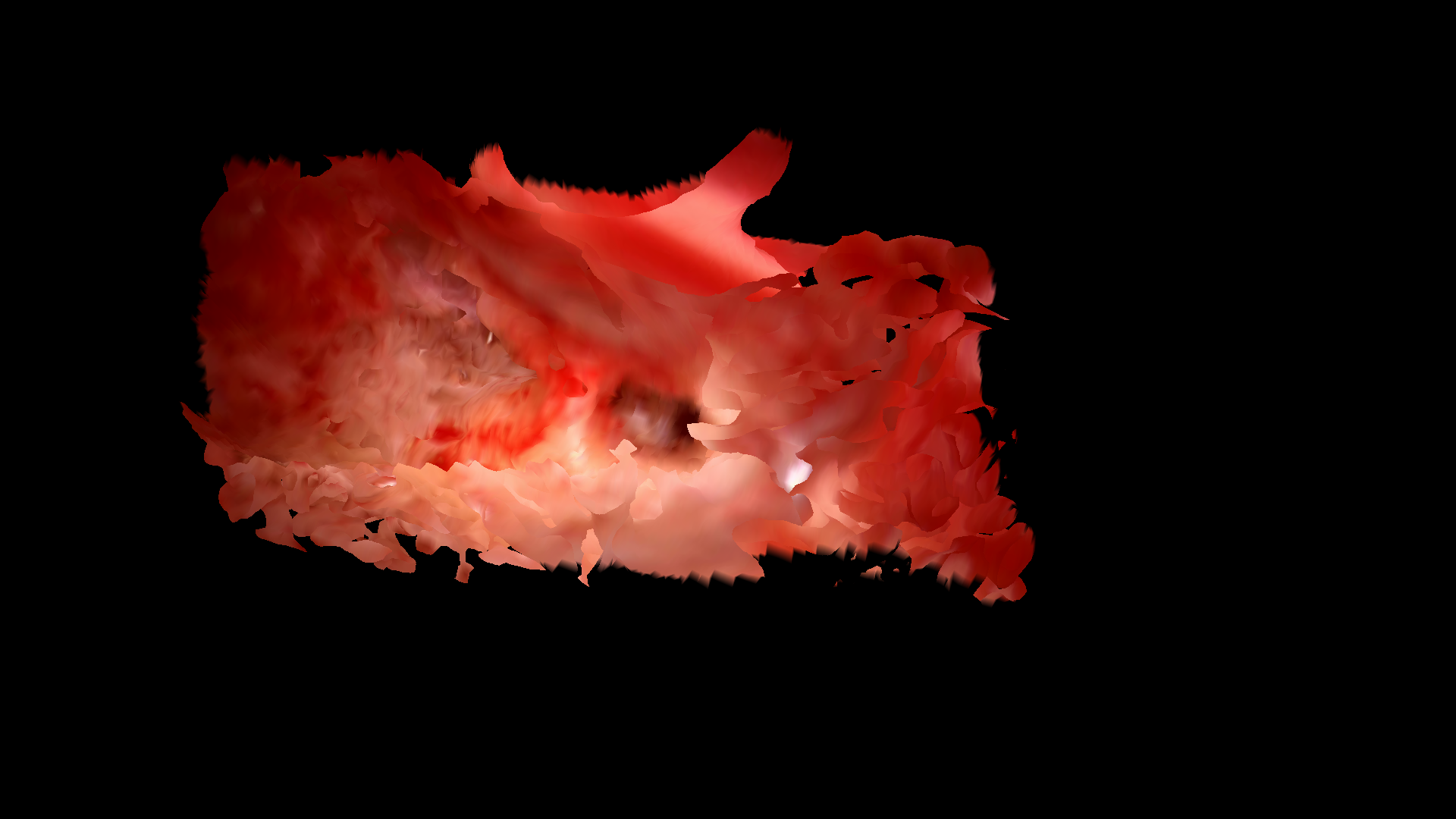}
        \subcaption{Pose 4 - Pytorch3D}
    \end{minipage}\hfill
    \begin{minipage}{\imgsize\linewidth}
        \centering
        \includegraphics[width=\linewidth]{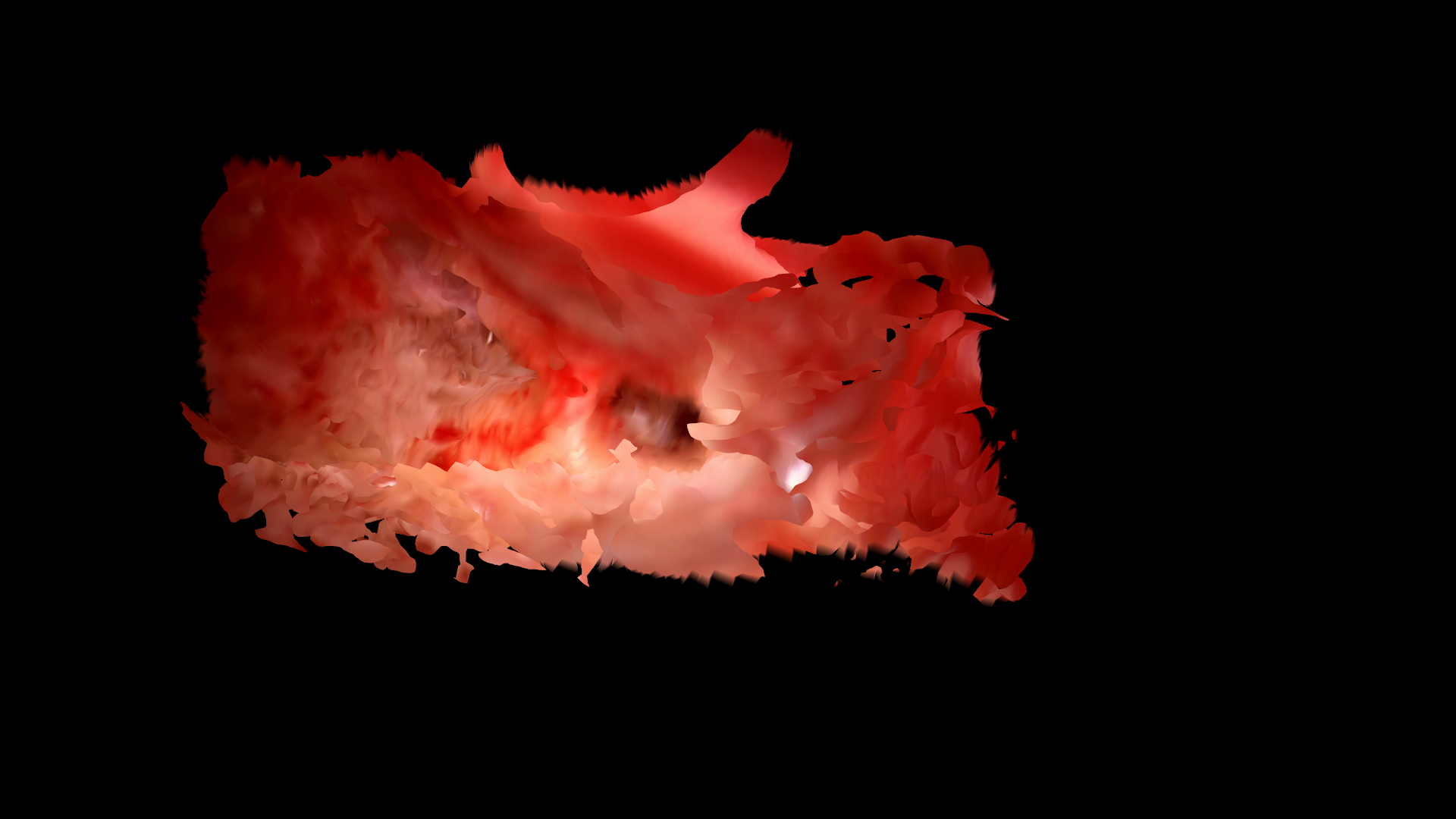}
        \subcaption{Pose 4 - PyVista}
    \end{minipage}
\end{minipage}

\vspace{\vspacing}
\begin{minipage}{\pairsize\textwidth}
    \centering
    \begin{minipage}{\imgsize\linewidth}
        \centering
        \includegraphics[width=\linewidth]{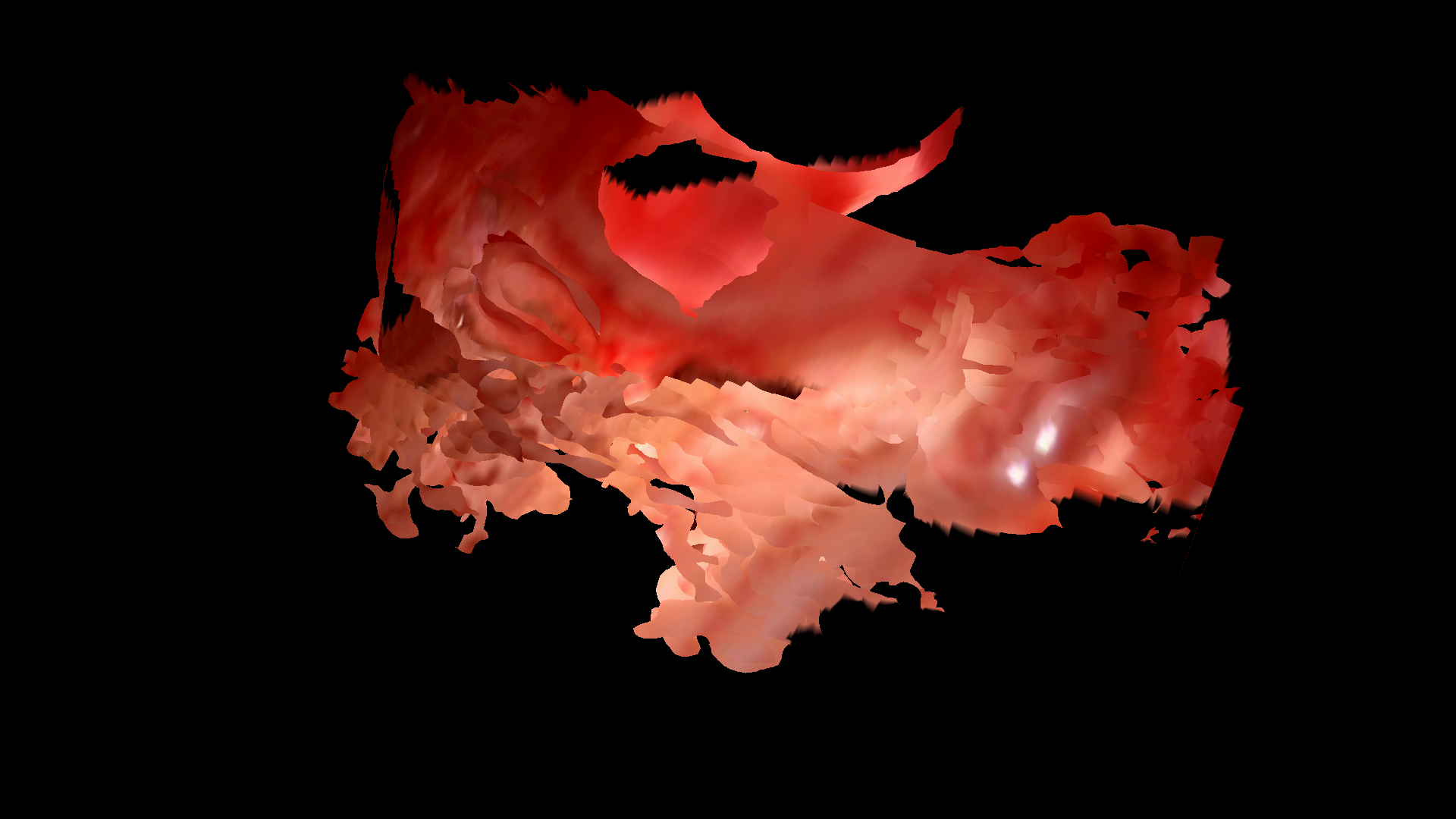}
        \subcaption{Pose 5 - Pytorch3D}
    \end{minipage}\hfill
    \begin{minipage}{\imgsize\linewidth}
        \centering
        \includegraphics[width=\linewidth]{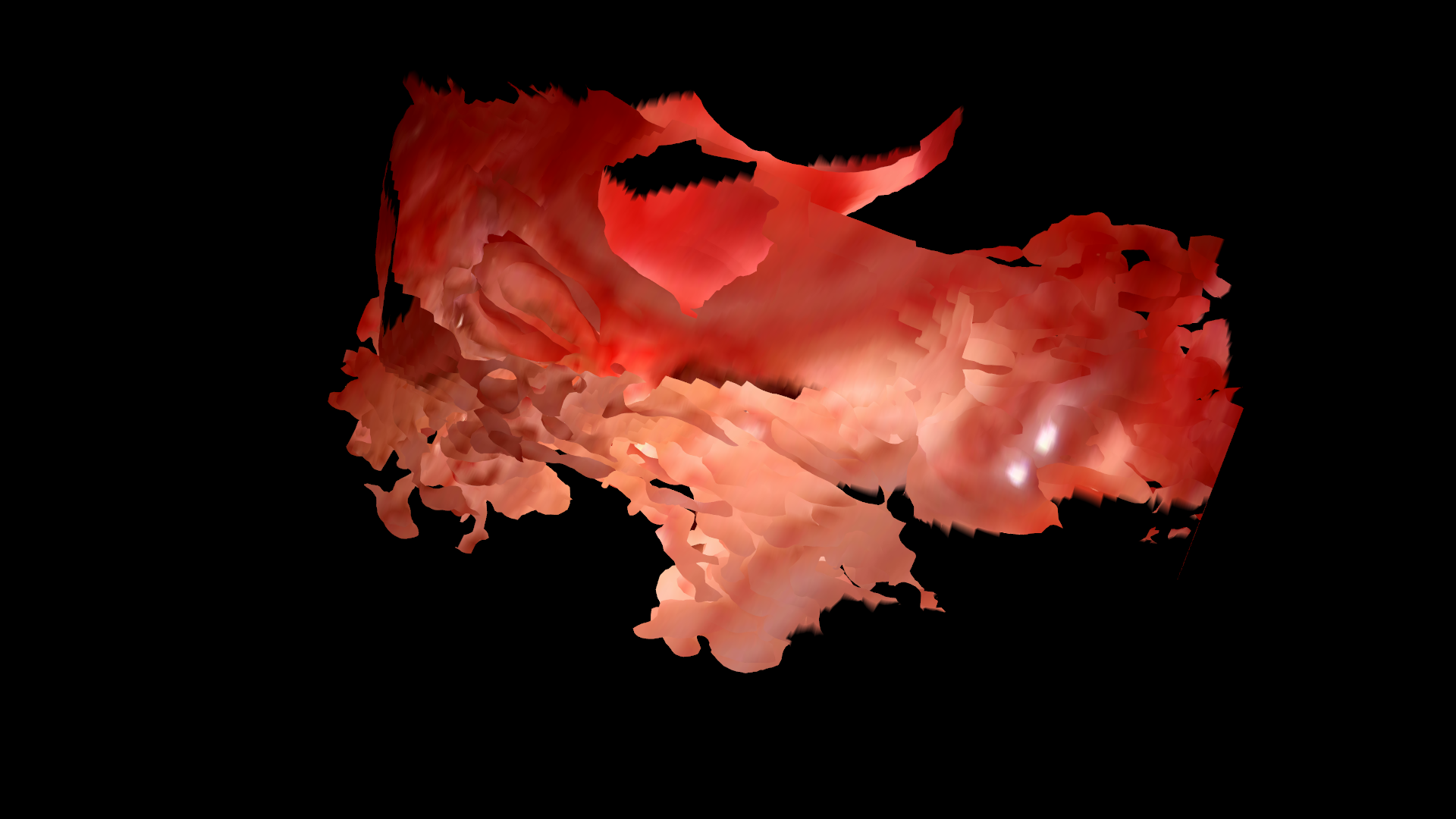}
        \subcaption{Pose 5 - PyVista}
    \end{minipage}
\end{minipage}\hfill
\begin{minipage}{\pairsize\textwidth}
    \centering
    \begin{minipage}{\imgsize\linewidth}
        \centering
        \includegraphics[width=\linewidth]{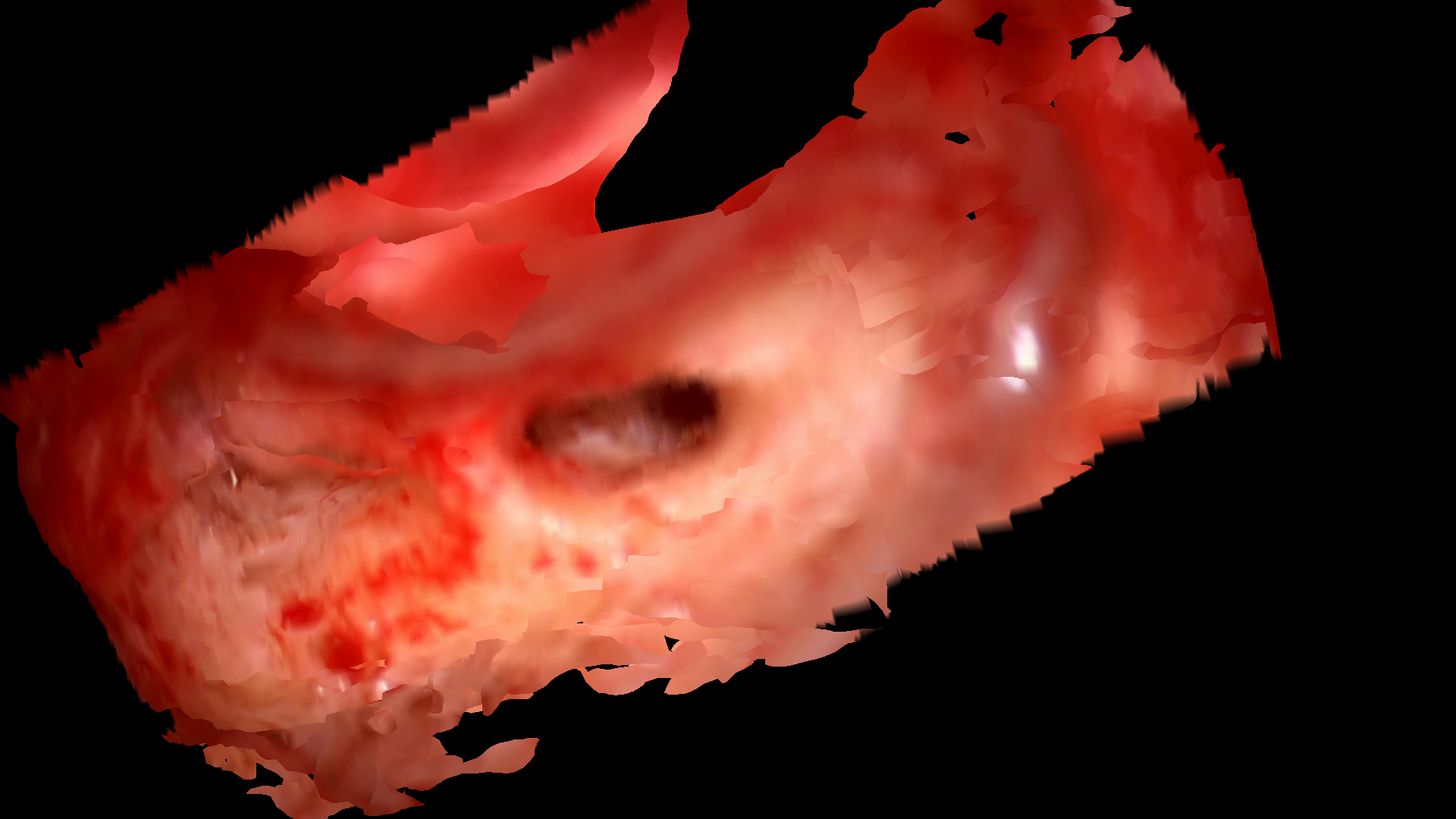}
        \subcaption{Pose 6 - Pytorch3D}
    \end{minipage}\hfill
    \begin{minipage}{\imgsize\linewidth}
        \centering
        \includegraphics[width=\linewidth]{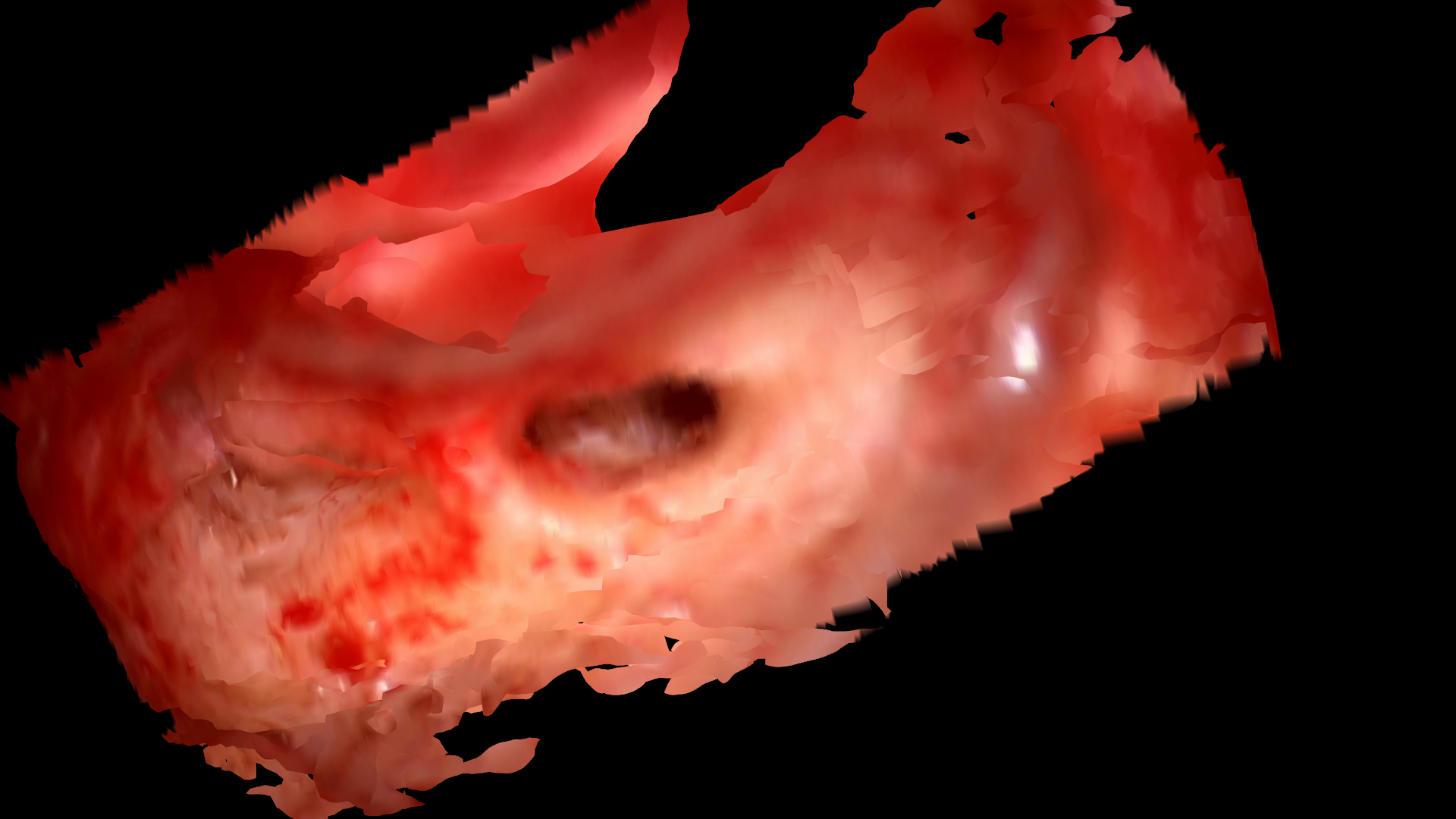}
        \subcaption{Pose 6 - PyVista}
    \end{minipage}
\end{minipage}

\vspace{\vspacing}
\begin{minipage}{\pairsize\textwidth}
    \centering
    \begin{minipage}{\imgsize\linewidth}
        \centering
        \includegraphics[width=\linewidth]{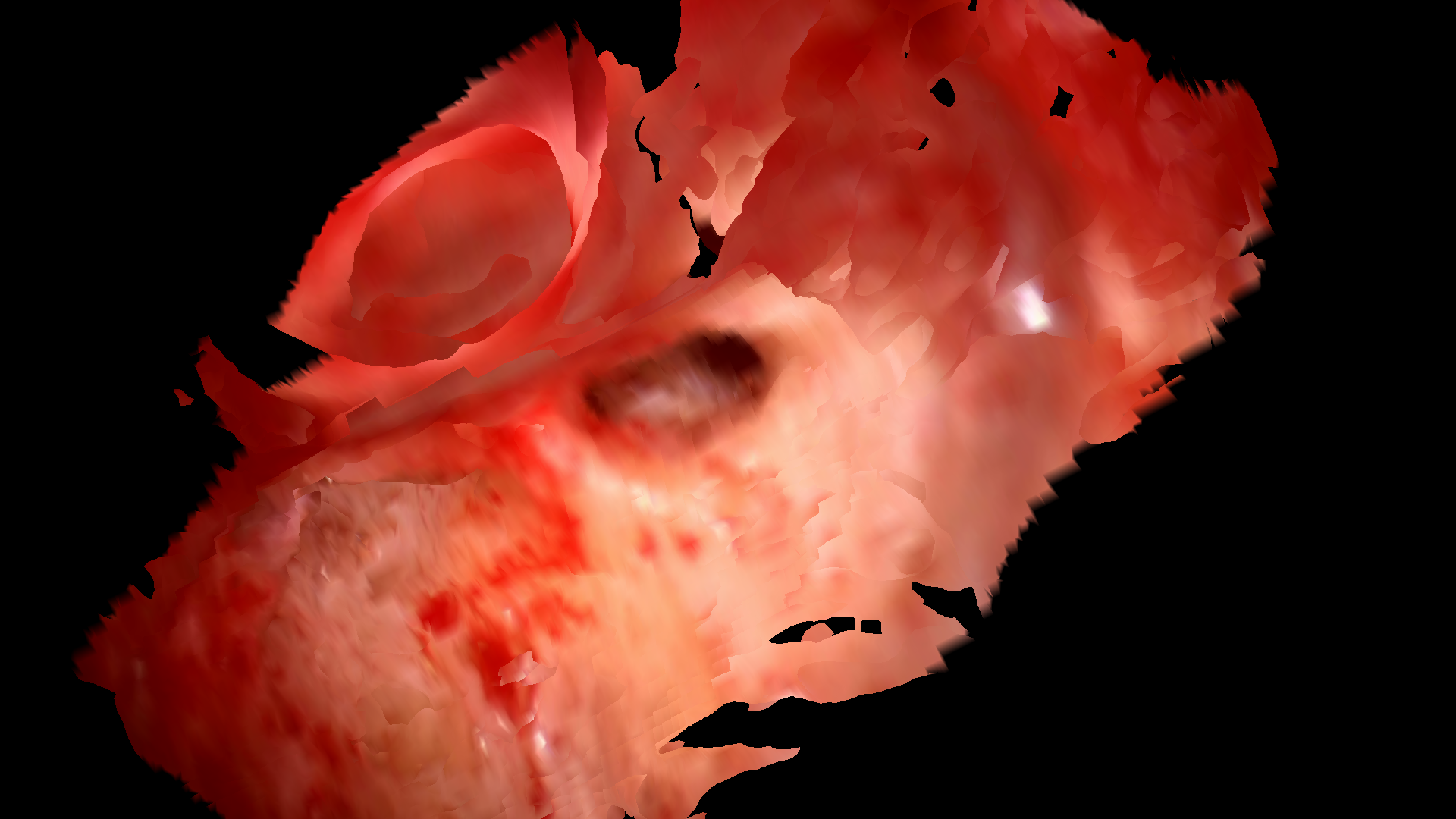}
        \subcaption{Pose 7 - Pytorch3D}
    \end{minipage}\hfill
    \begin{minipage}{\imgsize\linewidth}
        \centering
        \includegraphics[width=\linewidth]{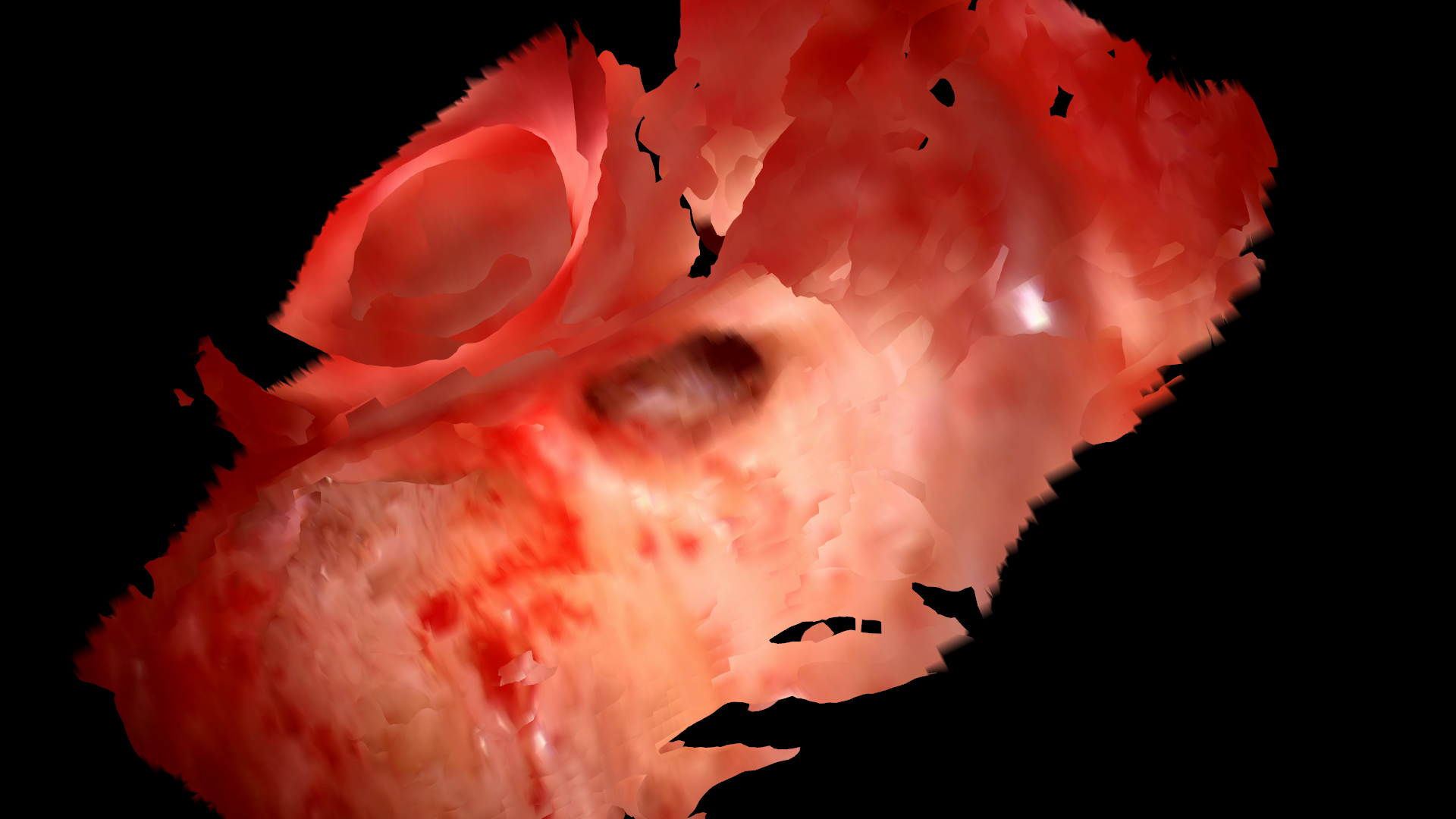}
        \subcaption{Pose 7 - PyVista}
    \end{minipage}
\end{minipage}\hfill
\begin{minipage}{\pairsize\textwidth}
    \centering
    \begin{minipage}{\imgsize\linewidth}
        \centering
        \includegraphics[width=\linewidth]{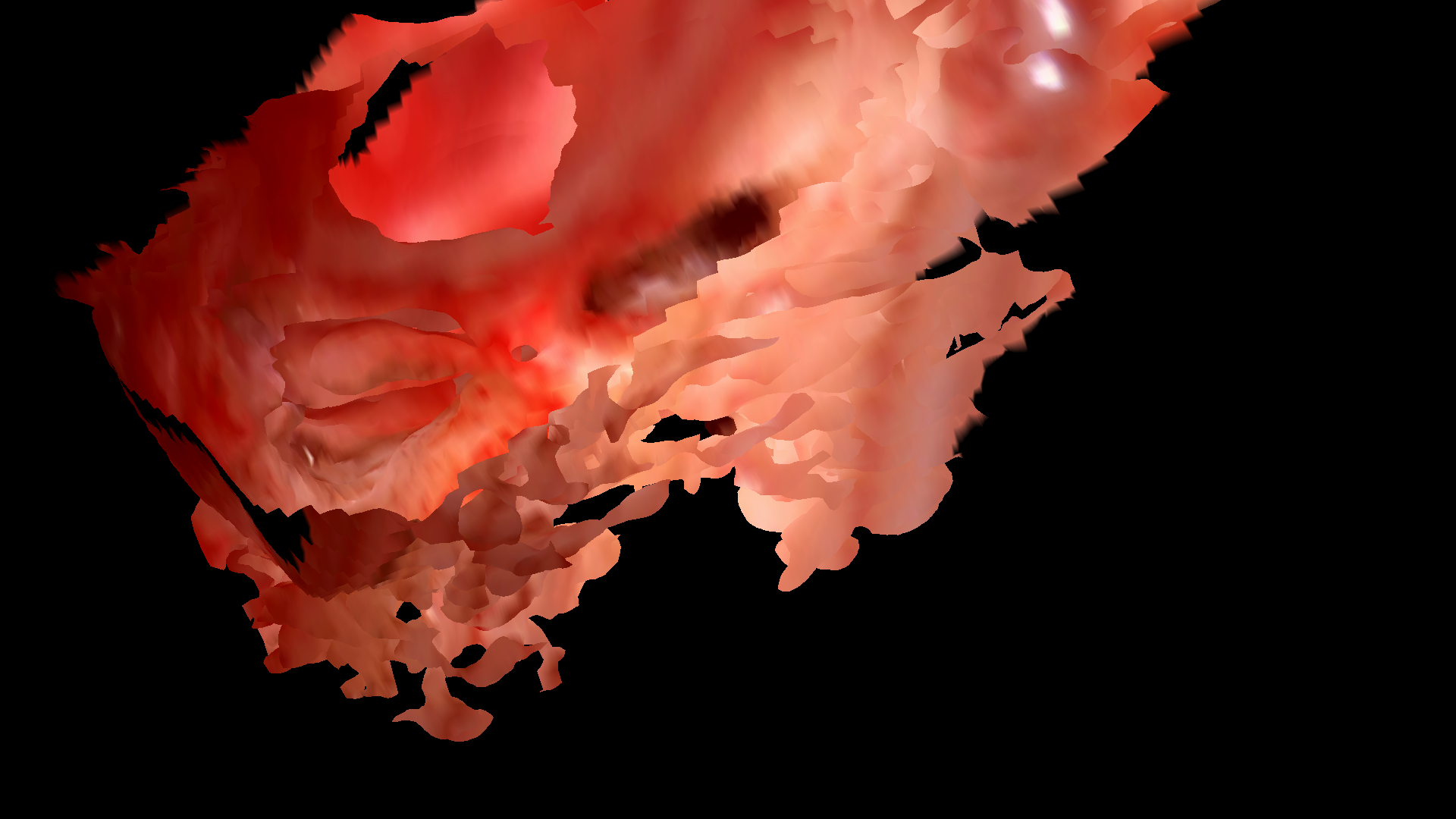}
        \subcaption{Pose 8 - Pytorch3D}
    \end{minipage}\hfill
    \begin{minipage}{\imgsize\linewidth}
        \centering
        \includegraphics[width=\linewidth]{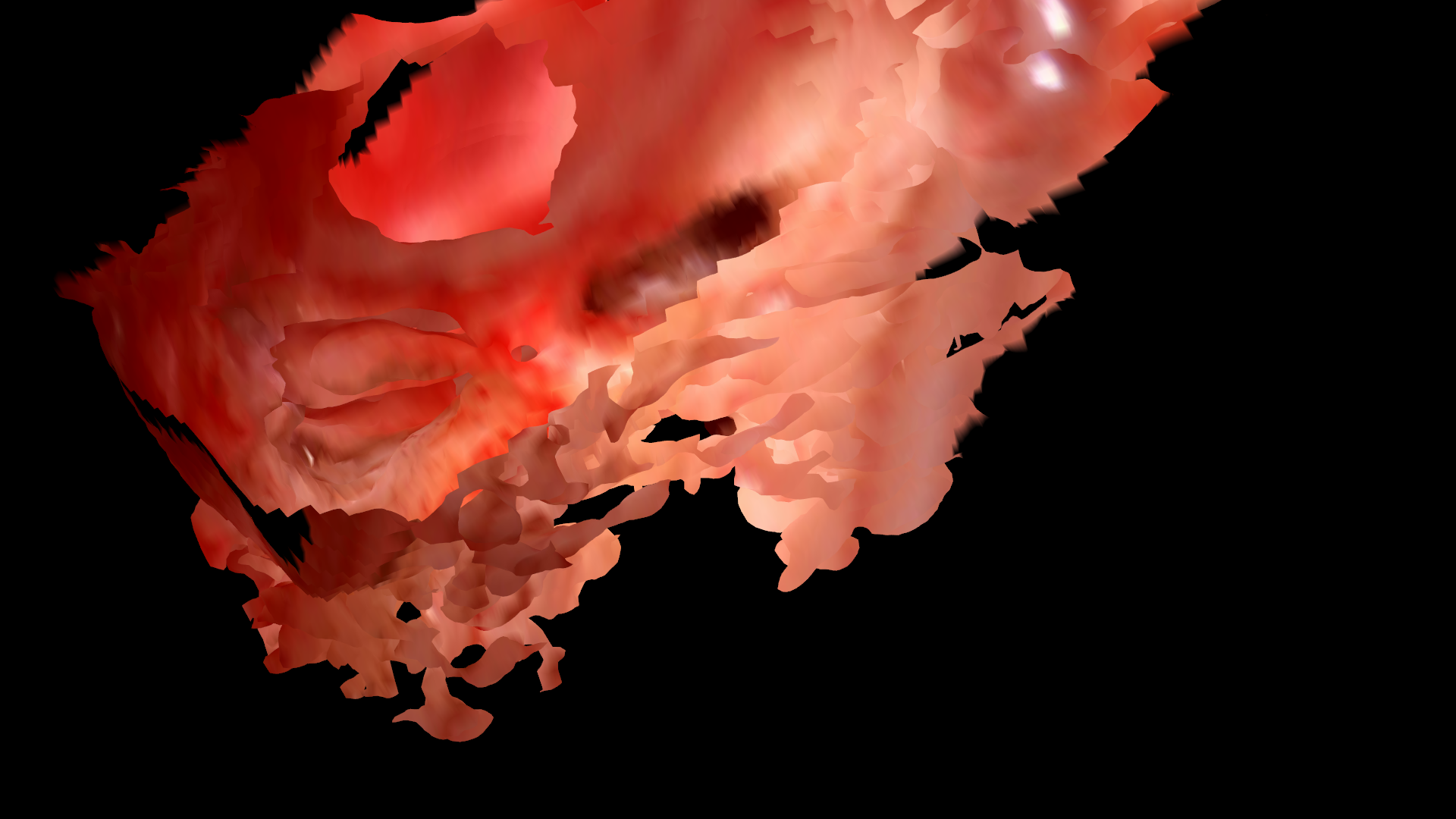}
        \subcaption{Pose 8 - PyVista}
    \end{minipage}
\end{minipage}

\vspace{\vspacing}
\begin{minipage}{\pairsize\textwidth}
    \centering
    \begin{minipage}{\imgsize\linewidth}
        \centering
        \includegraphics[width=\linewidth]{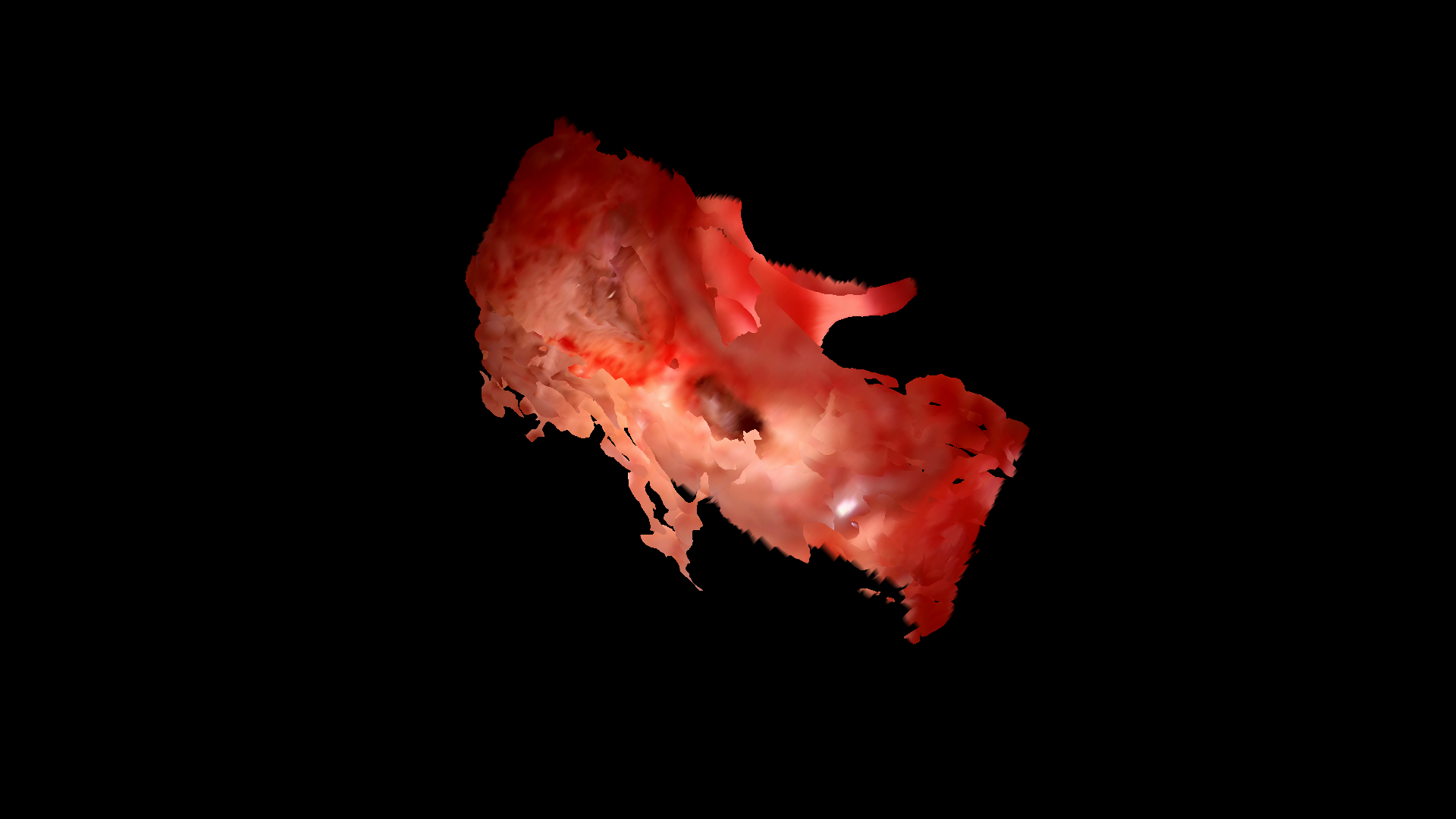}
        \subcaption{Pose 9 - Pytorch3D}
    \end{minipage}\hfill
    \begin{minipage}{\imgsize\linewidth}
        \centering
        \includegraphics[width=\linewidth]{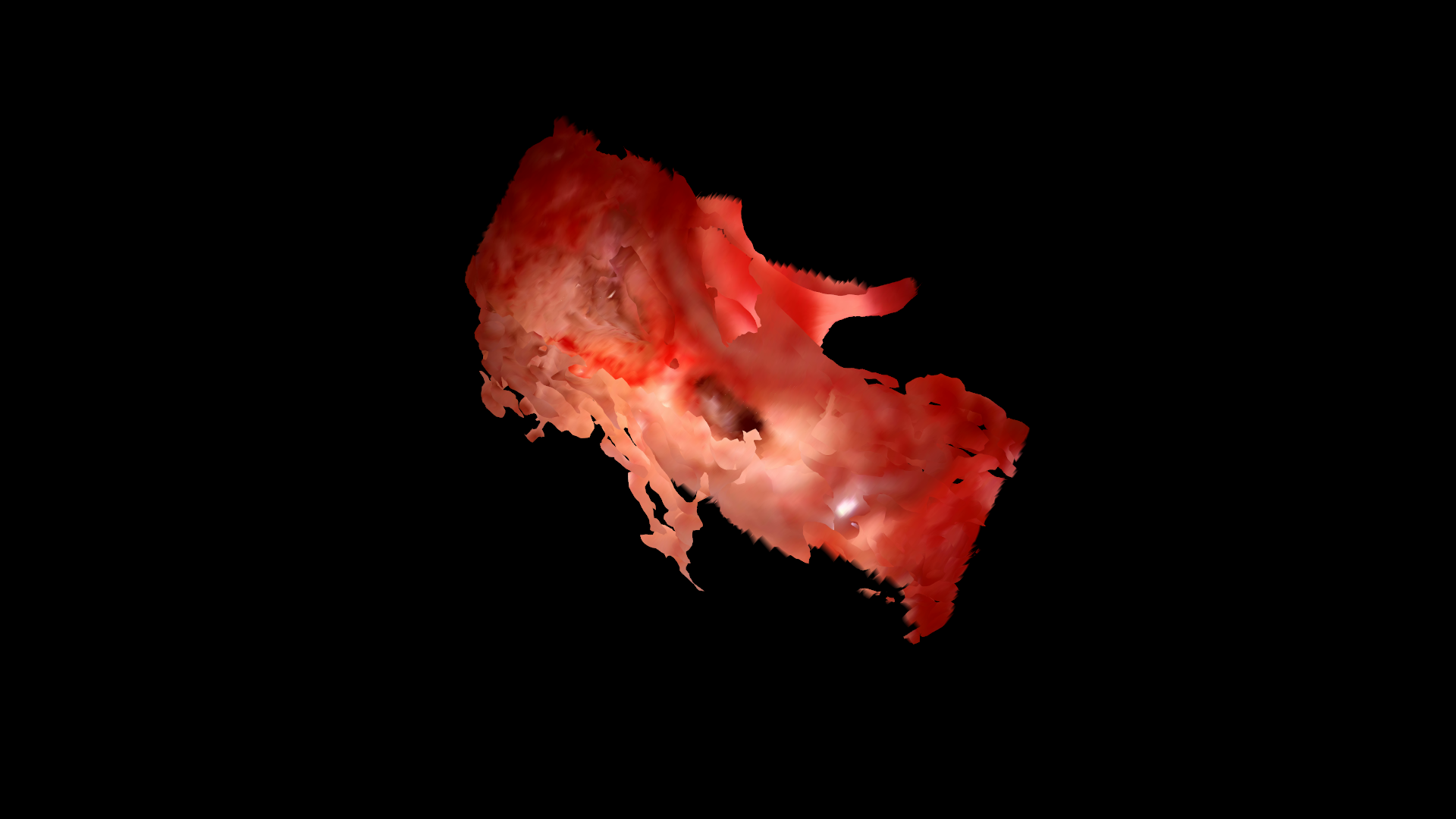}
        \subcaption{Pose 9 - PyVista}
    \end{minipage}
\end{minipage}\hfill
\begin{minipage}{\pairsize\textwidth}
    \centering
    \begin{minipage}{\imgsize\linewidth}
        \centering
        \includegraphics[width=\linewidth]{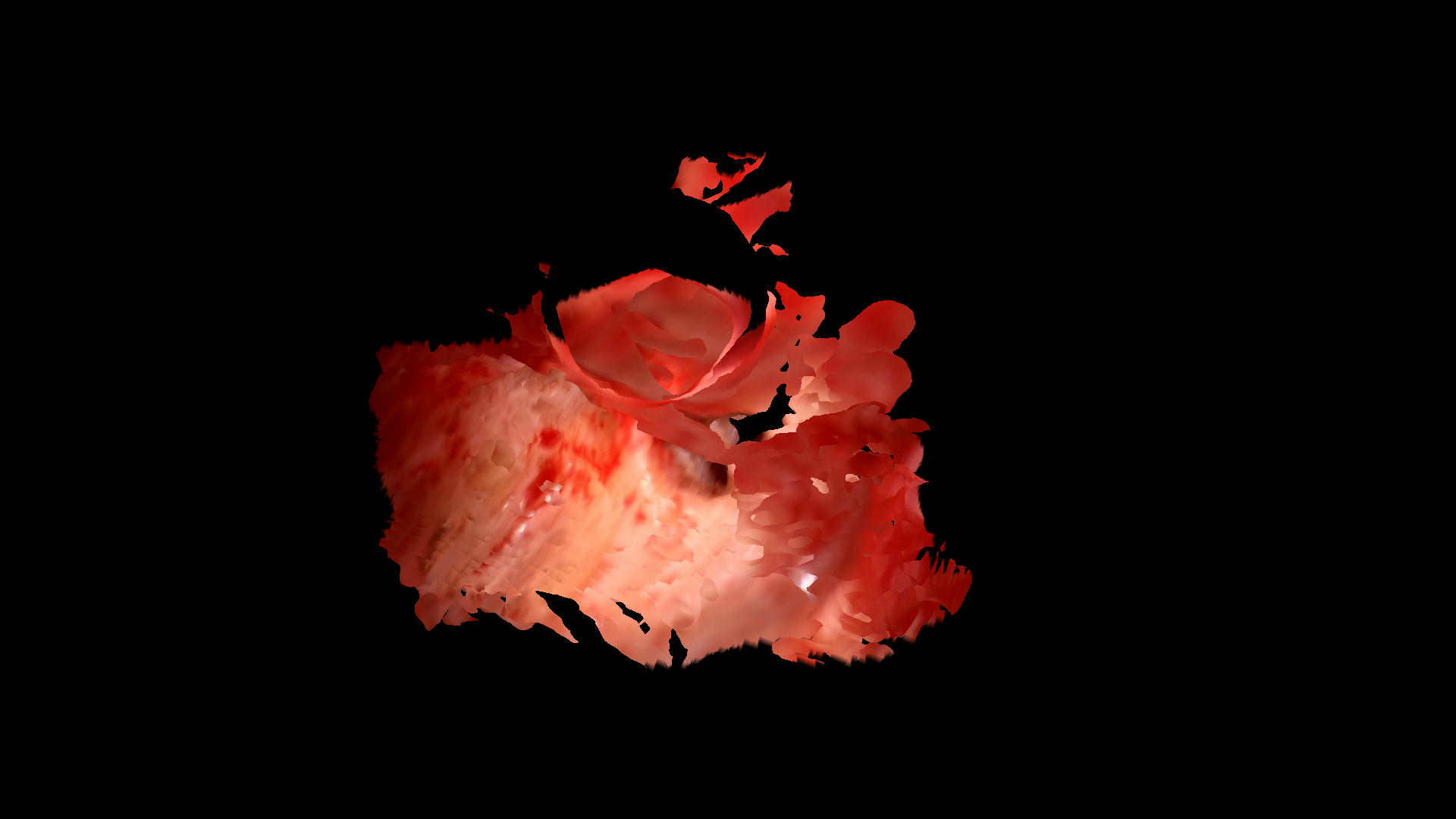}
        \subcaption{Pose 10 - Pytorch3D}
    \end{minipage}\hfill
    \begin{minipage}{\imgsize\linewidth}
        \centering
        \includegraphics[width=\linewidth]{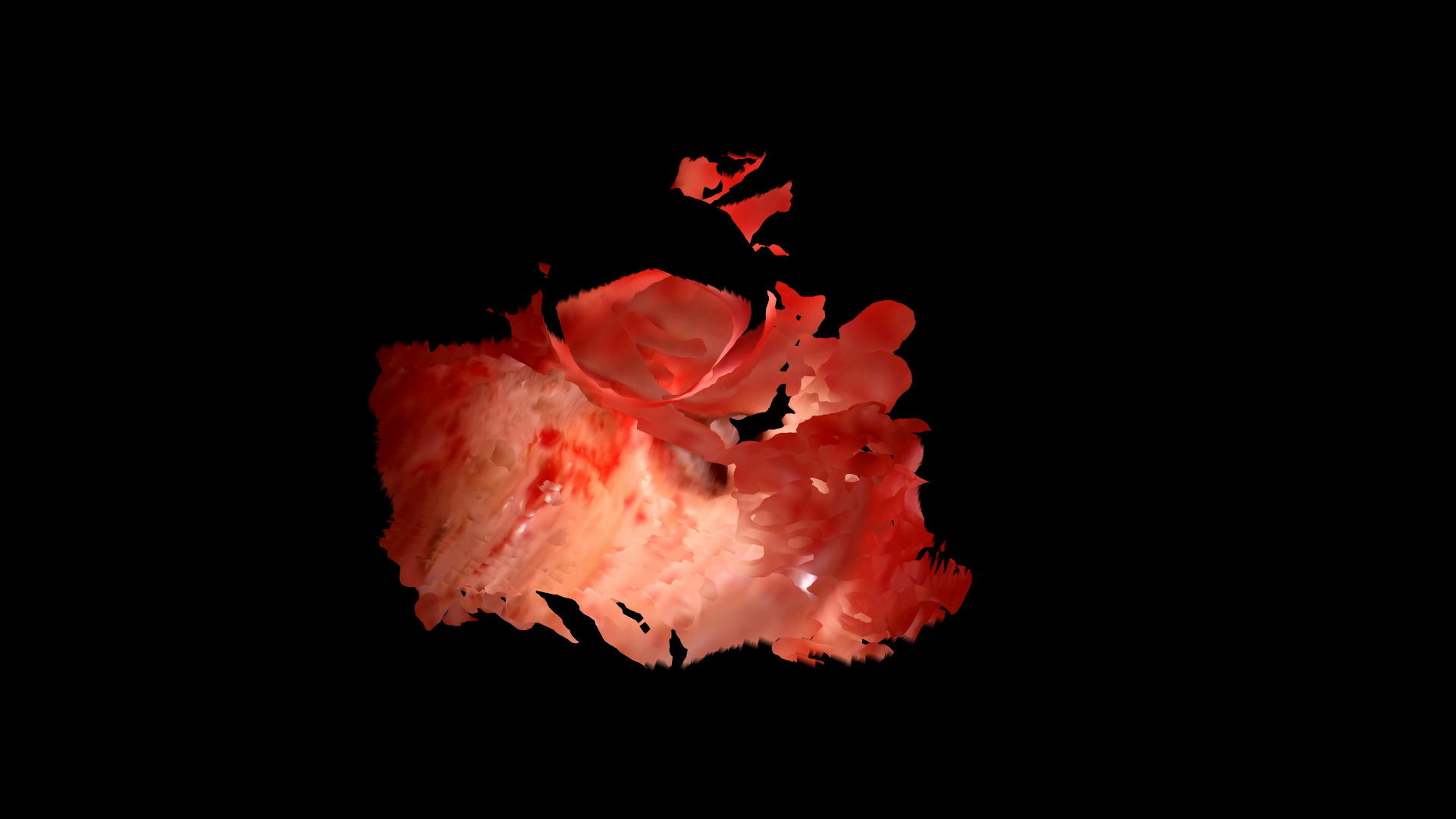}
        \subcaption{Pose 10 - PyVista}
    \end{minipage}
\end{minipage}

\vspace{\vspacing}
\begin{minipage}{\pairsize\textwidth}
    \centering
    \begin{minipage}{\imgsize\linewidth}
        \centering
        \includegraphics[width=\linewidth]{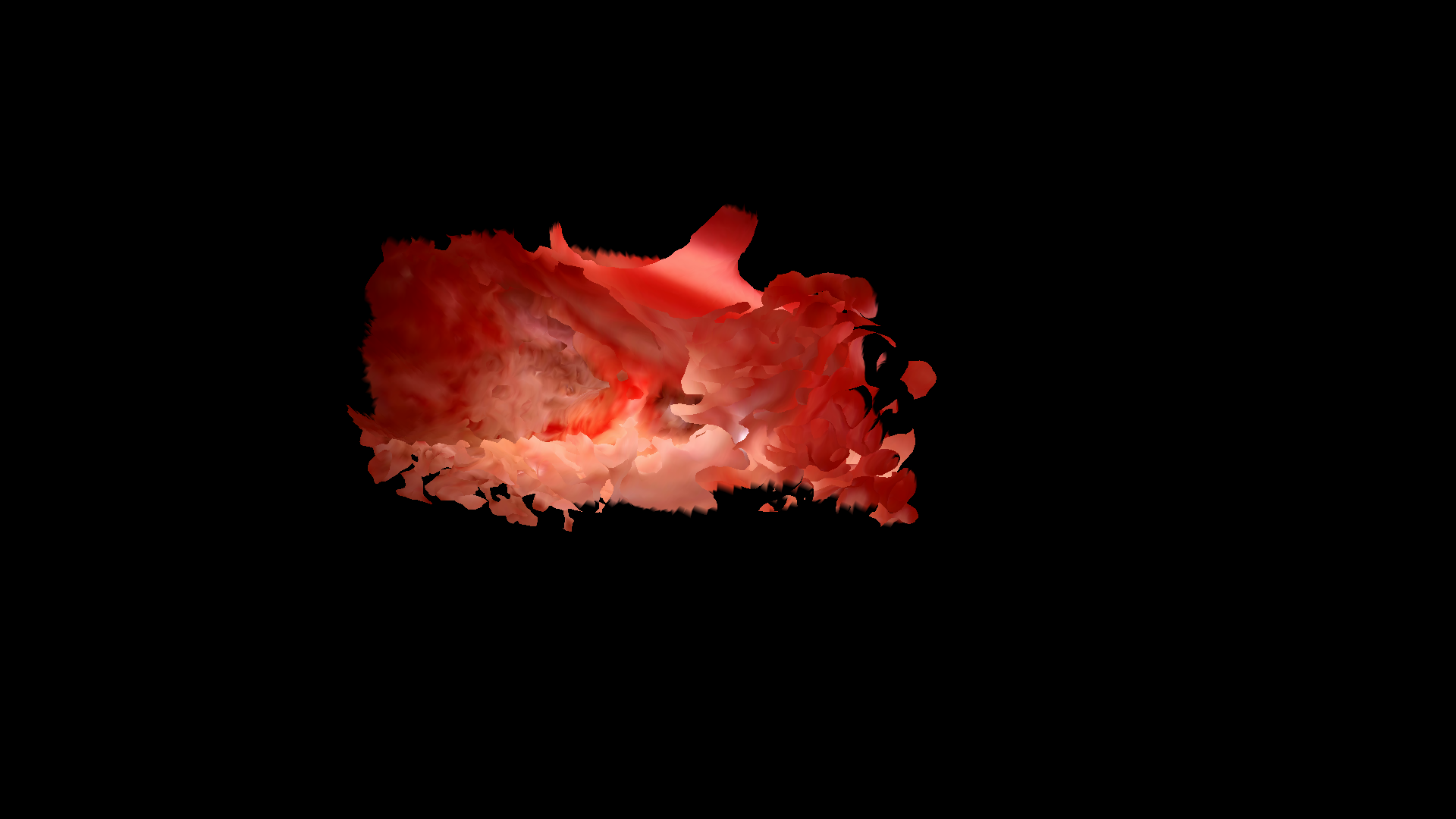}
        \subcaption{Pose 11 - Pytorch3D}
    \end{minipage}\hfill
    \begin{minipage}{\imgsize\linewidth}
        \centering
        \includegraphics[width=\linewidth]{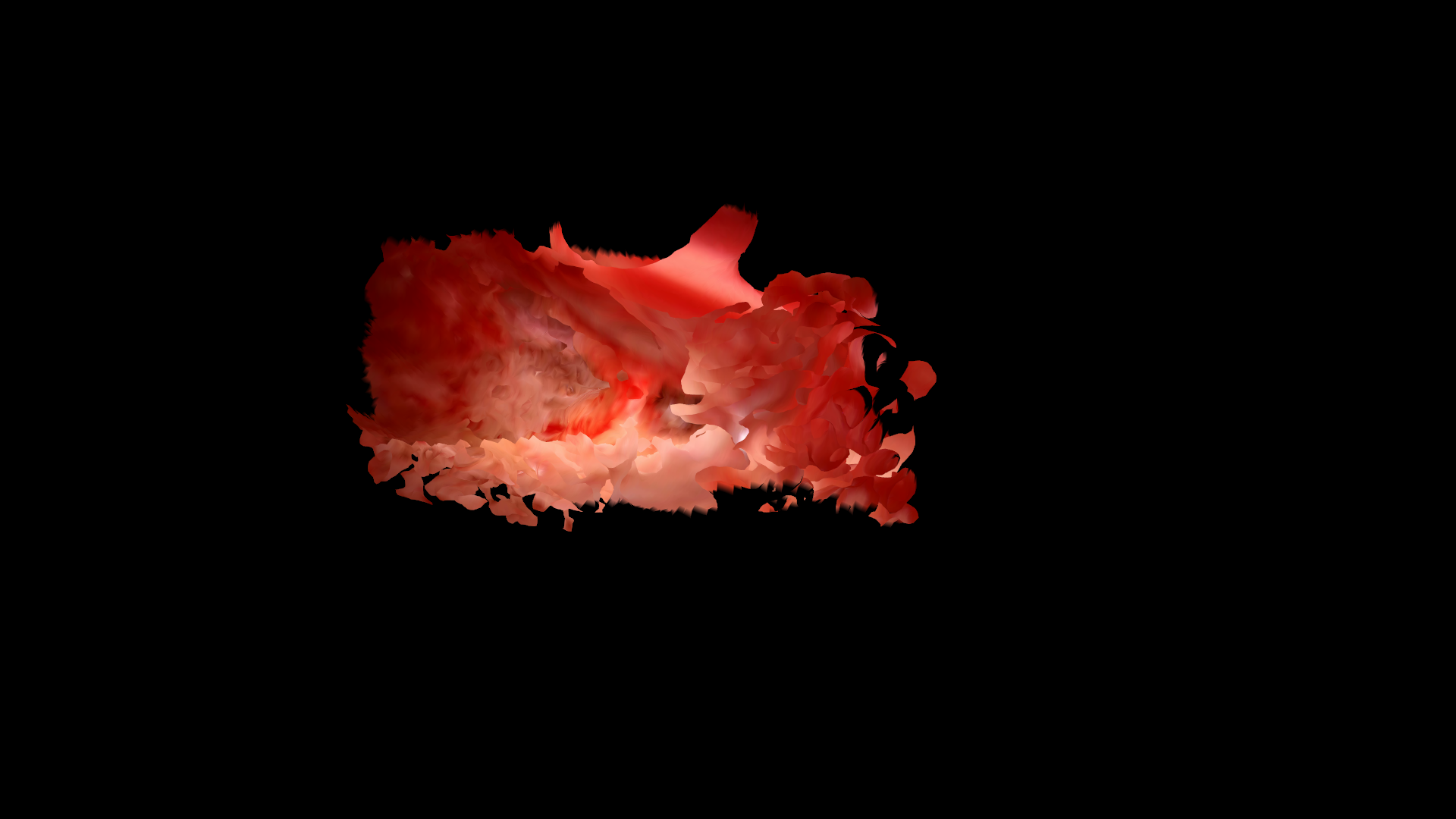}
        \subcaption{Pose 11 - PyVista}
    \end{minipage}
\end{minipage}\hfill
\begin{minipage}{\pairsize\textwidth}
    \centering
    \begin{minipage}{\imgsize\linewidth}
        \centering
        \includegraphics[width=\linewidth]{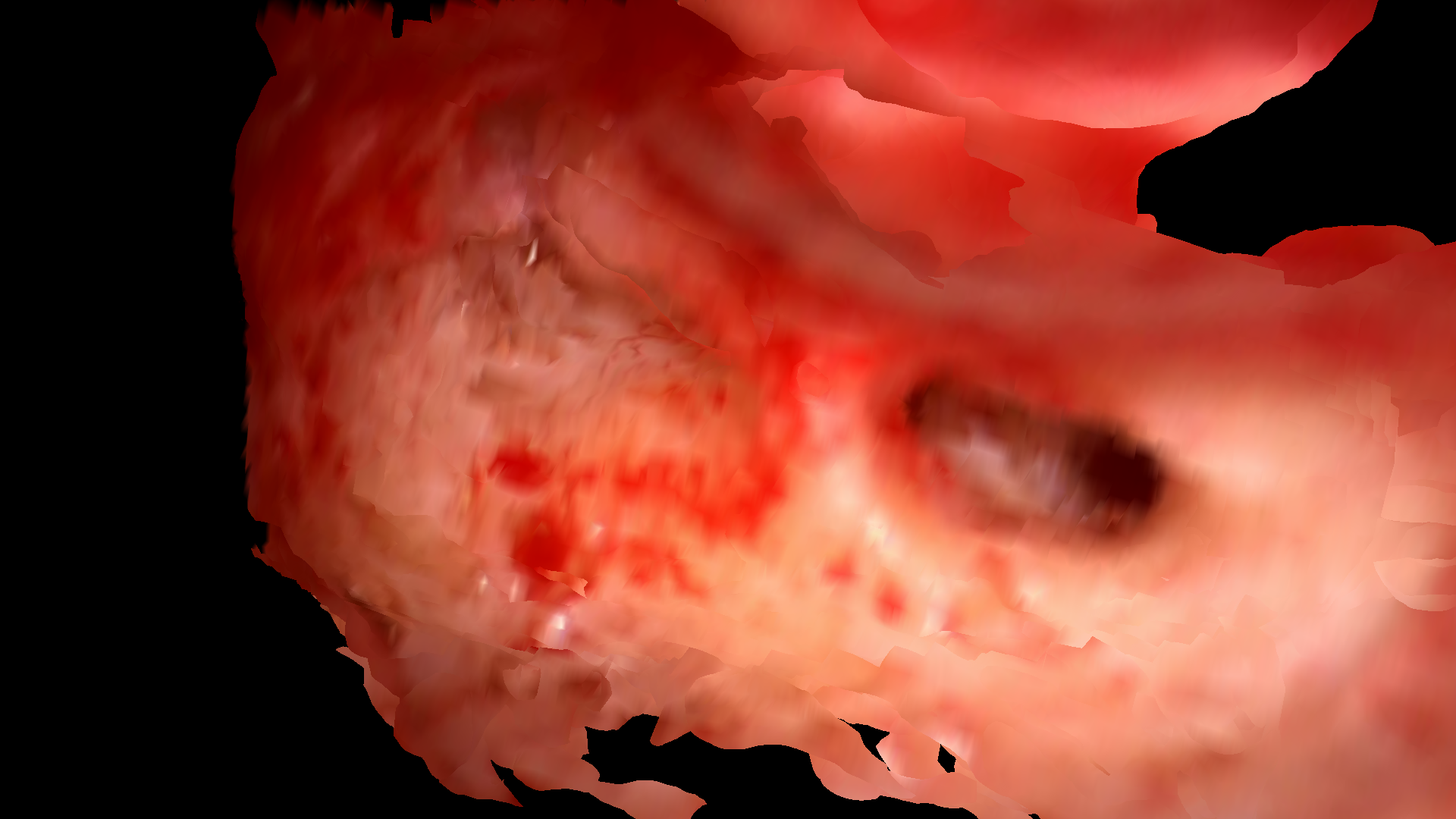}
        \subcaption{Pose 12 - Pytorch3D}
    \end{minipage}\hfill
    \begin{minipage}{\imgsize\linewidth}
        \centering
        \includegraphics[width=\linewidth]{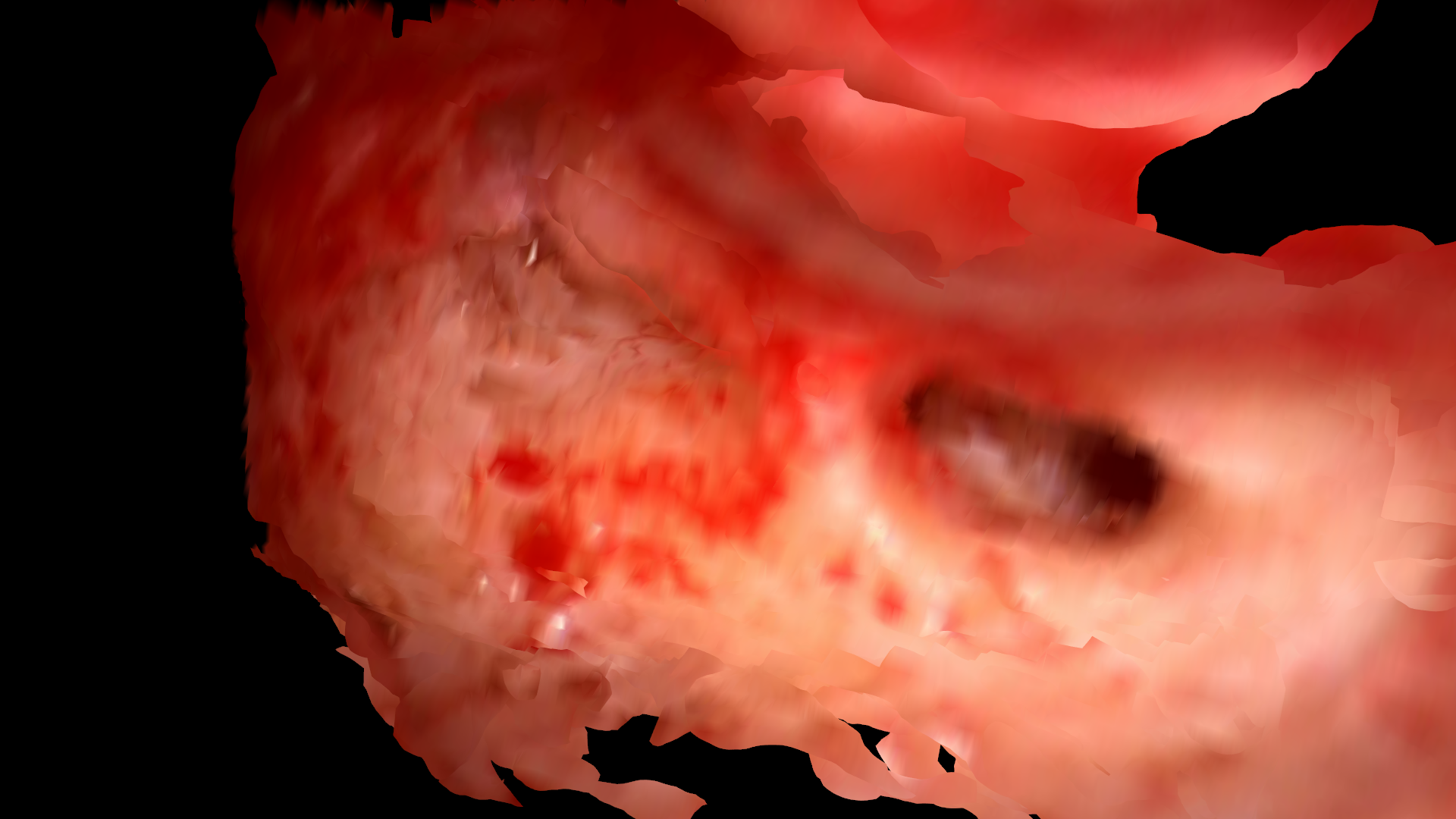}
        \subcaption{Pose 12 - PyVista}
    \end{minipage}
\end{minipage}




\caption{Multi-view Synthesis using Pytorch3D and PyVista render in each sub-figure.}
\label{fig:pytorch3d_pyvista_render}
\end{figure}

\section{Discussion and Conclusion}
Our paper introduces an efficient method for texturing a post-mastoidectomy mesh using corresponding surgical microscopy views. This will facilitate generating a large amount of synthetic views for training video processing methods and has the potential to be applied to other surgical procedures. In this work, we introduced a fast and convenient method for texturing a given post-mastoidectomy surface using a corresponding surgical frame. The resulting textures show a high degree of realism and open up numerous possibilities for various downstream tasks, including generating new views for synthesized videos and training methods to perform pose estimation from microscopy views. We are currently using these synthetic views to train such a pose estimation deep learning network and will report on the current results at the conference.

\acknowledgments
This work was supported in part by grants R01DC014037 and R01DC008408 from the NIDCD. This work is solely the responsibility of the authors and does not necessarily reflect the views of this institute.

\bibliographystyle{spiebib}
\bibliography{citation}
\end{document}